\documentclass[11pt]{article}

\usepackage[preprint]{acl}

\usepackage{times}
\usepackage{latexsym}

\usepackage[T1]{fontenc}

\usepackage[utf8]{inputenc}

\usepackage{microtype}

\usepackage{inconsolata}
\usepackage{booktabs}
\usepackage{multirow}
\usepackage{todonotes}
\usepackage{csquotes}

\usepackage{placeins}

\usepackage[table]{xcolor}
\usepackage{pgf}

\definecolor{deltaneg}{RGB}{178,24,43}
\definecolor{deltapos}{RGB}{33,102,172}
\newcommand{\maxintensity}{55}

\newcommand{\maxCOMET}{1}
\newcommand{\maxBLEU}{1}
\newcommand{\maxOLMES}{1}
\newcommand{\maxMTIF}{1}
\newcommand{\maxCOMETavg}{1}
\newcommand{\maxPct}{1}
\newcommand{\maxCOMETxling}{1}
\newcommand{\maxBLEUxling}{1}
\newcommand{\maxAccOne}{1}
\newcommand{\maxAccTwo}{1}
\newcommand{\maxAccThree}{1}
\newcommand{\maxFemAcc}{1}
\newcommand{\maxMascAcc}{1}
\newcommand{\maxAvgAcc}{1}
\newcommand{\maxFormalAcc}{1}
\newcommand{\maxInformalAcc}{1}
\newcommand{\maxAGIEval}{1}
\newcommand{\maxHumanEval}{1}
\newcommand{\maxHumanEvalPlus}{1}
\newcommand{\maxDROP}{1}
\newcommand{\maxGPQA}{1}
\newcommand{\maxGSMEightK}{1}
\newcommand{\maxIFEval}{1}
\newcommand{\maxTruthfulQA}{1}

\newcommand{\deltacell}[3]{%
  \pgfmathsetmacro{\ratio}{min(abs(#2)/(#3+0.0001),1)*\maxintensity}%
  \xdef\ratio{\ratio}%
  \ifdim #2 pt<0pt
    \cellcolor{deltaneg!\ratio!white}#1%
  \else
    \cellcolor{deltapos!\ratio!white}#1%
  \fi
}

\usepackage{graphicx}
\graphicspath{{figures/}{latex/figures/}}
\newcommand{\fitwidth}[1]{\resizebox{\ifdim\width>\linewidth\linewidth\else\width\fi}{!}{#1}}
\usepackage{tikz}
\usepackage{pifont}
\usepackage{amsmath}
\usepackage[nameinlink]{cleveref}
\crefname{appendixsec}{Appendix}{Appendices}
\Crefname{appendixsec}{Appendix}{Appendices}

\title{Fine-Tuning LLMs for Translation: General Forgetting Mitigation\\Does Not Preserve MT-Specific Instruction Following}

\author{Niklas Scholz\textsuperscript{1} \, David Thulke\textsuperscript{1,2} \, {Abdallah Nasir\textsuperscript{1,3}} \\ \textbf{Will Allred\textsuperscript{1}} \, \textbf{Evgeny Matusov\textsuperscript{1}} \, \textbf{Hermann Ney\textsuperscript{1,2}} \\ \\
        \textsuperscript{1}AppTek GmbH, Aachen, Germany\\
        \textsuperscript{2}Machine Learning and Human Language Technology, RWTH Aachen University, Germany\\
        \textsuperscript{3}Applied Science Private University, Amman, Jordan}

\begin{document}
\maketitle
\begin{abstract}
Fine-tuning large language models on parallel data improves translation quality but can cause catastrophic forgetting.
Mitigation methods are generally evaluated by retention on general benchmarks.
We ask whether these findings transfer to machine translation (MT) fine-tuning and to MT-specific instruction following (MT-IF):
instructions that modify a translation, such as formality, grammatical gender, and length control.
We compare methods anchored to auxiliary data, to model outputs, and to the base model parameters, first in a screening study with Llama 3.2 1B Instruct, then on Llama 3.1 8B Instruct fine-tuned on bidirectional Arabic-English or Spanish-English data.
Elastic Weight Consolidation preserves general capabilities best in both stages; on the 8B Spanish model the average score on general benchmarks drops 1.7 points versus 11.0 for standard fine-tuning, yet its scores for formality and grammatical gender control remain close to standard fine-tuning.
Only data mixing with control-task examples preserves these controls, but its gains do not transfer to unseen prompts for the same task.
\end{abstract}

\section{Introduction}

Instruction-tuned large language models (LLMs) translate well between high-resource languages \citep{robinson-etal-2023-chatgpt} and can modify their translations in response to natural language instructions.
Translation quality is lower for language pairs and domains that are poorly represented in the training data. Fine-tuning on parallel data can reduce this gap. In our experiments, it raises COMET from 45.6 to 71.5 for Amharic-to-English translation.

Fine-tuning on a narrow task can cause catastrophic forgetting, in which performance degrades on capabilities not represented in the fine-tuning data. One capability at risk during MT fine-tuning is following instructions that modify a translation. For example, a prompt can request a formal translation, select the masculine or feminine forms used for a person mentioned in the source, or impose a length constraint. Conventional neural machine translation (NMT) systems can support these controls, but typically require dedicated mechanisms and annotated training data.
In applications such as subtitling, the controls can be part of the translation task \citep{matusov-etal-2019-customizing}.

\begin{figure}[t]
\centering
\definecolor{teasergood}{RGB}{87,171,39}
\definecolor{teaserbad}{RGB}{204,7,30}
\newcommand{\teaseryes}[1]{\node[anchor=east, fill=white, inner xsep=3pt, inner ysep=1pt, font=\scriptsize, text=teasergood] at ([xshift=-3pt,yshift=1.7ex]#1.south east) {\ding{51}~formal};}
\newcommand{\teaserno}[1]{\node[anchor=east, fill=white, inner xsep=3pt, inner ysep=1pt, font=\scriptsize, text=teaserbad] at ([xshift=-3pt,yshift=1.7ex]#1.south east) {\ding{55}~informal};}
\newcommand{\teaserhead}[3]{\makebox[\linewidth]{\textbf{#1}\hfill{\scriptsize\textcolor{black!75}{COMET #2\enspace General #3}}}}
\begin{tikzpicture}
\tikzset{
  teaserbox/.style={draw=black!60, rounded corners=2pt, inner xsep=5pt,
    inner ysep=4pt, align=left, font=\footnotesize, anchor=north west,
    text width=\dimexpr\columnwidth-12pt\relax},
}
\node[teaserbox, fill=black!8] (prompt) at (0,0) {%
\textbf{Prompt:} Translate the following sentence into Spanish while maintaining a formal tone.\\
\textbf{Sentence:} I am with you there,\\ what is your favorite movie?};
\node[anchor=north west, inner xsep=5pt, inner ysep=1pt, font=\scriptsize,
  text=black!75] (legend) at ([yshift=-3pt]prompt.south west)
  {COMET: translation quality, General: avg. general capabilities, $\uparrow$ is better};
\node[teaserbox, draw=teasergood] (base) at ([yshift=-3pt]legend.south west) {%
\teaserhead{Llama 3.1 8B Instruct}{86.0}{\textbf{66.8}}\\
Estoy de acuerdo \textbf{\textcolor{teasergood}{con usted}},\\ ¿cuál es \textbf{\textcolor{teasergood}{su}} película favorita?};
\teaseryes{base}
\node[teaserbox, draw=teaserbad] (sft) at ([yshift=-6pt]base.south west) {%
\teaserhead{Standard SFT}{\textbf{87.0}}{55.8}\\
Estoy de acuerdo \textbf{\textcolor{teaserbad}{contigo}},\\ ¿cuál es \textbf{\textcolor{teaserbad}{tu}} película favorita?};
\teaserno{sft}
\node[teaserbox, draw=teaserbad] (ewc) at ([yshift=-6pt]sft.south west) {%
\teaserhead{Elastic Weight Consolidation}{\textbf{87.0}}{63.3}\\
Estoy de acuerdo \textbf{\textcolor{teaserbad}{contigo}},\\ ¿cuál es \textbf{\textcolor{teaserbad}{tu}} película favorita?};
\teaserno{ewc}
\node[teaserbox, draw=teasergood] (mix) at ([yshift=-6pt]ewc.south west) {%
\teaserhead{Data Mixing}{\textbf{87.0}}{60.6}\\
Estoy de acuerdo \textbf{\textcolor{teasergood}{con usted}},\\ ¿cuál es \textbf{\textcolor{teasergood}{su}} película favorita?};
\teaseryes{mix}
\end{tikzpicture}
\caption{Formality-controlled translation from CoCoA-MT \citep{nadejde-etal-2022-cocoa} before and after fine-tuning Llama 3.1 8B Instruct on 1M Spanish$\leftrightarrow$English sentence pairs, with each model's COMET (translation quality, FLORES es$\leftrightarrow$en) and average General capabilities scores. EWC recovers most of the general-capability loss but not the instructed formality; only Data Mixing with control-task examples retains it.}
\label{fig:teaser}
\end{figure}
}

A number of methods to mitigate forgetting have been proposed.
They differ in the reference signal they use to mitigate forgetting, which we call the anchor: auxiliary data that represents the capabilities to be kept, the output distribution of a model, or the parameters of the base model (\Cref{sec:methods}).
Existing evaluations fine-tune on data from domains such as mathematical reasoning and code, and measure retention with general benchmarks \citep{biderman2024loralearnsforgets,yang-etal-2024-self,diao_entropy-adaptive_2026}.
It remains unclear how well the mitigation methods transfer when MT is the fine-tuning task, and whether they protect MT-specific instruction following (MT-IF).

We address these questions for instruction-tuned models in two stages.
Stage 1 is a screening study with Llama 3.2 1B Instruct, fine-tuned separately on Amharic-to-English, Arabic-to-English, and Spanish-to-English data.
We compare the mitigation methods of \Cref{sec:methods} across hyperparameter settings, measuring translation quality and general capabilities (\Cref{sec:evalForgetting}).
Stage 2 is the main experiment.
We apply selected configurations to Llama 3.1 8B Instruct and fine-tune it on bidirectional Arabic-English or Spanish-English data.
In addition to translation quality and general capabilities, we evaluate MT-IF on formality, grammatical gender, and length control (\Cref{sec:mt-if-eval}).

On the 1B model, standard fine-tuning decreases the average score on the general benchmarks by 6.3 to 8.1 points across the three language pairs.
Elastic Weight Consolidation (EWC) limits this decrease to 2.5 or less, and its configurations trace the Pareto frontier of learning against forgetting.
The Stage 2 results show a different pattern for MT-IF. On the 8B model, EWC retains general capabilities, but its scores for formality and grammatical gender control remain close to those of standard fine-tuning (illustrated in \Cref{fig:teaser}). Among the methods evaluated in Stage 2, only data mixing with examples of the corresponding control task preserves these controls. Ablations show no consistent transfer between grammatical gender and formality control.
Our contributions are:
\begin{enumerate}
    \item A comparison of forgetting mitigation methods anchored to auxiliary data, to model outputs, and to the base model parameters, for MT fine-tuning of instruction-tuned LLMs, across three language pairs and two model sizes.
    \item Evidence from formality, grammatical gender, and length control that retention on general benchmarks can coexist with losses in MT-IF, so general benchmark scores alone can miss capabilities affected by MT fine-tuning.
\end{enumerate}

\section{Related Work}
\label{sec:related_work}

Our study draws on two lines of research, the mitigation of catastrophic forgetting during LLM fine-tuning and the control of translation attributes such as formality, grammatical gender, and length. %

\paragraph{Forgetting when fine-tuning LLMs on MT data.}
\citet{stap-etal-2024-fine} observe that fine-tuning pre-trained LLMs on parallel data improves translation quality but degrades formality steering, few-shot translation, and document-level translation.
They fine-tune pretrained-only base models, elicit the affected abilities with few-shot prompting, and evaluate a single mitigation method, mixing in monolingual data.
We instead start from instruction-tuned models, compare eleven methods, and measure zero-shot instruction following, the interface through which these controls are used in practice.
\citet{liu-niehues-2025-conditions} study the loss of translation ability for languages unseen during fine-tuning and find that the relative scale of model and data is a primary determinant, while parameter-efficient fine-tuning offers no clear advantage over full fine-tuning.

\paragraph{Forgetting mitigation for LLM fine-tuning.}
We draw the methods we compare from this literature \citep{kirkpatrick2017ewc,rolnick2019experiencereplaycontinuallearning,hu2021loralowrankadaptationlarge,tian_rethinking_2024,yang_model_2024,zheng_spurious_2025,wu_mitigating_2025,diao_entropy-adaptive_2026} and describe each of them in \Cref{sec:methods}.
The evidence behind them was collected outside MT.
Earlier approaches were validated on supervised and reinforcement learning benchmarks \citep{kirkpatrick2017ewc,rolnick2019experiencereplaycontinuallearning} and on vision and language classification tasks \citep{tian_rethinking_2024}, while recent work fine-tunes on mathematical reasoning, code, medical, or agentic data and measures retention with general benchmarks \citep{biderman2024loralearnsforgets,wu_mitigating_2025,diao_entropy-adaptive_2026}.

Our experiments focus on mitigation during off-policy SFT, where the training responses are fixed rather than sampled from the current model. Recent comparisons, including one on MT, find that on-policy reinforcement learning often retains prior capabilities better than SFT \citep{shenfeld2025rlsrazoronlinereinforcement,chen2026retainingdoingroleonpolicy,su-etal-2026-reinforcement}, although \citet{luo2026rlforgetscontinualpolicy} report substantial forgetting during continual reinforcement learning on multimodal reasoning tasks. These studies do not evaluate retention of translation-specific instruction following.

\paragraph{Controlled translation in conventional NMT.}
Conventional NMT supports controlled translation of attributes such as politeness, length, grammatical gender, and formality through labeled data and dedicated mechanisms, including source- and word-level tags, modified positional encodings, and constrained rescoring or reranking \citep{sennrich-etal-2016-controlling,lakew-etal-2019-controlling,saunders-etal-2020-neural,wu-etal-2023-improving}.
\citet{raunak-etal-2024-instruction} fine-tune a single NMT model on parallel and task-specific instruction data, where each instruction is a short task description, such as \enquote{informal}, prepended to the source inside dedicated demarcation tokens.
Instruction-tuned LLMs instead accept such requests as free-form natural language prompts.
We study whether fine-tuning on parallel data degrades this ability and whether forgetting-mitigation methods preserve it.

\paragraph{MT instruction-following benchmarks.}
Attribute-specific MT benchmarks evaluate phenomena relevant to controlled translation.
WinoMT \citep{stanovsky-etal-2019-evaluating} and MuST-SHE \citep{bentivogli-etal-2020-gender} target gender translation accuracy,
CoCoA-MT \citep{nadejde-etal-2022-cocoa} provides contrastive formal and informal references,
MT-GenEval \citep{currey-etal-2022-mt} adds gender-balanced counterfactual and contextual examples, and
mGeNTE \citep{savoldi-etal-2025-mind} instead evaluates gender-neutral translation.
We adapt CoCoA-MT and MT-GenEval, together with FLORES, into targeted evaluations of formality, grammatical gender, and length control, as described in \Cref{sec:mt-if-eval}.
Broader recent benchmarks assess translation and instruction adherence jointly.
IF-MT tests two to four verifiable instructions for English-to-Chinese and English-to-Spanish translation and scores translation quality and adherence separately \citep{rei-etal-2026-tower}.
Concurrent work introduces IFMTBench, which evaluates single and composed instructions across seven languages and several lexical, contextual, structural, and stylistic constraints \citep{sun2026ifmtbenchcomprehensivebenchmarkmultilingual}.

\section{Mitigation Methods}
\label{sec:methods}
The compared methods differ in the reference signal they use to mitigate forgetting during fine-tuning.
We call this signal the anchor and group the methods by it: auxiliary data that represents the capabilities to be kept (\Cref{sec:methods-data}), the output distribution of a model on the fine-tuning data (\Cref{sec:methods-outputs}), and the parameters of the base model (\Cref{sec:methods-params}).
The anchor bounds what a method can preserve, which we return to in \Cref{sec:discussion}.
Several of the methods are orthogonal and could be combined, but we evaluate each in isolation to compare their individual contributions.

\subsection{Anchoring to Auxiliary Data}
\label{sec:methods-data}
\paragraph{Data Mixing} \cite{rolnick2019experiencereplaycontinuallearning} replays a fraction of general-domain instruction data into the fine-tuning data instead of training on MT alone.

\paragraph{Model Merging} \cite{yang_model_2024} combines the parameters of two independently trained models. We merge a model fine-tuned on MT with a model trained on general-domain data by linear interpolation, that is, weight averaging.

\paragraph{Elastic Weight Consolidation (EWC)} \cite{kirkpatrick2017ewc} adds a quadratic penalty on the distance from the base parameters, scaled by a hyperparameter $\lambda$ and weighted per parameter by the diagonal of the empirical Fisher information estimated on the auxiliary data. Its penalty is in parameter space, but the importance estimate that shapes it comes from the auxiliary data, which is why we group it here.
\citet{liu2026elasticweightconsolidationright} propose \textbf{EWC-DR}, which reverses the logits when computing Fisher information and thereby reduces the redundant protection of parameters that the standard EWC applies.

\medskip\noindent All methods in this group draw their anchor from the same subset of the T\"ulu 3 SFT mixture (\Cref{sec:trainingDataInfo}), so what they retain is bounded by the capabilities that this data covers.

\subsection{Anchoring to Model Outputs}
\label{sec:methods-outputs}
\paragraph{KL Divergence Regularization.} We add the exact KL divergence between the output distributions of the model being trained and a frozen copy of the base model as a penalty scaled by $\lambda$ on the SFT loss \cite{shenfeld2025rlsrazoronlinereinforcement}.

\paragraph{Selective Token Masking (STM)} removes tokens from the loss instead of penalizing divergence. \citet{wu_mitigating_2025} observe that fine-tuning on LLM-generated responses forgets less than fine-tuning on ground-truth data, tracing this to the lower token-level perplexity of generated text, and therefore mask ground-truth tokens whose perplexity exceeds a fixed threshold.
We instead set the threshold dynamically and mask the $X\%$ of tokens with the highest perplexity under a frozen copy of the base model.

\paragraph{Confident Conflict Masking (CC).} \citet{diao_entropy-adaptive_2026} identify confident conflicts, tokens with low probability and low entropy, as a cause of forgetting, because fitting them overwrites a confident prior. Masking the tokens in the bottom $X\%$ of both quantities mitigates forgetting, but discards useful training signal.

\paragraph{Entropy-Adaptive Fine-Tuning (EAFT)} addresses this by weighting the loss of each token by its entropy normalized over the top-$K$ distribution, instead of discarding the token \cite{diao_entropy-adaptive_2026}.

\medskip\noindent CC and EAFT take the output from the model being trained rather than from a frozen copy. All methods in this group are computed on the MT fine-tuning data, so they bound how far the model moves on translation inputs and place no constraint on its behavior on other prompts.

\subsection{Anchoring to the Base Model Parameters}
\label{sec:methods-params}
\paragraph{Freeze} keeps the bottom $N$ layers at their base values and trains the remaining ones. \citet{zheng_spurious_2025} propose this on the grounds that task alignment is concentrated in the bottom layers and is disrupted during the first optimization steps.

\paragraph{Selective Projection Decay (SPD)} \cite{tian_rethinking_2024} modifies the Adam optimizer to apply weight decay toward the pre-trained weights selectively. A gradient condition determines for each parameter tensor whether the update is heading toward a higher-loss region, and only those tensors receive the penalty, weighted by $\lambda$, while the remaining ones are updated without intervention.

\paragraph{Low-Rank Adaptation (LoRA)} \citet{hu2021loralowrankadaptationlarge} restricts fine-tuning updates to a low-rank subspace, which is believed to keep the model close to its base weights and thus limit forgetting \citep{sun2023exploringimpactlowrankadaptation}. However, \citet{biderman2024loralearnsforgets} show this holds only insofar as LoRA also learns less, with high-rank LoRA forgetting nearly as much as full fine-tuning in some settings.

\begin{figure*}[t]
\centering
\includegraphics[width=\textwidth]{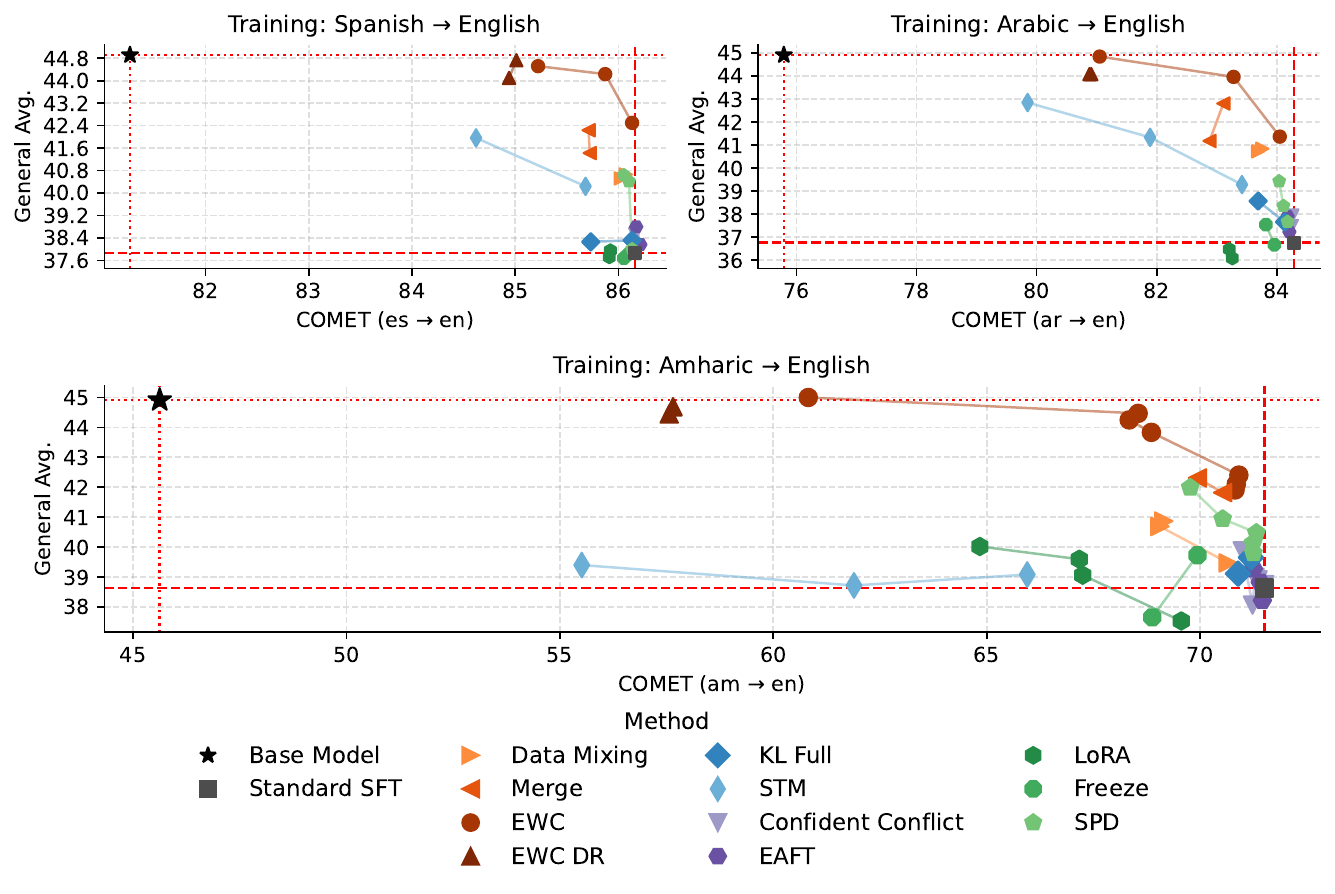}
\caption{Stage 1: Translation quality (COMET) versus general capabilities (General Avg) for all evaluated configurations per training language on the Llama 3.2 1B Instruct model. Lines connect hyperparameter settings within each method; dashed lines mark the SFT baseline.}
\label{fig:stage1Scatter}
\end{figure*}

\section{Evaluation of Model Capabilities}
\label{sec:evalForgetting}

We evaluate translation quality with COMET and BLEU, retention on general benchmarks, and three MT-specific controls: formality, grammatical gender, and length. These controlled-translation evaluations are targeted diagnostics of forgetting rather than a general benchmark of MT instruction following.

\subsection{Translation Quality}
\label{sec:mt-eval}
We report reference-based COMET with \texttt{Unbabel/wmt22-comet-da} \citep{rei-etal-2022-comet} and corpus BLEU with sacreBLEU \citep{post-2018-call}, using the default \texttt{13a} tokenizer for our language pairs. Before scoring, hypotheses and references are punctuation- and whitespace-normalized.

\subsection{General Capabilities}
Our general-capability benchmarks are a subset of the Tülu 3 evaluation suite \citep{lambert_tulu_2024}, which covers knowledge recall, reasoning, math, coding, and instruction following.
From its development split, we take GSM8K \citep{cobbe_training_2021}, DROP \citep{dua-etal-2019-drop}, TruthfulQA \citep{lin-etal-2022-truthfulqa}, Codex HumanEval \citep{chen_evaluating_2021}, Codex HumanEval+ \citep{liu_is_2023}, and IFEval \citep{zhou_instruction-following_2023}, the last of which assesses general instruction following.
From its unseen split, we take AGIEval English \citep{zhong-etal-2024-agieval} and GPQA \citep{rein_gpqa_2023}, both zero-shot with chain-of-thought prompting.
We run all of them with the task formulations and prompts of the Tülu 3 evaluation regime, using the OLMES evaluation standard \citep{gu-etal-2025-olmes}.
We report the mean of the eight task scores as the general average (General Avg).

\subsection{Machine Translation-Specific IF}
\label{sec:mt-if-eval}

To evaluate MT-IF capabilities, we assess models on the following three tasks closely related to MT:

\paragraph{Formality Control.} To assess formality, we prompt the model to produce a formal or informal translation on the CoCoA-MT dataset \citep{nadejde-etal-2022-cocoa} for \texttt{en}$\rightarrow$\texttt{de} and \texttt{en}$\rightarrow$\texttt{es}, and score with its metric: a hypothesis is classified as formal if it contains at least one of the annotated formal marked phrases from the reference pair and no informal ones (symmetrically for informal), and is left unmatched otherwise. The reported accuracy is the proportion of matched hypotheses whose classified formality matches the requested tone.

\paragraph{Grammatical Gender Control.} To assess control over grammatical gender, we use the MT-GenEval benchmark \citep{currey-etal-2022-mt}, which provides counterfactual sentence pairs that differ in the gender forms used for a person mentioned in the text.
We instruct the model to produce a translation with the opposite gender forms of the source sentence (i.e., masculine $\rightarrow$ feminine, feminine $\rightarrow$ masculine).
The instruction asks the model to override the forms suggested by the source, which separates instruction following from the model's default choice of gender forms.
We follow the same evaluation procedure from the benchmark by checking for the absence of unique opposite-gender words in the hypothesis.

\paragraph{Length Control.} We adapt the FLORES-101 and FLORES-200 benchmarks \citep{goyal-etal-2022-flores,Costa-jussa2024} for the bidirectional translation between English and the training language. %
In particular, we add additional instructions on producing shorter or longer translations than the character count of the source.
As the produced sequence length is heavily influenced by fine-tuning, we compare whether the instruction affected the output by comparing the translation with longer/shorter instructions to the baseline translation (instead of measuring pure accuracy on IF). This shows the effect of the additional instruction on length control and whether the model actually considered the instruction. We additionally report the target/source (tgt/src) ratios to understand length trends better.

\begin{table*}[t]
\centering
\small
\fitwidth{%
\renewcommand{\maxCOMET}{26.0}
\renewcommand{\maxBLEU}{10.8}
\renewcommand{\maxOLMES}{6.3}
\begin{tabular}{lllccccc}
\toprule
& & & \multicolumn{2}{c}{am$\rightarrow$en} & \multicolumn{2}{c}{de$\rightarrow$en} & \\
\cmidrule(lr){4-5} \cmidrule(lr){6-7}
Anchoring & Method & Hyperparameters & COMET & BLEU & COMET & BLEU & General Avg \\
\midrule
\multicolumn{3}{l}{Llama 3.2 1B Instruct}  & 45.6 & 0.2 & 82.9 & 27.6 & 44.9 \\
\midrule
\multicolumn{3}{l}{Standard SFT} & \deltacell{\underline{71.5}}{25.9}{\maxCOMET} & \deltacell{10.7}{10.5}{\maxBLEU} & \deltacell{83.6}{0.8}{\maxCOMET} & \deltacell{30.7}{3.1}{\maxBLEU} & \deltacell{38.6}{-6.3}{\maxOLMES} \\
\midrule
\multirow{4}{*}{\shortstack[l]{Auxiliary\\Data}} & Data Mixing & Tülu (full) & \deltacell{69.1}{23.4}{\maxCOMET} & \deltacell{8.5}{8.3}{\maxBLEU} & \deltacell{\textbf{85.7}}{2.7}{\maxCOMET} & \deltacell{\textbf{34.2}}{6.6}{\maxBLEU} & \deltacell{40.7}{-4.2}{\maxOLMES} \\
& Merge & Trans=0.75, Tülu=0.25 & \deltacell{70.5}{24.9}{\maxCOMET} & \deltacell{9.9}{9.7}{\maxBLEU} & \deltacell{\underline{84.2}}{1.3}{\maxCOMET} & \deltacell{\underline{32.1}}{4.5}{\maxBLEU} & \deltacell{\underline{41.8}}{-3.1}{\maxOLMES} \\
& EWC & $\lambda$=1e2, Tülu (75k) & \deltacell{70.9}{25.3}{\maxCOMET} & \deltacell{9.9}{9.7}{\maxBLEU} & \deltacell{83.9}{1.0}{\maxCOMET} & \deltacell{31.4}{3.8}{\maxBLEU} & \deltacell{\textbf{42.4}}{-2.5}{\maxOLMES} \\
& EWC-DR & $\lambda$=1e3, Tülu (10k) & \deltacell{71.3}{25.7}{\maxCOMET} & \deltacell{10.6}{10.4}{\maxBLEU} & \deltacell{83.7}{0.8}{\maxCOMET} & \deltacell{30.7}{3.1}{\maxBLEU} & \deltacell{38.6}{-6.3}{\maxOLMES} \\
\midrule
\multirow{4}{*}{\shortstack[l]{Model\\Outputs}} & KL-Full & $\lambda$=0.0025 & \deltacell{71.2}{25.6}{\maxCOMET} & \deltacell{10.6}{10.4}{\maxBLEU} & \deltacell{83.6}{0.7}{\maxCOMET} & \deltacell{30.7}{3.1}{\maxBLEU} & \deltacell{39.7}{-5.3}{\maxOLMES} \\
& STM & X=0.15 & \deltacell{65.9}{20.3}{\maxCOMET} & \deltacell{6.7}{6.5}{\maxBLEU} & \deltacell{82.7}{-0.2}{\maxCOMET} & \deltacell{27.6}{0}{\maxBLEU} & \deltacell{39.1}{-5.8}{\maxOLMES} \\
& CC & X=0.45 & \deltacell{71.0}{25.4}{\maxCOMET} & \deltacell{10.5}{10.3}{\maxBLEU} & \deltacell{83.9}{1.0}{\maxCOMET} & \deltacell{31.3}{3.7}{\maxBLEU} & \deltacell{39.9}{-5.0}{\maxOLMES} \\
& EAFT & K=20, Weight=1.0 & \deltacell{\textbf{71.6}}{26.0}{\maxCOMET} & \deltacell{\textbf{11.0}}{10.8}{\maxBLEU} & \deltacell{84.0}{1.1}{\maxCOMET} & \deltacell{31.8}{4.2}{\maxBLEU} & \deltacell{38.9}{-6.0}{\maxOLMES} \\
\midrule
\multirow{3}{*}{\shortstack[l]{Base Model\\Parameters}} & Freeze & N=1 & \deltacell{69.9}{24.3}{\maxCOMET} & \deltacell{9.4}{9.2}{\maxBLEU} & \deltacell{83.5}{0.6}{\maxCOMET} & \deltacell{30.5}{2.9}{\maxBLEU} & \deltacell{39.7}{-5.2}{\maxOLMES} \\
& SPD & $\lambda$=2 & \deltacell{71.3}{25.7}{\maxCOMET} & \deltacell{\underline{10.9}}{10.7}{\maxBLEU} & \deltacell{83.9}{1.0}{\maxCOMET} & \deltacell{31.4}{3.8}{\maxBLEU} & \deltacell{40.5}{-4.4}{\maxOLMES} \\
& LoRA & Rank=64, lr=1e-5 & \deltacell{67.2}{21.6}{\maxCOMET} & \deltacell{7.5}{7.3}{\maxBLEU} & \deltacell{83.9}{1.0}{\maxCOMET} & \deltacell{31.1}{3.5}{\maxBLEU} & \deltacell{39.6}{-5.3}{\maxOLMES} \\
\bottomrule
\end{tabular}%
}
\caption{Stage 1: Best-performing configuration per method for Amharic, selected by the highest sum of Amharic-language COMET and general average score. The cell shading encodes the absolute differences from the base instruct model, normalized relative to the largest observed change within each metric group (across languages).}
\label{tab:bestAmharicScores}
\end{table*}

\section{Stage 1: Transfer of Mitigation Techniques to Machine Translation Fine-tuning}
\subsection{Data}
We fine-tune models on unidirectional translation data for three language pairs: Amharic $\rightarrow$ English, Arabic $\rightarrow$ English, and Spanish $\rightarrow$ English, each containing 1,000,068 samples. For Data Mixing, Model Merging, and for computing the Fisher matrix in EWC, we use subsets of the Tülu 3 SFT mixture \citep{lambert_tulu_2024} of varying sizes, from which we remove synthetic instruction-following data to prevent overlap with the IFEval benchmark used for evaluation (see \Cref{sec:trainingDataInfo} for more details on the distribution of our training data). 

\subsection{Experimental Setup}
\label{sec:stage1Setup}
In the first stage of our experiments, we fine-tune the Llama 3.2 1B Instruct model with the mitigation methods previously introduced, comparing various hyperparameter settings. We compare the fine-tuned models against two baselines: the unmodified instruct model and standard SFT.

\paragraph{Evaluation.} We report BLEU and COMET scores for translation on the training language pair and on German-to-English, included as an out-of-distribution language pair (\Cref{sec:mt-eval}). We use FLORES-101 \citep{goyal-etal-2022-flores} for Amharic and Spanish, and FLORES-200 \citep{Costa-jussa2024} for Arabic and German. We additionally evaluate the forgetting of general capabilities, as described in \Cref{sec:evalForgetting}.

\subsection{Results}
We present the MT scores and the general average of the best-performing configurations for Amharic in \Cref{tab:bestAmharicScores} (Arabic and Spanish in \Cref{app:stage1BestConfig}). These configurations are chosen based on the highest sum of COMET and general average scores. Additionally, we visualize the results of all 40 Amharic, 27 Spanish, and 28 Arabic configuration training runs in \Cref{fig:stage1Scatter} showing the trade-off between general capabilities and MT performance. 

Overall, we observe the SFT baselines of all three languages to maintain similar forgetting rates (Amharic: 6.3, Arabic: 8.1, Spanish: 7.0) despite the different improvements in learning the MT task (Amharic: 25.9, Arabic: 8.5, Spanish: 4.9). Even though German was not included as a language pair in training, German MT scores slightly improve after fine-tuning, indicating no forgetting on a similar translation task. 

\paragraph{Most methods collapse toward the SFT baseline.} Across all three training languages, the majority of evaluated configurations are clustered around the SFT baseline in \Cref{fig:stage1Scatter} (dashed lines). Only a small number of methods consistently separate from this cluster by forgetting less on general capabilities (General Avg) at comparable COMET scores. In particular, the EWC family (EWC, EWC-DR) and Merge appear in the upper-right region in all three plots, while Data Mixing and SPD show mediocre mitigation of forgetting. 

\paragraph{Most methods are consistent across languages.} The only two techniques showing clear differences between languages are Selective Token Masking (STM) and LoRA, each diverging in opposite directions when comparing Amharic with Spanish and Arabic. LoRA performs comparatively better on Amharic by trading a smaller gain on MT performance (COMET: +21.6 vs. +25.9 in standard SFT) for a slightly smaller drop in the general average ($-5.3$), whereas STM performs better on Arabic and Spanish than for Amharic where it significantly hindered learning (COMET: 65.9). 

\paragraph{EWC lies on the Pareto frontier.} Individual EWC configurations trade translation quality against general capabilities, spreading along the upper-right boundary of \Cref{fig:stage1Scatter} between standard SFT and the base model rather than occupying a single point.
Any method with a continuous coefficient produces such a curve between the two baselines, and the sweeps of Merge, Data Mixing, KL, and STM reach the region where COMET falls below standard SFT.
Their curves lie below the EWC curve: for almost every configuration of another method, some EWC setting reaches both a higher COMET and a higher general average, in all three languages.
The remaining frontier points are configurations of standard SFT and a few other methods that gain up to a few tenths of COMET over the best EWC setting at the cost of several points on the general average.

\paragraph{Implications for Stage 2 Experiments.} 
The language-dependent differences of LoRA and STM noted above suggest that the extent of learning required, rather than the language itself, drives the differences in forgetting mitigation trends.
As we are not aware of resources to evaluate MT-IF on Amharic, we reduce the Stage 2 experiments to Spanish and Arabic.
Additionally, as EWC performed better than EWC-DR, we drop the latter.

\renewcommand{\maxCOMET}{1.0}
\renewcommand{\maxBLEU}{3.8}
\renewcommand{\maxOLMES}{13.8}
\renewcommand{\maxMTIF}{39.6}

\begin{table*}[t]
\centering
\small
\fitwidth{%
\begin{tabular}{lllcccccc}
\toprule
& & & \multicolumn{2}{c}{es$\leftrightarrow$en} & \multicolumn{2}{c}{de$\leftrightarrow$en} & & es \\
\cmidrule(lr){4-5} \cmidrule(lr){6-7} \cmidrule(lr){9-9}
Anchoring & Method & Hyperparameters & COMET & BLEU & COMET & BLEU & General Avg & MT-IF Avg \\
\midrule
\multicolumn{3}{l}{Llama 3.1 8B Instruct} & 86.0 & 26.7 & 87.6 & 37.3 & 66.8 & 71.1 \\
\midrule
\multicolumn{3}{l}{Standard SFT} & \deltacell{87.0}{1.0}{\maxCOMET} & \deltacell{30.4}{3.7}{\maxBLEU} & \deltacell{87.4}{-0.2}{\maxCOMET} & \deltacell{37.2}{-0.1}{\maxBLEU} & \deltacell{55.8}{-11.0}{\maxOLMES} & \deltacell{32.3}{-38.8}{\maxMTIF} \\
\midrule
\multirow{12}{*}{\shortstack[l]{Auxiliary\\Data}} & \multirow{8}{*}{\shortstack[l]{Data\\Mixing}} & Tülu (full) & \deltacell{86.9}{0.9}{\maxCOMET} & \deltacell{30.1}{3.4}{\maxBLEU} & \deltacell{87.9}{0.3}{\maxCOMET} & \deltacell{38.2}{0.9}{\maxBLEU} & \deltacell{60.0}{-6.8}{\maxOLMES} & \deltacell{34.5}{-36.6}{\maxMTIF} \\
& & MT-IF (1600) & \deltacell{87.0}{1.0}{\maxCOMET} & \deltacell{30.5}{3.8}{\maxBLEU} & \deltacell{87.9}{0.3}{\maxCOMET} & \deltacell{38.4}{1.1}{\maxBLEU} & \deltacell{56.5}{-10.3}{\maxOLMES} & \deltacell{65.8}{-5.3}{\maxMTIF} \\
& & Tülu (800) + MT-IF (1600) & \deltacell{87.0}{1.0}{\maxCOMET} & \deltacell{30.3}{3.6}{\maxBLEU} & \deltacell{87.8}{0.2}{\maxCOMET} & \deltacell{37.8}{0.5}{\maxBLEU} & \deltacell{60.6}{-6.2}{\maxOLMES} & \deltacell{67.0}{-4.1}{\maxMTIF} \\
& & \quad + diff MT-IF prompt & \deltacell{87.0}{1.0}{\maxCOMET} & \deltacell{30.4}{3.7}{\maxBLEU} & \deltacell{87.9}{0.3}{\maxCOMET} & \deltacell{38.0}{0.7}{\maxBLEU} & \deltacell{60.9}{-5.9}{\maxOLMES} & \deltacell{33.7}{-37.4}{\maxMTIF} \\
& & Tülu (800) + Gender (800) & \deltacell{87.0}{1.0}{\maxCOMET} & \deltacell{30.4}{3.7}{\maxBLEU} & \deltacell{87.8}{0.2}{\maxCOMET} & \deltacell{37.7}{0.4}{\maxBLEU} & \deltacell{61.7}{-5.1}{\maxOLMES} & \deltacell{51.2}{-19.9}{\maxMTIF} \\
& & Tülu (800) + Formality (800) & \deltacell{87.0}{1.0}{\maxCOMET} & \deltacell{30.3}{3.6}{\maxBLEU} & \deltacell{87.4}{-0.2}{\maxCOMET} & \deltacell{37.1}{-0.2}{\maxBLEU} & \deltacell{61.6}{-5.2}{\maxOLMES} & \deltacell{45.5}{-25.6}{\maxMTIF} \\
\cmidrule(lr){2-9}
& Merge & Trans=0.5, Tülu+MT-IF=0.5 & \deltacell{87.0}{1.0}{\maxCOMET} & \deltacell{30.3}{3.6}{\maxBLEU} & \deltacell{87.8}{0.2}{\maxCOMET} & \deltacell{37.9}{0.6}{\maxBLEU} & \deltacell{61.5}{-5.3}{\maxOLMES} & \deltacell{36.8}{-34.3}{\maxMTIF} \\
\cmidrule(lr){2-9}
& \multirow{3}{*}{EWC} & $\lambda$=1e3, Tülu (10k) & \deltacell{87.0}{1.0}{\maxCOMET} & \deltacell{30.0}{3.3}{\maxBLEU} & \deltacell{87.6}{0.0}{\maxCOMET} & \deltacell{37.7}{0.4}{\maxBLEU} & \deltacell{65.1}{-1.7}{\maxOLMES} & \deltacell{35.3}{-35.8}{\maxMTIF} \\
& & $\lambda$=1e3, MT-IF (1600) & \deltacell{87.0}{1.0}{\maxCOMET} & \deltacell{30.4}{3.7}{\maxBLEU} & \deltacell{87.7}{0.1}{\maxCOMET} & \deltacell{38.0}{0.7}{\maxBLEU} & \deltacell{58.1}{-8.7}{\maxOLMES} & \deltacell{34.8}{-36.3}{\maxMTIF} \\
& & $\lambda$=1e3, Tülu (800) + MT-IF (1600) & \deltacell{87.0}{1.0}{\maxCOMET} & \deltacell{30.2}{3.5}{\maxBLEU} & \deltacell{87.7}{0.1}{\maxCOMET} & \deltacell{37.5}{0.2}{\maxBLEU} & \deltacell{63.3}{-3.5}{\maxOLMES} & \deltacell{35.0}{-36.1}{\maxMTIF} \\
\midrule
\multirow{4}{*}{\shortstack[l]{Model\\Outputs}} & KL-Full & $\lambda$=0.05 & \deltacell{86.9}{0.9}{\maxCOMET} & \deltacell{30.1}{3.4}{\maxBLEU} & \deltacell{87.6}{0.0}{\maxCOMET} & \deltacell{37.5}{0.2}{\maxBLEU} & \deltacell{57.8}{-9.0}{\maxOLMES} & \deltacell{32.1}{-39.0}{\maxMTIF} \\
& STM & X=0.25 & \deltacell{86.5}{0.5}{\maxCOMET} & \deltacell{28.2}{1.5}{\maxBLEU} & \deltacell{87.1}{-0.5}{\maxCOMET} & \deltacell{35.3}{-2.0}{\maxBLEU} & \deltacell{53.0}{-13.8}{\maxOLMES} & \deltacell{32.5}{-38.6}{\maxMTIF} \\
& CC & X=0.25 & \deltacell{87.0}{1.0}{\maxCOMET} & \deltacell{30.4}{3.7}{\maxBLEU} & \deltacell{87.4}{-0.2}{\maxCOMET} & \deltacell{36.8}{-0.5}{\maxBLEU} & \deltacell{55.0}{-11.8}{\maxOLMES} & \deltacell{31.5}{-39.6}{\maxMTIF} \\
& EAFT & K=20, Weight=1.0 & \deltacell{87.0}{1.0}{\maxCOMET} & \deltacell{30.2}{3.5}{\maxBLEU} & \deltacell{87.6}{0.0}{\maxCOMET} & \deltacell{37.4}{0.1}{\maxBLEU} & \deltacell{57.8}{-9.0}{\maxOLMES} & \deltacell{31.9}{-39.2}{\maxMTIF}\\
\midrule
\multirow{3}{*}{\shortstack[l]{Base Model\\Parameters}} & Freeze & N=4 & \deltacell{87.0}{1.0}{\maxCOMET} & \deltacell{30.3}{3.6}{\maxBLEU} & \deltacell{87.5}{-0.1}{\maxCOMET} & \deltacell{37.2}{-0.1}{\maxBLEU} & \deltacell{54.3}{-12.5}{\maxOLMES} & \deltacell{32.5}{-38.6}{\maxMTIF} \\
& SPD & $\lambda$=5 & \deltacell{86.8}{0.8}{\maxCOMET} & \deltacell{29.9}{3.2}{\maxBLEU} & \deltacell{87.8}{0.2}{\maxCOMET} & \deltacell{37.7}{0.4}{\maxBLEU} & \deltacell{60.2}{-6.6}{\maxOLMES} & \deltacell{53.3}{-17.8}{\maxMTIF} \\
& LoRA & Rank=32, Alpha=64 & \deltacell{86.9}{0.9}{\maxCOMET} & \deltacell{30.1}{3.4}{\maxBLEU} & \deltacell{87.6}{0.0}{\maxCOMET} & \deltacell{37.8}{0.5}{\maxBLEU} & \deltacell{61.3}{-5.5}{\maxOLMES} & \deltacell{33.2}{-37.9}{\maxMTIF} \\
\bottomrule
\end{tabular}%
}
\caption{Stage 2: MT, General Avg, and Avg MT-IF scores for bidirectional Spanish training on the Llama 3.1 8B Instruct model. Color shade of cells encodes the absolute difference to the base model, normalized within each metric group (across languages).}
\label{tab:spanish-bidrectional-overview}
\end{table*}

\begin{figure*}[t]
\centering
\includegraphics[width=0.9\textwidth]{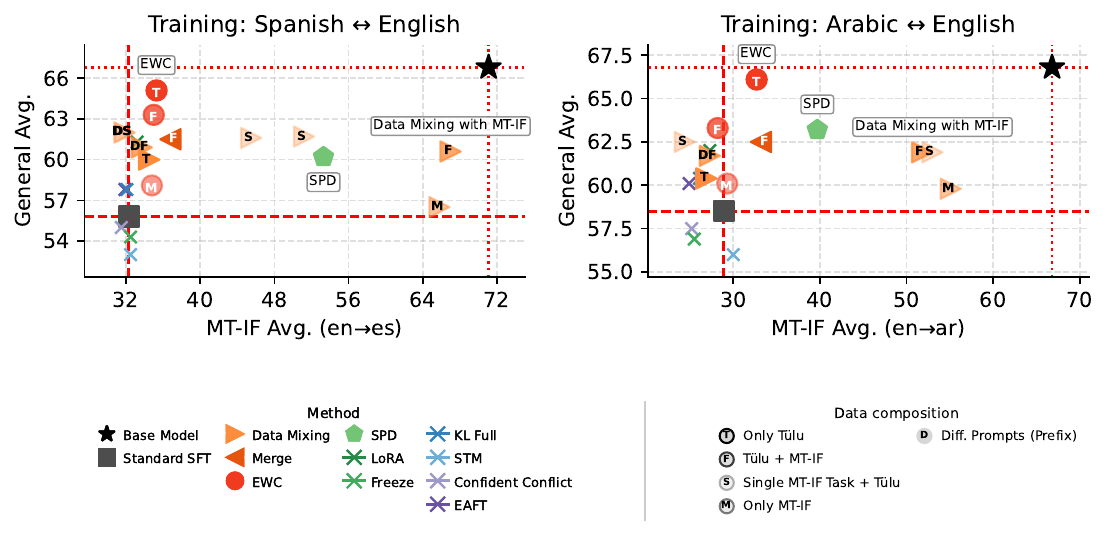}
\caption{Stage 2: General capabilities (General Avg) versus MT-IF capabilities per training language for fine-tuned, bidirectional Llama 3.1 8B Instruct models (MT-IF averaged over Formality, Grammatical Gender, and Length Control for Spanish; Length and Grammatical Gender Control for Arabic). Dotted/dashed red lines mark the base model / standard SFT reference points. Additional text labels within markers indicate the data composition.}
\label{fig:stage2MTIFScatterByLanguage}
\end{figure*}

\section{Stage 2: Comparison to Machine Translation Instruction Following Tasks}
\subsection{Data}
We fine-tune models with bidirectional, open data from two languages to improve reproducibility: Spanish $\leftrightarrow$ English and Arabic $\leftrightarrow$ English.
Each dataset has a total size of 1,000,000 samples, consisting of 500,000 pairs per direction (see \Cref{sec:trainingDataInfo} for detailed information on the training data distribution).

Alongside the Tülu 3 subsets used in Stage 1, we mix in data closely related to our MT-IF tasks: the training split of CoCoA-MT (Spanish, N=400; German N=400) \citep{nadejde-etal-2022-cocoa} and the dev split of MT-GenEval (Arabic, Spanish, and German evenly distributed; N=800) \citep{currey-etal-2022-mt}.
We primarily use the same prompts as during evaluation, but additionally run variants with a different prompt to assess whether the observed effects generalize to unseen prompts (see \Cref{app:prompts-formality,app:prompts-gender} for prompts used).

\subsection{Experimental Setup}
We move to Llama 3.1 8B Instruct, whose base performance on the MT-IF tasks is high enough to measure their forgetting, and to open, bidirectional training data.
Model size, translation directions, and data source therefore change together between the two stages.
We take the best performing configurations per method identified in the previous stage as a starting point and only slightly re-tune them for the larger model trained on open, bidirectional data, in contrast to re-screening all possible configurations.

\paragraph{Evaluation.}
We reuse the Stage 1 evaluation setup (COMET, BLEU, and the general average; \Cref{sec:stage1Setup}), but evaluate translation in both directions.
Beyond these scores, we evaluate our models on the three MT-IF control tasks, as defined in \Cref{sec:mt-if-eval}: Formality Control (Spanish, German; only for Spanish training runs), Grammatical Gender Control (Training Language, German), and Length Control (Training Language).
We evaluate MT-IF only in Stage 2, as CoCoA-MT and MT-GenEval only cover translations from English to other languages and the MT-IF capabilities of the 1B model are too low to measure forgetting.
We summarize the controls as the MT-IF average (MT-IF Avg): the mean of the training-language scores of the three tasks, where formality and grammatical gender contribute their average accuracy and length control contributes the mean compliance rate under the shorter and longer instructions.
For Arabic, the average covers only grammatical gender and length control, as CoCoA-MT provides no Arabic formality data.

\subsection{Results}
\subsubsection{Forgetting of General Capabilities}
We present the MT scores and the general average of our bidirectional training runs on Spanish in \Cref{tab:spanish-bidrectional-overview} and on Arabic in \Cref{tab:arabic-bidrectional-overview} (Appendix).
Standard SFT forgets more on general capabilities at 8B than at 1B in Stage 1, with the general average (base 66.8) dropping 11.0 points for Spanish and 8.3 points for Arabic, despite much smaller MT learning gains (COMET +1.0 vs. +4.9 at 1B for Spanish).
Since model size, translation directions, and data source change together between the stages, we cannot attribute this difference to one of them.
The ordering of the methods is largely stable across stages: EWC retains the most in both, Merge, SPD, and Data Mixing retain moderately, and the methods anchored to model outputs stay within two points of standard SFT.
Changes in ordering occur where Stage 1 differences were already small. CC and Freeze, within 1.3 points of standard SFT at 1B, now fall 0.8 to 1.6 points below it.
The exception is STM, which retained 4.1 to 4.5 points more than standard SFT for Spanish and Arabic in Stage 1 at the cost of COMET, and at 8B loses 2.5 to 2.8 points more than standard SFT (53.0 for Spanish, 56.0 for Arabic) while still learning less.

\paragraph{EWC retains general capabilities; other methods only partially.}
Consistent with Stage 1 of our experiments, EWC with Tülu for parameter importance estimation keeps the drop in the general average close to zero.
With only the Tülu subset, EWC limits the drop to $-1.7$ for Spanish and $-0.7$ for Arabic, far smaller than the $-3.6$ to $-13.8$ drops of other methods.
This mirrors the Stage 1 pattern for EWC, now under bidirectional translation with a stronger base model.
LoRA, Merge, and SPD all show moderate retention of the general average. %

\subsubsection{Forgetting of MT-IF Capabilities}
We observe that while general capabilities have been preserved strongly or moderately by specific methods, such as EWC, this is not the case for MT-IF (i.e., Formality Control, Grammatical Gender Control, Length Control) (see \Cref{fig:stage2MTIFScatterByLanguage}).
Standard SFT drops the average MT-IF score by $-38.8$ for Spanish (71.1 $\rightarrow$ 32.3) and $-37.8$ for Arabic (66.8 $\rightarrow$ 28.9).
The only methods that separate from the large cluster around this baseline are SPD, with drops of $-17.8$ for Spanish and $-27.1$ for Arabic, and Data Mixing with MT-IF data (using the same prompts as evaluation), ranging from $-4.1$ to $-25.6$ for Spanish and from $-11.7$ to $-15.0$ for Arabic.
The Arabic mix containing only formality data, a control task that is not part of the Arabic MT-IF average, even underperforms SFT by 4.6 points.

\paragraph{EWC does not preserve MT-IF.} Although EWC retains general capabilities best, it falls into the cluster around the SFT baseline on MT-IF (see \Cref{fig:stage2MTIFScatterByLanguage}).
EWC with the 10k Tülu subset, the configuration with the smallest drop in the general average for both languages, reaches MT-IF averages of 35.3 for Spanish and 32.7 for Arabic, close to standard SFT (32.3 and 28.9).
Estimating parameter importance on additional MT-IF data does not change this: the general average drops further while the MT-IF averages stay near the SFT baseline.

\paragraph{Data Mixing preserves only the target MT-IF capability.}
We observe that Data Mixing with MT-IF data is the only method that retains most of the MT-IF capabilities (see \Cref{fig:stage2MTIFScatterByLanguage}).
However, this mitigation is limited to the specific MT-IF task that is included in the training data.
For example, Data Mixing with Tülu (800) + Gender (800) is only actively improving on grammatical gender control capabilities (to 77.9\% accuracy vs. standard SFT's 17.8\%), while showing more forgetting than standard SFT on Formality Control (53.7\% accuracy vs. 56.1\%) and Length Control (\Cref{tab:spanish-length-control}).
This shows that the retention of MT-IF capabilities is limited to the specific task that is included in the training data, and does not transfer to other MT-IF tasks.
On the mixed-in task itself, scores surpass even the base model in all cases except German formality control.

\paragraph{Same-prompt gains do not transfer to unseen prompts on the same task.} However, we observe clear differences between using the same prompt as during evaluation and different prompts in Spanish training runs (see \Cref{fig:stage2MTIFScatterByLanguage} for the dashed-edge markers that distinguish these variants). Evidently, for Data Mixing with different Formality Control Prompts, Formality scores reduce to an average accuracy of 53.8, which is even lower than the score of standard SFT (56.1). The recovery of MT-IF capabilities with Data Mixing is thus partially because of prompt memorization, rather than general MT-IF capability retention. 

\paragraph{Grammatical Gender and Formality Control.} Per-method scores are in \Cref{tab:spanish-gender-control,tab:arabic-gender-control,tab:spanish-formality-control-detailed} (Appendix).
Standard SFT reduces grammatical gender control scores in both training runs and formality control scores in the Spanish run by more than 30 points.
The majority of methods (CC, EAFT, Freeze, KL, LoRA, STM) stay close to these scores on both tasks.
SPD and Merge stay above these scores on both tasks.
On the training language pair, SPD retains more in the Spanish run (38.5 versus 23.4 for Merge on gender control, 72.0 versus 61.0 on formality control), while Merge retains more gender control in the Arabic run (34.3 versus 29.7).
\paragraph{Length Control.} Per-method length control scores are in \Cref{tab:spanish-length-control,tab:arabic-length-control} (Appendix).
The base model follows shorter and longer instructions relatively well.
After standard SFT, compliance collapses and the tgt/src ratios of the three control settings converge; almost every other method stays within this range.
SPD is the only method that keeps compliance with both instructions clearly above the SFT level.
While EWC (Tülu 10k) also partially follows the longer-translation instruction, its COMET scores under this instruction drop to 74.3 for Spanish and 71.1 for Arabic, well below other methods.
This suggests that EWC pushes the model toward producing longer sequences without ensuring correctness of translations.

\begin{figure*}[t]
\centering
 \includegraphics[width=\textwidth]{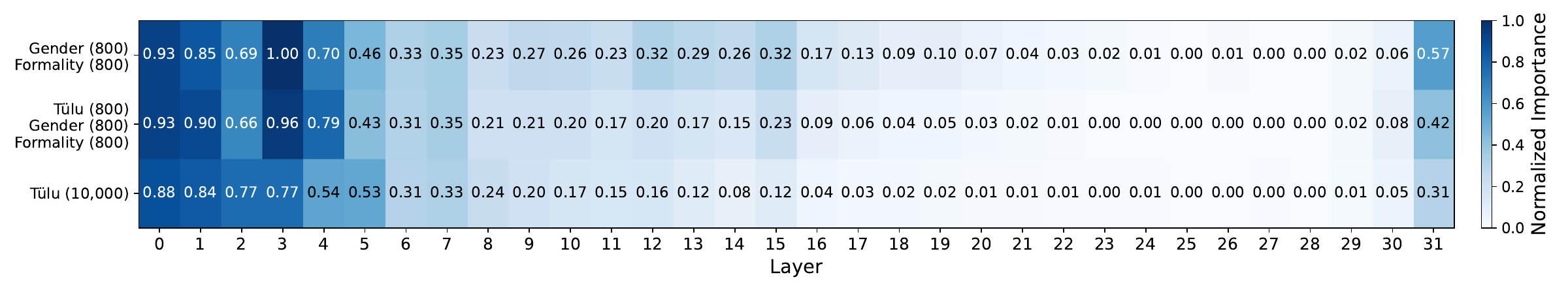}
\caption{Normalized layer-wise Fisher importance on Llama 3.1 8B Instruct, estimated on Tülu, MT-IF data, and their combination. The importance profiles largely coincide.}
\label{fig:fisherHeatmap}
\end{figure*}

\section{Analysis and Discussion}
\label{sec:discussion}

\paragraph{Why does EWC protect general capabilities but not MT-IF?}
\Cref{fig:fisherHeatmap} speaks against the simplest cause, a Fisher estimated on Tülu that misses the parameters behind MT-IF: the normalized layer-wise Fisher importance of the three anchor datasets used in Stage 2 has nearly the same profile, with most importance in layers 0--5 and the final layer (parameter-level differences remain possible).
Consistent with this, a Fisher estimated on MT-IF data alone did not preserve MT-IF in the Spanish run (average 34.8 versus 32.3 for standard SFT) and lowered the general average (58.1 versus 65.1 with the Tülu Fisher).
We therefore attribute the asymmetry to the MT gradients rather than to the importance estimate.
EWC with the Tülu Fisher reaches the same COMET as standard SFT (87.0) while keeping the general average at 65.1 versus 55.8, which is only possible if MT fine-tuning updates the parameters behind the general benchmarks weakly.
The MT-IF controls, in contrast, are exercised on translation inputs, so the parameters that carry them receive the MT gradient throughout fine-tuning, which a fixed quadratic penalty does not hold back.
We did not measure per-parameter drift, so this explanation remains a hypothesis.

\paragraph{Why do the methods anchored to model outputs fail?}
The methods in this group (KL, STM, CC, EAFT) compute their constraint on the MT training batches (\Cref{sec:methods-outputs}), so they bound the drift only on inputs on which the model must change and leave the evaluated behavior unconstrained.
KL regularization, for instance, penalizes divergence from the base model on the translation inputs themselves and places no constraint on the prompts used by the general benchmarks or the MT-IF evaluations.
CC additionally masks almost nothing on MT data.
In the Spanish 8B run (X=0.25), each condition alone is met by roughly a quarter of the tokens in a batch (on average 2773 with low entropy and 2619 with low probability), but fewer than one token per batch (0.36) meets both.
Tokens that are improbable under the base model are thus uncertain rather than confidently wrong, and with almost no tokens masked, CC reduces to standard SFT, in line with its scores in \Cref{tab:spanish-bidrectional-overview}.

\paragraph{SPD is a partial exception.}
Among the methods without auxiliary data, only SPD separates from the SFT cluster on MT-IF (average drops of 17.8 for Spanish and 27.1 for Arabic, against 38.8 and 37.8 for standard SFT), plausibly because its penalty applies per tensor and step, when the update heads toward a higher-loss region, rather than following a fixed importance estimate.
Our results do not isolate this mechanism, and the retained scores remain well below those of data mixing.

\paragraph{Implications.}
Methods anchored to the base model parameters assume that the fine-tuning task does not need to update the weights that carry the capability to be preserved.
Under MT fine-tuning this holds for general capabilities but not for MT-IF, which is itself translation behavior.
Data mixing escapes this limit by continuing to train the control rather than constraining the model, but as the Spanish formality run shows, the retention is tied to the training prompt.

\section{Conclusion}
We compared forgetting mitigation methods anchored to auxiliary data, to model outputs, and to the base model parameters, for fine-tuning Llama 3.2 1B Instruct and Llama 3.1 8B Instruct on parallel data for three language pairs.
EWC retains general capabilities best on both models. On the 8B model, it limits the drop in the general average to 1.7 points for Spanish and 0.7 for Arabic, against 11.0 and 8.3 for standard fine-tuning.
Its scores for formality and grammatical gender control nevertheless remain close to those of standard fine-tuning, even when the data used to estimate parameter importance includes control-task examples.
Of the methods compared, only mixing control-task examples into the training data preserves the corresponding control, and the effect does not extend to other control tasks or to unseen prompts for the same task.
Among the methods without control-task data, SPD retains the most MT-IF, at scores well below data mixing.

We conclude that retention on general benchmarks does not establish that MT-specific instruction following is preserved. Evaluations of MT fine-tuning should measure the controls the application requires, and preserving them currently requires matching data in the training mix. Open questions are whether broader control-task data, for example synthetic examples covering more tasks and prompt variants, can preserve MT-IF without overfitting to the seen prompts, and whether the findings extend to further controls such as glossary constraints and gender-neutral translation.

\section*{Limitations}

Our comparison is restricted to SFT-based adaptation and does not include on-policy reinforcement learning or preference optimization.
As discussed in \Cref{sec:related_work}, recent prior work showed that on-policy methods retain prior capabilities better than SFT.
Including them would have required model-generated trajectories and task-specific rewards or preference pairs, a substantial addition to our framework that would complicate a fair comparison.

We adapt the evaluation for MT-IF from related work and inherit some limitations.
For control of formality and grammatical gender, we follow the evaluation approaches proposed in CoCoA-MT and MT-GenEval.
These rely on predefined marked phrases and words and some translation variants may be missed.
MT-GenEval's masculine/feminine annotation, which our prompts follow, treats gender as binary and excludes non-binary referents.
We use it because its scoring is deterministic and cheap to run, whereas gender-neutral translation currently requires a trained classifier or an LLM judge for evaluation \citep{savoldi-etal-2025-mind}.
Our results therefore cover the selection between masculine and feminine forms and do not show whether MT fine-tuning degrades gender-neutral translation.
The MT-IF evaluation also covers only translation from English into Spanish, German, and Arabic, where CoCoA-MT and MT-GenEval provide data, and formality control only for Spanish and German.
The length-control score compares each instructed translation with the model's own uninstructed translation, so it measures whether the instruction changed the output length and not whether a length target was met.

Model size, translation directions, and data source change together between Stage 1 and Stage 2, and the configurations selected on the 1B model are only slightly re-tuned for the 8B model.
The design therefore does not show which of these factors causes the larger forgetting of standard SFT at 8B or the reversal for STM, and the Stage 2 configurations may not be optimal for the larger model.
We only use data from our internal standard data preparation pipeline which may introduce some biases.
Experiments are further limited to two model sizes of the Llama 3 family and to unidirectional and bidirectional translation of English and three target languages (two high and one low-resource).
While we assume that our findings generalize to other languages and models, this is still untested.

\makeatletter%
\ifacl@anonymize\else%
\section*{Acknowledgments}
This work was partially supported by NeuroSys, which as part of the initiative \enquote{Clusters4Future} is
funded by the Federal Ministry of Education and Research BMBF (funding ID 03ZU2106DD).

Generative AI assistants (OpenAI Codex and Claude Code) were used to proofread and help rephrasing parts of the manuscript, to help with LaTeX formatting, and to support with the implementation of the experiments.
All model generated suggestions were reviewed and verified by the authors, who take full responsibility for the content of this work.

\fi%
\makeatother%

\bibliography{anthology-1,anthology-2,custom}

\clearpage
\appendix
\crefalias{section}{appendixsec}
\crefalias{subsection}{appendixsec}
\crefalias{subsubsection}{appendixsec}

\section{Appendix}
\label{sec:appendix}

\subsection{Information on Training Data}
\label{sec:trainingDataInfo}
The MT training data consists of two groups. \textit{Base} comprises sentence-level parallel segments drawn from OPUS and other sources. \textit{CONCAT} is built from \textit{Base} segments for which document-level metadata is available: consecutive sentences from the same document are concatenated into a single training example.
\paragraph{Stage 1.}
For unidirectional training runs on Llama 3.2 1B Instruct, we use Amharic~$\rightarrow$~English, Arabic~$\rightarrow$~English, and Spanish~$\rightarrow$~English data (\Cref{tab:data-sizes-am-s1,tab:data-sizes-ar-s1,tab:data-sizes-es-s1}). Each language pair comprises 1,000,000 \textit{Base} sentence pairs plus 68 additional \textit{CONCAT} samples (1,000,068 total). Non-OPUS sources are aggregated as proprietary data; Amharic is drawn almost entirely from OPUS (primarily NLLB).

\begin{table}[t]
\centering
\small
\begin{tabular}{lrr}
\toprule
\textbf{Corpus} & \textbf{Base} & \textbf{CONCAT} \\
\midrule
OPUS-NLLB                    & 872,787 & --     \\
OPUS-CCAligned               & 47,563  & --     \\
OPUS-Tanzil                  & 39,617  & --     \\
OPUS-bible-uedin             & 23,635  & --     \\
OPUS-GoURMET                 & 6,279   & --     \\
OPUS-MultiCCAligned          & 4,093   & --     \\
OPUS-XLEnt                   & 3,803   & --     \\
OPUS-OpenSubtitles           & 1,289   & --     \\
OPUS-GlobalVoices            & 316     & 21     \\
OPUS-GNOME                   & 296     & 4      \\
OPUS-Ubuntu                  & 172     & 39     \\
OPUS-TED2020                 & 93      & 4      \\
OPUS-wikimedia               & 41      & --     \\
Proprietary data             & 16      & --     \\
\midrule
\textbf{Total}               & \textbf{1,000,000} & \textbf{68} \\
\bottomrule
\end{tabular}
\caption{Number of Amharic sentence pairs contributed by each corpus for the Stage~1 SFT training dataset. Non-OPUS sources are aggregated as proprietary data.}
\label{tab:data-sizes-am-s1}
\end{table}

\begin{table}[t]
\centering
\small
\begin{tabular}{lrr}
\toprule
\textbf{Corpus} & \textbf{Base} & \textbf{CONCAT} \\
\midrule
Proprietary data                   & 762,628 & 22     \\
OPUS-TED2020                       & 61,929  & 2      \\
OPUS-NeuLab-TedTalks               & 56,368  & 3      \\
OPUS-MultiUN                       & 35,934  & 19     \\
OPUS-UNPC                          & 28,406  & 18     \\
OPUS-GlobalVoices                  & 13,369  & --     \\
OPUS-OpenSubtitles                 & 13,150  & 4      \\
OPUS-Tatoeba                       & 9,211   & --     \\
OPUS-News-Commentary               & 4,659   & --     \\
OPUS-TED2013                       & 3,906   & --     \\
OPUS-ELRC                          & 3,846   & --     \\
OPUS-KDE4                          & 2,624   & --     \\
OPUS-Mozilla-I10n                  & 2,114   & --     \\
OPUS-tico-19                       & 946     & --     \\
OPUS-Ubuntu                        & 711     & --     \\
OPUS-UN                            & 124     & --     \\
OPUS-EUbookshop                    & 75      & --     \\
\midrule
\textbf{Total}                     & \textbf{1,000,000} & \textbf{68} \\
\bottomrule
\end{tabular}
\caption{Number of Arabic sentence pairs contributed by each corpus for the Stage~1 SFT training dataset. Non-OPUS sources are aggregated as proprietary data.}
\label{tab:data-sizes-ar-s1}
\end{table}

\begin{table}[t]
\centering
\fitwidth{
\small
\begin{tabular}{lrr}
\toprule
\textbf{Corpus} & \textbf{Base} & \textbf{CONCAT} \\
\midrule
Proprietary data                   & 203,757 & 15     \\
OPUS-ELRC                          & 90,449  & --     \\
OPUS-DGT                           & 88,367  & 3      \\
OPUS-Europarl                      & 78,739  & 2      \\
OPUS-EUbookshop                    & 78,002  & 8      \\
OPUS-SciELO                        & 48,193  & --     \\
OPUS-GlobalVoices                  & 47,387  & 1      \\
OPUS-ELITR-ECA                     & 40,974  & 4      \\
OPUS-MultiUN                       & 39,451  & 3      \\
OPUS-JRC-Acquis                    & 38,903  & 4      \\
OPUS-Tatoeba                       & 35,037  & --     \\
es\_LA.OPUS-OpenSubtitles2018      & 34,928  & 12     \\
OPUS-UNPC                          & 34,514  & 4      \\
OPUS-TED2020                       & 31,017  & 6      \\
OPUS-NeuLab-TedTalks               & 25,279  & 3      \\
OPUS-EMEA                          & 20,498  & --     \\
OPUS-EuroPat                       & 16,707  & --     \\
OPUS-OpenSubtitles                 & 13,306  & 1      \\
OPUS-KDE4                          & 9,155   & 1      \\
OPUS-News-Commentary               & 7,048   & 1      \\
OPUS-ECB                           & 6,475   & --     \\
OPUS-Mozilla-I10n                  & 3,733   & --     \\
OPUS-Books                         & 3,424   & --     \\
OPUS-OpenOffice                    & 1,435   & --     \\
OPUS-MDN\_Web\_Docs                & 620     & --     \\
OPUS-Ubuntu                        & 539     & --     \\
OPUS-TED2013                       & 412     & --     \\
OPUS-tico-19                       & 406     & --     \\
OPUS-ECDC                          & 292     & --     \\
OPUS-UN                            & 290     & --     \\
OPUS-tldr-pages                    & 289     & --     \\
OPUS-EUconst                       & 234     & --     \\
OPUS-GNOME                         & 140     & --     \\
\midrule
\textbf{Total}                     & \textbf{1,000,000} & \textbf{68} \\
\bottomrule
\end{tabular}}
\caption{Number of Spanish sentence pairs contributed by each corpus for the Stage~1 SFT training dataset. Non-OPUS sources are aggregated as proprietary data.}
\label{tab:data-sizes-es-s1}
\end{table}

\paragraph{Stage 2.}
For bidirectional training runs on Llama 3.1 8B Instruct, we use only public data from OPUS (\Cref{tab:data-sizes-es} for Spanish, \Cref{tab:data-sizes-ar} for Arabic). During sampling, we explicitly minimized contributions from larger corpora, since these tend to consist of noisier, crawled data. For both \textit{Base} and \textit{CONCAT}, each language's 500,000 sentence pairs are split evenly across both translation directions to enable bidirectional training.
\begin{table}[t]
\centering
\small
\begin{tabular}{lrr}
\toprule
\textbf{Corpus} & \textbf{Base} & \textbf{CONCAT} \\
\midrule
OPUS-SciELO                    & 48,193 & --     \\
OPUS-GlobalVoices              & 47,387 & 18,945 \\
OPUS-ELITR-ECA                 & 40,974 & 63,626 \\
OPUS-JRC-Acquis                & 38,903 & 25,130 \\
OPUS-Tatoeba                   & 35,037 & --     \\
OPUS-ELRC                      & 33,584 & --     \\
OPUS-DGT                       & 32,866 & 38,634 \\
OPUS-TED2020                   & 31,017 & 58,439 \\
OPUS-Europarl                  & 29,490 & 31,800 \\
OPUS-EUbookshop                & 29,230 & 86,181 \\
OPUS-NeuLab-TedTalks           & 25,279 & 46,687 \\
OPUS-EMEA                      & 20,498 & --     \\
OPUS-MultiUN                   & 15,022 & 16,802 \\
es\_LA.OPUS-OpenSubtitles2018  & 13,318 & 51,304 \\
OPUS-UNPC                      & 13,160 & 29,092 \\
OPUS-KDE4                      & 9,155  & 9,325  \\
OPUS-News-Commentary           & 7,048  & 15,760 \\
OPUS-ECB                       & 6,475  & --     \\
OPUS-EuroPat                   & 6,421  & --     \\
OPUS-OpenSubtitles             & 5,129  & 6,615  \\
OPUS-Mozilla-I10n              & 3,733  & --     \\
OPUS-Books                     & 3,424  & --     \\
OPUS-OpenOffice                & 1,435  & --     \\
OPUS-MDN\_Web\_Docs            & 620    & 1,020  \\
OPUS-Ubuntu                    & 539    & 503    \\
OPUS-TED2013                   & 412    & --     \\
OPUS-tico-19                   & 406    & --     \\
OPUS-ECDC                      & 292    & --     \\
OPUS-UN                        & 290    & --     \\
OPUS-tldr-pages                & 289    & --     \\
OPUS-EUconst                   & 234    & --     \\
OPUS-GNOME                     & 140    & 137    \\
\midrule
\textbf{Total}                 & \textbf{500,000} & \textbf{500,000} \\
\bottomrule
\end{tabular}
\caption{Number of Spanish sentence pairs contributed by each OPUS corpus for the Stage~2 SFT training dataset.}
\label{tab:data-sizes-es}
\end{table}

\begin{table}[t]
\centering
\small
\begin{tabular}{lrr}
\toprule
\textbf{Corpus} & \textbf{Base} & \textbf{CONCAT} \\
\midrule
OPUS-MultiUN          & 157,691 & 144,588 \\
OPUS-UNPC             & 124,691 & 202,562 \\
OPUS-TED2020          & 61,929  & 44,055  \\
OPUS-OpenSubtitles    & 57,736  & 64,417  \\
OPUS-NeuLab-TedTalks  & 56,368  & 40,398  \\
OPUS-GlobalVoices     & 13,369  & 2,396   \\
OPUS-Tatoeba          & 9,211   & --      \\
OPUS-News-Commentary  & 4,659   & 463     \\
OPUS-TED2013          & 3,906   & --      \\
OPUS-ELRC             & 3,846   & --      \\
OPUS-KDE4             & 2,624   & 744     \\
OPUS-Mozilla-I10n     & 2,114   & --      \\
OPUS-tico-19          & 946     & --      \\
OPUS-Ubuntu           & 711     & 371     \\
OPUS-UN               & 124     & --      \\
OPUS-EUbookshop       & 75      & 6       \\
\midrule
\textbf{Total}        & \textbf{500,000} & \textbf{500,000} \\
\bottomrule
\end{tabular}
\caption{Number of Arabic sentence pairs contributed by each OPUS corpus for the Stage~2 SFT training dataset.}
\label{tab:data-sizes-ar}
\end{table}
 
\begin{table}[t]
\centering
\small
\begin{tabular}{lr}
\toprule
Category & \# Prompts \\
\midrule
General & 116,872 \\
Knowledge Recall & 104,982 \\
Math Reasoning & 334,252 \\
Coding & 142,275 \\
Safety \& Non-Compliance & 110,983 \\
Multilingual & 100,000 \\
\midrule
Instruction Following (removed) & 29,980 \\
\bottomrule
\end{tabular}
\caption{Composition of the Tülu 3 SFT-Mixture by category. We remove the synthetic instruction-following category to avoid overlap with the instructions evaluated in IFEval.}
\label{tab:TüluComposition}
\end{table}

\subsection{Training \& Evaluation Settings}
\subsubsection{Training}
\begin{itemize}
     \item \textbf{Epochs:} 1.0
    \item \textbf{Learning Rate (Non-PEFT):} 1e-6
    \item \textbf{Learning Rate (PEFT):} 1e-5
    \item \textbf{Optimizer:} AdamW (default weight decay; weight decay and AdamW only adapted for SPD regularization)
    \item \textbf{Warmup Ratio:} 0.003
    \item \textbf{Effective Batch Size:} 64
    \item \textbf{Max Sequence Length:} 1024 tokens
    \item \textbf{Chat Template}: Llama-3 Instruct format, with loss masked to assistant-turn content only.
\end{itemize}

\subsubsection{Evaluation Sampling Arguments}
\paragraph{General Capabilities.} We use the OLMES setup \citep{gu-etal-2025-olmes} for evaluating general capabilities, inheriting the task-specific sampling arguments of their implementations to ensure reproducibility. 

\paragraph{MT(-IF).} For MT-IF Evaluation, we use greedy decoding (temperature = 0) with a maximum output length of 512 tokens. 

\subsection{Computational Resources}
For the unidirectional training runs on Llama 3.2 1B Instruct, we used either one \texttt{A100} with 80GB or one GPU with 48GB. We report the average computation time across languages of the training techniques used for the bidirectional Llama 3.1 8B Instruct training in \Cref{tab:training-compute}. 

\begin{table}[t]
\centering
\small
\begin{tabular}{lcr}
\toprule
& & \textbf{Avg. Training} \\
\textbf{Training Technique} & \textbf{\# GPUs} & \textbf{Time (h)} \\
\midrule
Standard SFT                                  & 4x \texttt{A100} & 16.89\\
CC                                             & 6x \texttt{A100} & 19.26 \\
EAFT                                           & 6x \texttt{A100} & 17.25 \\
Freeze                                         & 4x \texttt{A100} & 14.15 \\
KL-Full                                        & 4x \texttt{H200} & 6.31 \\
LoRA                                           & 4x \texttt{A100} & 22.77 \\
SPD                                            & 6x \texttt{A100} & 18.04 \\
STM                                            & 6x \texttt{A100} & 21.23 \\
\midrule
Data Mixing & 4x \texttt{A100} & 53.50 \\
(T\"ulu full) \\
Data Mixing                        & 4x \texttt{A100} & 15.11 \\
(MT-IF 1600) \\
Data Mixing            & 4x \texttt{A100} & 15.54 \\
(T\"ulu + MT-IF; 2400)  \\
Data Mixing  & 4x \texttt{A100} & 16.84 \\
 (Single MT-IF \\
 \quad + T\"ulu; 1600) \\
\midrule
EWC                               & 2x \texttt{H200} & 14.79 \\
(10k T\"ulu) \\
EWC                            & 2x \texttt{H200} & 14.69 \\
 (MT-IF 1600)   \\
EWC                     & 2x \texttt{H200} & 14.99 \\
(T\"ulu + MT-IF; 2400) \\
\bottomrule
\end{tabular}
\caption{GPUs used and average training time (in hours) for each technique on the reported bidirectional Llama 3.1 8B Instruct training runs. }
\label{tab:training-compute}
\end{table}
\subsection{Prompt for MT Training \& Evaluation}
\noindent
\begin{quote}
\ttfamily \textbf{System:} You are a translation assistant who always responds with an \{target language\} translation. You should write the translation only without any formatting.
\\
\textbf{User:} Translate the given sentence into \{target language\}. \\
Sentence:\\
\{source sentence\}\\
\end{quote}

\subsection{Prompts for MT-IF Formality Control}
\label{app:prompts-formality}
\subsubsection{Training \& Evaluation}

\noindent
\begin{quote}
\ttfamily \textbf{System:} You are a translation assistant who always responds in an \{target language\} translation. You should only write the translation without any formatting and without any reasoning.
\\
\textbf{User:} Translate the following sentence into \{target language\} while maintaining [a|an] [formal|informal] tone.\\
Sentence:\\
\{source sentence\}\\
\end{quote}
\subsubsection{Different Training Prompts}

\noindent
\begin{quote}
\ttfamily \textbf{System:} Provide a \{target language\} translation only. Do not include explanations, formatting, or any text besides the translation itself.
\\
\textbf{User:} Translate the following text into \{target language\}. Use the [informal|formal] form of address.\\
Text to translate: \\
\{source sentence\}
\end{quote}
\subsection{Prompts for MT-IF Grammatical Gender Control}
\label{app:prompts-gender}

The prompts use the \texttt{male}/\texttt{female} labels of the MT-GenEval annotation and are reproduced as used in our experiments.

\subsubsection{Training \& Evaluation}
\noindent
\begin{quote}
\ttfamily \textbf{System:} You are a translation assistant who always responds with a \{target language\} translation. You should write the translation only without any formatting.
\\
\textbf{User:} Translate the given sentence into \{target language\}. In your translation the person referred to should be [male|female]. Translate all gendered words in the sentence accordingly, even if it uses opposite-gender pronouns. \\
Sentence: \\
\{source sentence\}
\end{quote}

\subsubsection{Different Training Prompts}
\noindent
\begin{quote}
\ttfamily \textbf{System:} Provide a \{target language\} translation only. Do not include explanations, formatting, or any text besides the translation itself.
\\
\textbf{User:} Translate the following text into \{target language\}. Ensure the translation represents the subject as [male|female]. Apply [male|female] gender forms to all applicable words, regardless of the original pronouns.\\
Text to translate: \\
\{source sentence\}
\end{quote}

\subsection{Prompts for MT-IF Length Control}
\subsubsection{Baseline}
\noindent
\begin{quote}
\ttfamily \textbf{System:} You are a translation assistant who always responds in an \{target language\} translation. You should only write the translation without any formatting and without any reasoning.
\\
\textbf{User:} Translate the following sentence into \{target language\}. \\
Sentence:\\
\{source sentence\}\\
\end{quote}
 \subsubsection{Shorter / Longer}
\noindent
\begin{quote}
\ttfamily \textbf{System:} You are a translation assistant who always responds in an \{target language\} translation. You should only write the translation without any formatting and without any reasoning.
\\
\textbf{User v1:} Translate the following sentence into \{target language \} while ensuring it has [more|fewer] characters than the original sentence.\\
Sentence:\\
\{source sentence\}\\
\textbf{User v2:} Translate the given sentence into \{target language\}. \\
Additional instruction to follow while translating: Produce a translated sentence with [more|fewer] than \{src chars\} characters. \\
Sentence:\\
\{source sentence\}\\
\textbf{User v3:} Translate the following sentence into \{target language\} ensuring it has [more|fewer] than \{src chars\} characters.\\
Sentence:\\
\{source sentence\}\\
\end{quote}

\FloatBarrier
\begin{table*}[t]
\refstepcounter{subsection}
\subsection*{\thesubsection\quad Stage 1: Best-Performing Configurations of Spanish \& Arabic Training}
\label{app:stage1BestConfig}
\centering
\small
\fitwidth{%
\renewcommand{\maxCOMET}{8.5}
\renewcommand{\maxBLEU}{13.8}
\renewcommand{\maxOLMES}{8.4}

\begin{tabular}{lllccccc}
\toprule
& & & \multicolumn{2}{c}{ar$\rightarrow$en} & \multicolumn{2}{c}{de$\rightarrow$en} & \\
\cmidrule(lr){4-5} \cmidrule(lr){6-7}
Anchoring & Method & Hyperparameters & COMET & BLEU & COMET & BLEU & General Avg \\
\midrule
\multicolumn{3}{l}{Llama 3.2 1B Instruct} & 75.8 & 18.1 & 82.9 & 27.6 & 44.9 \\
\midrule
\multicolumn{3}{l}{Standard SFT} & \deltacell{\textbf{84.3}}{8.5}{\maxCOMET} & \deltacell{\textbf{31.9}}{13.8}{\maxBLEU} & \deltacell{85.9}{3.0}{\maxCOMET} & \deltacell{34.9}{7.3}{\maxBLEU} & \deltacell{36.8}{-8.1}{\maxOLMES} \\
\midrule
\multirow{4}{*}{\shortstack[l]{Auxiliary\\Data}} & Data Mixing & Tülu (500k) & \deltacell{83.8}{8.0}{\maxCOMET} & \deltacell{31.2}{13.1}{\maxBLEU} & \deltacell{\underline{86.2}}{3.3}{\maxCOMET} & \deltacell{35.3}{7.7}{\maxBLEU} & \deltacell{40.8}{-4.1}{\maxOLMES} \\
& Merge & Trans=0.5, Tülu=0.5 & \deltacell{83.1}{7.3}{\maxCOMET} & \deltacell{29.7}{11.6}{\maxBLEU} & \deltacell{\textbf{86.4}}{3.6}{\maxCOMET} & \deltacell{\textbf{36.0}}{8.4}{\maxBLEU} & \deltacell{42.8}{-2.1}{\maxOLMES} \\
& EWC & $\lambda$=1e3, Tülu (10k) & \deltacell{83.3}{7.5}{\maxCOMET} & \deltacell{30.1}{12.0}{\maxBLEU} & \deltacell{86.1}{3.2}{\maxCOMET} & \deltacell{35.0}{7.4}{\maxBLEU} & \deltacell{\underline{44.0}}{-0.9}{\maxOLMES} \\
& EWC-DR & $\lambda$=1e4, Tülu (10k) & \deltacell{80.9}{5.1}{\maxCOMET} & \deltacell{24.8}{6.7}{\maxBLEU} & \deltacell{86.1}{3.2}{\maxCOMET} & \deltacell{33.8}{6.2}{\maxBLEU} & \deltacell{\textbf{44.1}}{-0.8}{\maxOLMES} \\
\midrule
\multirow{4}{*}{\shortstack[l]{Model\\Outputs}} & KL-Full & $\lambda$=0.03 & \deltacell{83.7}{7.9}{\maxCOMET} & \deltacell{30.2}{12.1}{\maxBLEU} & \deltacell{85.8}{2.9}{\maxCOMET} & \deltacell{33.7}{6.1}{\maxBLEU} & \deltacell{38.6}{-6.3}{\maxOLMES} \\
& STM & X=0.25 & \deltacell{81.9}{6.1}{\maxCOMET} & \deltacell{28.2}{10.1}{\maxBLEU} & \deltacell{85.8}{2.9}{\maxCOMET} & \deltacell{34.6}{7.0}{\maxBLEU} & \deltacell{41.3}{-3.6}{\maxOLMES} \\
& CC & X=0.2 & \deltacell{\textbf{84.3}}{8.5}{\maxCOMET} & \deltacell{\underline{31.8}}{13.7}{\maxBLEU} & \deltacell{85.9}{3.0}{\maxCOMET} & \deltacell{34.2}{6.6}{\maxBLEU} & \deltacell{37.9}{-7.0}{\maxOLMES} \\
& EAFT & K=20, Weight=1.5 & \deltacell{\underline{84.2}}{8.4}{\maxCOMET} & \deltacell{31.7}{13.6}{\maxBLEU} & \deltacell{85.9}{3.1}{\maxCOMET} & \deltacell{35.1}{7.5}{\maxBLEU} & \deltacell{37.9}{-7.0}{\maxOLMES} \\
\midrule
\multirow{3}{*}{\shortstack[l]{Base Model\\Parameters}} & Freeze & N=2 & \deltacell{83.8}{8.0}{\maxCOMET} & \deltacell{\underline{31.8}}{13.7}{\maxBLEU} & \deltacell{86.0}{3.1}{\maxCOMET} & \deltacell{35.2}{7.6}{\maxBLEU} & \deltacell{37.5}{-7.4}{\maxOLMES} \\
& SPD & $\lambda$=5 & \deltacell{84.0}{8.3}{\maxCOMET} & \deltacell{31.7}{13.6}{\maxBLEU} & \deltacell{\underline{86.2}}{3.3}{\maxCOMET} & \deltacell{35.2}{7.6}{\maxBLEU} & \deltacell{39.4}{-5.5}{\maxOLMES} \\
& LoRA & Rank=128, Alpha=64, lr=1e-5 & \deltacell{83.2}{7.4}{\maxCOMET} & \deltacell{29.7}{11.6}{\maxBLEU} & \deltacell{85.4}{2.5}{\maxCOMET} & \deltacell{34.0}{6.4}{\maxBLEU} & \deltacell{36.5}{-8.4}{\maxOLMES} \\
\bottomrule
\end{tabular}%
}
\caption{Best-performing configuration per method for Arabic, selected by the highest sum of Arabic COMET and general average score. Cell shading encodes the absolute difference from the base instruct model, normalized relative to the largest observed change within each metric group (across languages).}
\label{tab:bestArabicScores}
\end{table*}
 \begin{table*}[t]
\centering
\small
\fitwidth{%
\renewcommand{\maxCOMET}{4.9}
\renewcommand{\maxBLEU}{8.2}
\renewcommand{\maxOLMES}{7.1}

\begin{tabular}{lllccccc}
\toprule
& & & \multicolumn{2}{c}{es$\rightarrow$en} & \multicolumn{2}{c}{de$\rightarrow$en} & \\
\cmidrule(lr){4-5} \cmidrule(lr){6-7}
Anchoring & Method & Hyperparameters & COMET & BLEU & COMET & BLEU & General Avg \\
\midrule
\multicolumn{3}{l}{Llama 3.2 1B Instruct} & 81.3 & 20.3 & 82.9 & 27.6 & 44.9 \\
\midrule
\multicolumn{3}{l}{Standard SFT} & \deltacell{\textbf{86.2}}{4.9}{\maxCOMET} & \deltacell{\underline{28.4}}{8.1}{\maxBLEU} & \deltacell{84.3}{1.4}{\maxCOMET} & \deltacell{32.3}{4.7}{\maxBLEU} & \deltacell{37.9}{-7.0}{\maxOLMES} \\
\midrule
\multirow{4}{*}{\shortstack[l]{Auxiliary\\Data}} & Data Mixing & Tülu (500k) & \deltacell{\underline{86.1}}{4.8}{\maxCOMET} & \deltacell{28.1}{7.8}{\maxBLEU} & \deltacell{85.6}{2.7}{\maxCOMET} & \deltacell{33.8}{6.2}{\maxBLEU} & \deltacell{40.7}{-4.2}{\maxOLMES} \\
& Merge & Trans=0.5, Tülu=0.5 & \deltacell{85.7}{4.4}{\maxCOMET} & \deltacell{27.2}{6.9}{\maxBLEU} & \deltacell{\textbf{86.5}}{3.6}{\maxCOMET} & \deltacell{\textbf{35.8}}{8.2}{\maxBLEU} & \deltacell{42.2}{-2.7}{\maxOLMES} \\
& EWC & $\lambda$=1e3, Tülu (10k) & \deltacell{85.9}{4.6}{\maxCOMET} & \deltacell{27.5}{7.2}{\maxBLEU} & \deltacell{85.4}{2.5}{\maxCOMET} & \deltacell{34.0}{6.4}{\maxBLEU} & \deltacell{\underline{44.2}}{-0.7}{\maxOLMES} \\
& EWC-DR & $\lambda$=1e4, Tülu (10k) & \deltacell{85.0}{3.7}{\maxCOMET} & \deltacell{25.6}{5.3}{\maxBLEU} & \deltacell{\underline{85.9}}{3.0}{\maxCOMET} & \deltacell{34.3}{6.7}{\maxBLEU} & \deltacell{\textbf{44.7}}{-0.2}{\maxOLMES} \\
\midrule
\multirow{4}{*}{\shortstack[l]{Model\\Outputs}} & KL-Full & $\lambda$=0.003 & \deltacell{\underline{86.1}}{4.9}{\maxCOMET} & \deltacell{28.2}{7.9}{\maxBLEU} & \deltacell{84.6}{1.7}{\maxCOMET} & \deltacell{32.5}{4.9}{\maxBLEU} & \deltacell{38.3}{-6.6}{\maxOLMES} \\
& STM & X=0.25 & \deltacell{84.6}{3.3}{\maxCOMET} & \deltacell{24.7}{4.4}{\maxBLEU} & \deltacell{85.6}{2.7}{\maxCOMET} & \deltacell{34.2}{6.6}{\maxBLEU} & \deltacell{42.0}{-2.9}{\maxOLMES} \\
& CC & X=0.15 & \deltacell{\textbf{86.2}}{4.9}{\maxCOMET} & \deltacell{28.3}{8.0}{\maxBLEU} & \deltacell{84.2}{1.3}{\maxCOMET} & \deltacell{32.4}{4.8}{\maxBLEU} & \deltacell{38.5}{-6.4}{\maxOLMES} \\
& EAFT & K=20, Weight=1.0 & \deltacell{\textbf{86.2}}{4.9}{\maxCOMET} & \deltacell{28.2}{7.9}{\maxBLEU} & \deltacell{84.2}{1.4}{\maxCOMET} & \deltacell{31.9}{4.3}{\maxBLEU} & \deltacell{38.8}{-6.1}{\maxOLMES} \\
\midrule
\multirow{3}{*}{\shortstack[l]{Base Model\\Parameters}} & Freeze & N=1 & \deltacell{\underline{86.1}}{4.8}{\maxCOMET} & \deltacell{28.2}{7.9}{\maxBLEU} & \deltacell{85.7}{2.8}{\maxCOMET} & \deltacell{\underline{34.5}}{6.9}{\maxBLEU} & \deltacell{37.8}{-7.1}{\maxOLMES} \\
& SPD & $\lambda$=5 & \deltacell{\underline{86.1}}{4.8}{\maxCOMET} & \deltacell{28.0}{7.7}{\maxBLEU} & \deltacell{85.4}{2.5}{\maxCOMET} & \deltacell{34.1}{6.5}{\maxBLEU} & \deltacell{40.6}{-4.3}{\maxOLMES} \\
& LoRA & Rank=64, Alpha=64, lr=1e-5 & \deltacell{85.9}{4.6}{\maxCOMET} & \deltacell{27.8}{7.5}{\maxBLEU} & \deltacell{82.6}{-0.3}{\maxCOMET} & \deltacell{30.1}{2.5}{\maxBLEU} & \deltacell{38.0}{-7.0}{\maxOLMES} \\
\bottomrule
\end{tabular}%
}
\caption{Best-performing configuration per method for Spanish, selected by the highest sum of Spanish-language COMET and general average score. Cell shading encodes the absolute difference from the base instruct model, normalized relative to the largest observed change within each metric group (across languages).}
\label{tab:bestSpanishScores}
\end{table*}

\begin{table*}[t]
\centering
\small
\renewcommand{\maxAGIEval}{6.1}
\renewcommand{\maxHumanEval}{11.9}
\renewcommand{\maxHumanEvalPlus}{10.6}
\renewcommand{\maxDROP}{4.9}
\renewcommand{\maxGPQA}{2.5}
\renewcommand{\maxGSMEightK}{13.0}
\renewcommand{\maxIFEval}{17.6}
\renewcommand{\maxTruthfulQA}{5.7}
\renewcommand{\maxOLMES}{6.8}

\fitwidth{%
\begin{tabular}{lllccccccccc}
\toprule
Anchoring & Method & Hyperparameters & AGI-Eval & HumanEval & HumanEval+ & DROP & GPQA & GSM8K & IFEval & TruthfulQA & General Avg \\
\midrule
\multicolumn{3}{l}{Llama 3.2 1B Instruct} & 38.0 & 65.9 & 59.3 & 32.2 & 25.9 & 44.7 & 54.3 & 39.0 & 44.9 \\
\midrule
\multicolumn{3}{l}{Standard SFT} & \deltacell{32.6}{-5.3}{\maxAGIEval} & \deltacell{54.5}{-11.4}{\maxHumanEval} & \deltacell{49.4}{-9.9}{\maxHumanEvalPlus} & \deltacell{30.0}{-2.3}{\maxDROP} & \deltacell{26.8}{0.9}{\maxGPQA} & \deltacell{33.9}{-10.8}{\maxGSMEightK} & \deltacell{39.9}{-14.4}{\maxIFEval} & \deltacell{41.9}{3.0}{\maxTruthfulQA} & \deltacell{38.6}{-6.3}{\maxOLMES} \\
\midrule
\multirow{4}{*}{\shortstack[l]{Auxiliary\\Data}} & Data Mixing & Tülu (full) & \deltacell{\textbf{34.4}}{-3.6}{\maxAGIEval} & \deltacell{56.7}{-9.2}{\maxHumanEval} & \deltacell{53.8}{-5.6}{\maxHumanEvalPlus} & \deltacell{\textbf{33.0}}{0.7}{\maxDROP} & \deltacell{23.7}{-2.2}{\maxGPQA} & \deltacell{\textbf{42.8}}{-1.9}{\maxGSMEightK} & \deltacell{40.3}{-14.0}{\maxIFEval} & \deltacell{41.0}{2.0}{\maxTruthfulQA} & \deltacell{40.7}{-4.2}{\maxOLMES} \\
& Merge & Trans=0.75, Tülu=0.25 & \deltacell{\underline{34.2}}{-3.8}{\maxAGIEval} & \deltacell{59.1}{-6.7}{\maxHumanEval} & \deltacell{\textbf{55.7}}{-3.7}{\maxHumanEvalPlus} & \deltacell{\underline{30.9}}{-1.4}{\maxDROP} & \deltacell{24.6}{-1.3}{\maxGPQA} & \deltacell{\underline{40.7}}{-3.9}{\maxGSMEightK} & \deltacell{\underline{47.9}}{-6.5}{\maxIFEval} & \deltacell{41.6}{2.6}{\maxTruthfulQA} & \deltacell{\underline{41.8}}{-3.1}{\maxOLMES} \\
& EWC & $\lambda$=1e2, Tülu (75k) & \deltacell{33.7}{-4.3}{\maxAGIEval} & \deltacell{\textbf{61.0}}{-4.9}{\maxHumanEval} & \deltacell{\underline{55.1}}{-4.2}{\maxHumanEvalPlus} & \deltacell{30.1}{-2.1}{\maxDROP} & \deltacell{\textbf{28.3}}{2.5}{\maxGPQA} & \deltacell{37.4}{-7.3}{\maxGSMEightK} & \deltacell{\textbf{49.4}}{-5.0}{\maxIFEval} & \deltacell{\textbf{44.3}}{5.3}{\maxTruthfulQA} & \deltacell{\textbf{42.4}}{-2.5}{\maxOLMES} \\
& EWC-DR & $\lambda$=1e3, Tülu (10k) & \deltacell{32.6}{-5.4}{\maxAGIEval} & \deltacell{54.0}{-11.9}{\maxHumanEval} & \deltacell{49.2}{-10.1}{\maxHumanEvalPlus} & \deltacell{29.8}{-2.4}{\maxDROP} & \deltacell{27.0}{1.1}{\maxGPQA} & \deltacell{35.0}{-9.7}{\maxGSMEightK} & \deltacell{39.0}{-15.3}{\maxIFEval} & \deltacell{42.1}{3.1}{\maxTruthfulQA} & \deltacell{38.6}{-6.3}{\maxOLMES} \\
\midrule
\multirow{6}{*}{\shortstack[l]{Model\\Outputs}} & KL & $\lambda$=0.003 & \deltacell{32.5}{-5.5}{\maxAGIEval} & \deltacell{54.1}{-11.8}{\maxHumanEval} & \deltacell{48.7}{-10.6}{\maxHumanEvalPlus} & \deltacell{29.6}{-2.6}{\maxDROP} & \deltacell{27.0}{1.1}{\maxGPQA} & \deltacell{34.3}{-10.4}{\maxGSMEightK} & \deltacell{36.8}{-17.6}{\maxIFEval} & \deltacell{41.9}{2.9}{\maxTruthfulQA} & \deltacell{38.1}{-6.8}{\maxOLMES} \\
& KL-Distill & $\lambda$=0.05, $\tau=5$ & \deltacell{32.8}{-5.2}{\maxAGIEval} & \deltacell{\underline{59.6}}{-6.3}{\maxHumanEval} & \deltacell{52.1}{-7.2}{\maxHumanEvalPlus} & \deltacell{29.9}{-2.3}{\maxDROP} & \deltacell{24.1}{-1.8}{\maxGPQA} & \deltacell{33.3}{-11.4}{\maxGSMEightK} & \deltacell{39.9}{-14.4}{\maxIFEval} & \deltacell{\underline{44.2}}{5.2}{\maxTruthfulQA} & \deltacell{39.5}{-5.4}{\maxOLMES} \\
& KL-Full & $\lambda$=0.0025 & \deltacell{33.5}{-4.5}{\maxAGIEval} & \deltacell{57.4}{-8.5}{\maxHumanEval} & \deltacell{51.9}{-7.4}{\maxHumanEvalPlus} & \deltacell{29.4}{-2.8}{\maxDROP} & \deltacell{27.2}{1.3}{\maxGPQA} & \deltacell{33.7}{-11.0}{\maxGSMEightK} & \deltacell{42.0}{-12.4}{\maxIFEval} & \deltacell{42.2}{3.2}{\maxTruthfulQA} & \deltacell{39.7}{-5.3}{\maxOLMES} \\
& STM & X=0.15 & \deltacell{31.9}{-6.1}{\maxAGIEval} & \deltacell{56.1}{-9.7}{\maxHumanEval} & \deltacell{53.1}{-6.2}{\maxHumanEvalPlus} & \deltacell{30.0}{-2.2}{\maxDROP} & \deltacell{\underline{28.1}}{2.2}{\maxGPQA} & \deltacell{33.2}{-11.4}{\maxGSMEightK} & \deltacell{37.7}{-16.6}{\maxIFEval} & \deltacell{42.5}{3.5}{\maxTruthfulQA} & \deltacell{39.1}{-5.8}{\maxOLMES} \\
& CC & X=0.45 & \deltacell{32.5}{-5.5}{\maxAGIEval} & \deltacell{57.3}{-8.6}{\maxHumanEval} & \deltacell{53.5}{-5.9}{\maxHumanEvalPlus} & \deltacell{30.0}{-2.3}{\maxDROP} & \deltacell{26.8}{0.9}{\maxGPQA} & \deltacell{35.2}{-9.5}{\maxGSMEightK} & \deltacell{42.5}{-11.8}{\maxIFEval} & \deltacell{41.3}{2.3}{\maxTruthfulQA} & \deltacell{39.9}{-5.0}{\maxOLMES} \\
& EAFT & K=20, Weight=1.0 & \deltacell{32.3}{-5.7}{\maxAGIEval} & \deltacell{55.2}{-10.7}{\maxHumanEval} & \deltacell{50.3}{-9.1}{\maxHumanEvalPlus} & \deltacell{30.2}{-2.0}{\maxDROP} & \deltacell{27.5}{1.6}{\maxGPQA} & \deltacell{35.1}{-9.6}{\maxGSMEightK} & \deltacell{39.2}{-15.2}{\maxIFEval} & \deltacell{41.7}{2.7}{\maxTruthfulQA} & \deltacell{38.9}{-6.0}{\maxOLMES} \\
\midrule
\multirow{3}{*}{\shortstack[l]{Base Model\\Parameters}} & Freeze & N=1 & \deltacell{33.4}{-4.6}{\maxAGIEval} & \deltacell{58.0}{-7.9}{\maxHumanEval} & \deltacell{50.8}{-8.6}{\maxHumanEvalPlus} & \deltacell{28.9}{-3.3}{\maxDROP} & \deltacell{27.7}{1.8}{\maxGPQA} & \deltacell{33.9}{-10.8}{\maxGSMEightK} & \deltacell{43.3}{-11.1}{\maxIFEval} & \deltacell{42.0}{3.0}{\maxTruthfulQA} & \deltacell{39.7}{-5.2}{\maxOLMES} \\
& SPD & $\lambda$=2 & \deltacell{33.1}{-4.9}{\maxAGIEval} & \deltacell{58.2}{-7.7}{\maxHumanEval} & \deltacell{54.6}{-4.8}{\maxHumanEvalPlus} & \deltacell{30.5}{-1.7}{\maxDROP} & \deltacell{26.6}{0.7}{\maxGPQA} & \deltacell{35.6}{-9.0}{\maxGSMEightK} & \deltacell{42.1}{-12.2}{\maxIFEval} & \deltacell{43.1}{4.1}{\maxTruthfulQA} & \deltacell{40.5}{-4.4}{\maxOLMES} \\
& LoRA & Rank=64, lr=1e-5 & \deltacell{32.4}{-5.6}{\maxAGIEval} & \deltacell{59.8}{-6.1}{\maxHumanEval} & \deltacell{53.1}{-6.2}{\maxHumanEvalPlus} & \deltacell{27.3}{-4.9}{\maxDROP} & \deltacell{28.3}{2.5}{\maxGPQA} & \deltacell{31.6}{-13.0}{\maxGSMEightK} & \deltacell{39.6}{-14.8}{\maxIFEval} & \deltacell{44.7}{5.7}{\maxTruthfulQA} & \deltacell{39.6}{-5.3}{\maxOLMES} \\
\bottomrule
\end{tabular}%
}
\caption{Per-task scores on the general benchmarks for the best-performing configuration of each method for Amharic (same models as in \Cref{tab:bestAmharicScores}).}
\label{tab:olmesAmharicDetailed}
\end{table*} 

\FloatBarrier
\begin{table*}[t]
\refstepcounter{subsection}
\subsection*{\thesubsection\quad Per-Task General Benchmark Scores for All Training Runs}
\centering
\small
\renewcommand{\maxAGIEval}{12.5}
\renewcommand{\maxHumanEval}{12.8}
\renewcommand{\maxHumanEvalPlus}{10.9}
\renewcommand{\maxDROP}{5.4}
\renewcommand{\maxGPQA}{6.5}
\renewcommand{\maxGSMEightK}{13.8}
\renewcommand{\maxIFEval}{18.1}
\renewcommand{\maxTruthfulQA}{6.8}
\renewcommand{\maxOLMES}{8.4}

\fitwidth{%
\begin{tabular}{lllccccccccc}
\toprule
Anchoring & Method & Hyperparameters & AGI-Eval & HumanEval & HumanEval+ & DROP & GPQA & GSM8K & IFEval & TruthfulQA & General Avg \\
\midrule
\multicolumn{3}{l}{Llama 3.2 1B Instruct} & 38.0 & 65.9 & 59.3 & 32.2 & 25.9 & 44.7 & 54.3 & 39.0 & 44.9 \\
\midrule
\multicolumn{3}{l}{Standard SFT} & \deltacell{30.5}{-7.5}{\maxAGIEval} & \deltacell{53.1}{-12.8}{\maxHumanEval} & \deltacell{48.4}{-10.9}{\maxHumanEvalPlus} & \deltacell{28.3}{-4.0}{\maxDROP} & \deltacell{22.3}{-3.6}{\maxGPQA} & \deltacell{30.9}{-13.8}{\maxGSMEightK} & \deltacell{39.0}{-15.3}{\maxIFEval} & \deltacell{41.6}{2.6}{\maxTruthfulQA} & \deltacell{36.8}{-8.1}{\maxOLMES} \\
\midrule
\multirow{4}{*}{\shortstack[l]{Auxiliary\\Data}} & Data Mixing & Tülu (500k) & \deltacell{35.4}{-2.6}{\maxAGIEval} & \deltacell{59.3}{-6.5}{\maxHumanEval} & \deltacell{51.3}{-8.0}{\maxHumanEvalPlus} & \deltacell{\textbf{33.1}}{0.9}{\maxDROP} & \deltacell{23.2}{-2.7}{\maxGPQA} & \deltacell{\textbf{42.7}}{-2.0}{\maxGSMEightK} & \deltacell{41.4}{-12.9}{\maxIFEval} & \deltacell{40.2}{1.2}{\maxTruthfulQA} & \deltacell{40.8}{-4.1}{\maxOLMES} \\
& Merge & Trans=0.5, Tülu=0.5 & \deltacell{\underline{37.0}}{-1.0}{\maxAGIEval} & \deltacell{\underline{62.5}}{-3.4}{\maxHumanEval} & \deltacell{56.3}{-3.0}{\maxHumanEvalPlus} & \deltacell{32.2}{-0.1}{\maxDROP} & \deltacell{\underline{24.6}}{-1.3}{\maxGPQA} & \deltacell{41.7}{-3.0}{\maxGSMEightK} & \deltacell{46.6}{-7.8}{\maxIFEval} & \deltacell{41.6}{2.6}{\maxTruthfulQA} & \deltacell{42.8}{-2.1}{\maxOLMES} \\
& EWC & $\lambda$=1e3, Tülu (10k) & \deltacell{36.5}{-1.4}{\maxAGIEval} & \deltacell{\textbf{64.1}}{-1.8}{\maxHumanEval} & \deltacell{\textbf{58.9}}{-0.4}{\maxHumanEvalPlus} & \deltacell{31.6}{-0.6}{\maxDROP} & \deltacell{21.9}{-4.0}{\maxGPQA} & \deltacell{39.4}{-5.2}{\maxGSMEightK} & \deltacell{\textbf{55.1}}{0.7}{\maxIFEval} & \deltacell{\underline{44.2}}{5.2}{\maxTruthfulQA} & \deltacell{\underline{44.0}}{-0.9}{\maxOLMES} \\
& EWC-DR & $\lambda$=1e4, Tülu (10k) & \deltacell{\textbf{37.7}}{-0.3}{\maxAGIEval} & \deltacell{62.0}{-3.8}{\maxHumanEval} & \deltacell{\underline{58.2}}{-1.1}{\maxHumanEvalPlus} & \deltacell{\underline{32.4}}{0.2}{\maxDROP} & \deltacell{22.1}{-3.8}{\maxGPQA} & \deltacell{\underline{41.8}}{-2.8}{\maxGSMEightK} & \deltacell{\underline{52.5}}{-1.8}{\maxIFEval} & \deltacell{\textbf{45.8}}{6.8}{\maxTruthfulQA} & \deltacell{\textbf{44.1}}{-0.8}{\maxOLMES} \\
\midrule
\multirow{4}{*}{\shortstack[l]{Model\\Outputs}} & KL-Full & $\lambda$=0.03 & \deltacell{32.1}{-5.9}{\maxAGIEval} & \deltacell{54.3}{-11.6}{\maxHumanEval} & \deltacell{51.5}{-7.8}{\maxHumanEvalPlus} & \deltacell{28.5}{-3.7}{\maxDROP} & \deltacell{\textbf{25.2}}{-0.7}{\maxGPQA} & \deltacell{32.5}{-12.1}{\maxGSMEightK} & \deltacell{41.6}{-12.8}{\maxIFEval} & \deltacell{42.9}{3.9}{\maxTruthfulQA} & \deltacell{38.6}{-6.3}{\maxOLMES} \\
& STM & X=0.25 & \deltacell{35.2}{-2.8}{\maxAGIEval} & \deltacell{61.4}{-4.5}{\maxHumanEval} & \deltacell{54.1}{-5.2}{\maxHumanEvalPlus} & \deltacell{30.5}{-1.7}{\maxDROP} & \deltacell{20.8}{-5.1}{\maxGPQA} & \deltacell{38.6}{-6.1}{\maxGSMEightK} & \deltacell{47.9}{-6.5}{\maxIFEval} & \deltacell{42.2}{3.2}{\maxTruthfulQA} & \deltacell{41.3}{-3.6}{\maxOLMES} \\
& CC & X=0.2 & \deltacell{31.4}{-6.6}{\maxAGIEval} & \deltacell{55.7}{-10.1}{\maxHumanEval} & \deltacell{51.9}{-7.4}{\maxHumanEvalPlus} & \deltacell{28.5}{-3.7}{\maxDROP} & \deltacell{23.7}{-2.2}{\maxGPQA} & \deltacell{31.3}{-13.3}{\maxGSMEightK} & \deltacell{39.4}{-15.0}{\maxIFEval} & \deltacell{41.6}{2.6}{\maxTruthfulQA} & \deltacell{37.9}{-7.0}{\maxOLMES} \\
& EAFT & K=20, Weight=1.5 & \deltacell{31.7}{-6.3}{\maxAGIEval} & \deltacell{54.4}{-11.5}{\maxHumanEval} & \deltacell{50.5}{-8.8}{\maxHumanEvalPlus} & \deltacell{29.0}{-3.2}{\maxDROP} & \deltacell{23.0}{-2.9}{\maxGPQA} & \deltacell{32.8}{-11.8}{\maxGSMEightK} & \deltacell{39.9}{-14.4}{\maxIFEval} & \deltacell{41.9}{2.9}{\maxTruthfulQA} & \deltacell{37.9}{-7.0}{\maxOLMES} \\
\midrule
\multirow{3}{*}{\shortstack[l]{Base Model\\Parameters}} & Freeze & N=2 & \deltacell{31.3}{-6.7}{\maxAGIEval} & \deltacell{54.3}{-11.5}{\maxHumanEval} & \deltacell{50.3}{-9.0}{\maxHumanEvalPlus} & \deltacell{27.0}{-5.2}{\maxDROP} & \deltacell{22.1}{-3.8}{\maxGPQA} & \deltacell{34.3}{-10.3}{\maxGSMEightK} & \deltacell{38.4}{-15.9}{\maxIFEval} & \deltacell{42.5}{3.6}{\maxTruthfulQA} & \deltacell{37.5}{-7.4}{\maxOLMES} \\
& SPD & $\lambda$=5 & \deltacell{31.4}{-6.6}{\maxAGIEval} & \deltacell{59.4}{-6.5}{\maxHumanEval} & \deltacell{53.7}{-5.7}{\maxHumanEvalPlus} & \deltacell{29.5}{-2.7}{\maxDROP} & \deltacell{22.3}{-3.6}{\maxGPQA} & \deltacell{34.7}{-9.9}{\maxGSMEightK} & \deltacell{40.7}{-13.7}{\maxIFEval} & \deltacell{43.8}{4.8}{\maxTruthfulQA} & \deltacell{39.4}{-5.5}{\maxOLMES} \\
& LoRA & Rank=128, Alpha=64, lr=1e-5 & \deltacell{25.5}{-12.5}{\maxAGIEval} & \deltacell{56.8}{-9.1}{\maxHumanEval} & \deltacell{52.9}{-6.4}{\maxHumanEvalPlus} & \deltacell{26.8}{-5.4}{\maxDROP} & \deltacell{19.4}{-6.5}{\maxGPQA} & \deltacell{31.4}{-13.3}{\maxGSMEightK} & \deltacell{36.2}{-18.1}{\maxIFEval} & \deltacell{42.7}{3.7}{\maxTruthfulQA} & \deltacell{36.5}{-8.4}{\maxOLMES} \\
\bottomrule
\end{tabular}%
}
\caption{Per-task scores on the general benchmarks for the best-performing configuration of each method for Arabic (same models as in \Cref{tab:bestArabicScores}).}
\label{tab:olmes-arabic-detailed}
\end{table*}

 \begin{table*}[t]
\centering
\small
\renewcommand{\maxAGIEval}{7.3}
\renewcommand{\maxHumanEval}{11.3}
\renewcommand{\maxHumanEvalPlus}{11.9}
\renewcommand{\maxDROP}{4.1}
\renewcommand{\maxGPQA}{4.9}
\renewcommand{\maxGSMEightK}{13.5}
\renewcommand{\maxIFEval}{18.1}
\renewcommand{\maxTruthfulQA}{7.8}
\renewcommand{\maxOLMES}{7.1}
\fitwidth{%
\begin{tabular}{lllccccccccc}
\toprule
Anchoring & Method & Hyperparameters & AGI-Eval & HumanEval & HumanEval+ & DROP & GPQA & GSM8K & IFEval & TruthfulQA & General Avg \\
\midrule
\multicolumn{3}{l}{Llama 3.2 1B Instruct} & 38.0 & 65.9 & 59.3 & 32.2 & 25.9 & 44.7 & 54.3 & 39.0 & 44.9 \\
\midrule
\multicolumn{3}{l}{Standard SFT} & \deltacell{32.1}{-5.9}{\maxAGIEval} & \deltacell{56.4}{-9.5}{\maxHumanEval} & \deltacell{48.6}{-10.7}{\maxHumanEvalPlus} & \deltacell{29.4}{-2.8}{\maxDROP} & \deltacell{24.1}{-1.8}{\maxGPQA} & \deltacell{31.7}{-13.0}{\maxGSMEightK} & \deltacell{37.5}{-16.8}{\maxIFEval} & \deltacell{43.2}{4.2}{\maxTruthfulQA} & \deltacell{37.9}{-7.0}{\maxOLMES} \\
\midrule
\multirow{4}{*}{\shortstack[l]{Auxiliary\\Data}} & & Tülu (500k) & \deltacell{34.6}{-3.4}{\maxAGIEval} & \deltacell{57.4}{-8.5}{\maxHumanEval} & \deltacell{50.8}{-8.5}{\maxHumanEvalPlus} & \deltacell{\textbf{33.3}}{1.1}{\maxDROP} & \deltacell{25.4}{-0.5}{\maxGPQA} & \deltacell{\textbf{42.1}}{-2.6}{\maxGSMEightK} & \deltacell{41.6}{-12.7}{\maxIFEval} & \deltacell{40.1}{1.1}{\maxTruthfulQA} & \deltacell{40.7}{-4.2}{\maxOLMES} \\
& Merge & Trans=0.5, Tülu=0.5 & \deltacell{36.3}{-1.7}{\maxAGIEval} & \deltacell{59.7}{-6.2}{\maxHumanEval} & \deltacell{54.0}{-5.3}{\maxHumanEvalPlus} & \deltacell{32.2}{0}{\maxDROP} & \deltacell{24.6}{-1.3}{\maxGPQA} & \deltacell{41.2}{-3.5}{\maxGSMEightK} & \deltacell{47.7}{-6.6}{\maxIFEval} & \deltacell{42.2}{3.2}{\maxTruthfulQA} & \deltacell{42.2}{-2.7}{\maxOLMES} \\
& EWC & $\lambda$=1e3, Tülu (10k) & \deltacell{\underline{38.0}}{0}{\maxAGIEval} & \deltacell{61.9}{-4.0}{\maxHumanEval} & \deltacell{\textbf{58.6}}{-0.7}{\maxHumanEvalPlus} & \deltacell{31.4}{-0.8}{\maxDROP} & \deltacell{25.0}{-0.9}{\maxGPQA} & \deltacell{40.8}{-3.9}{\maxGSMEightK} & \deltacell{\textbf{53.8}}{-0.5}{\maxIFEval} & \deltacell{44.4}{5.4}{\maxTruthfulQA} & \deltacell{\underline{44.2}}{-0.7}{\maxOLMES} \\
& EWC-DR & $\lambda$=1e4, Tülu (10k) & \deltacell{\textbf{38.3}}{0.3}{\maxAGIEval} & \deltacell{\underline{63.5}}{-2.4}{\maxHumanEval} & \deltacell{58.3}{-1.0}{\maxHumanEvalPlus} & \deltacell{\underline{32.5}}{0.3}{\maxDROP} & \deltacell{\textbf{26.1}}{0.2}{\maxGPQA} & \deltacell{41.7}{-3.0}{\maxGSMEightK} & \deltacell{\underline{52.9}}{-1.4}{\maxIFEval} & \deltacell{44.5}{5.5}{\maxTruthfulQA} & \deltacell{\textbf{44.7}}{-0.2}{\maxOLMES} \\
\midrule
\multirow{4}{*}{\shortstack[l]{Model\\Outputs}} & KL-Full & $\lambda$=0.003 & \deltacell{32.5}{-5.5}{\maxAGIEval} & \deltacell{54.6}{-11.3}{\maxHumanEval} & \deltacell{50.6}{-8.7}{\maxHumanEvalPlus} & \deltacell{29.6}{-2.6}{\maxDROP} & \deltacell{\underline{25.7}}{-0.2}{\maxGPQA} & \deltacell{32.0}{-12.7}{\maxGSMEightK} & \deltacell{39.0}{-15.3}{\maxIFEval} & \deltacell{42.6}{3.6}{\maxTruthfulQA} & \deltacell{38.3}{-6.6}{\maxOLMES} \\
& STM & X=0.25 & \deltacell{37.1}{-0.9}{\maxAGIEval} & \deltacell{62.4}{-3.5}{\maxHumanEval} & \deltacell{55.1}{-4.2}{\maxHumanEvalPlus} & \deltacell{31.7}{-0.5}{\maxDROP} & \deltacell{21.0}{-4.9}{\maxGPQA} & \deltacell{39.9}{-4.8}{\maxGSMEightK} & \deltacell{49.0}{-5.3}{\maxIFEval} & \deltacell{39.6}{0.6}{\maxTruthfulQA} & \deltacell{42.0}{-2.9}{\maxOLMES} \\
& CC & X=0.15 & \deltacell{31.9}{-6.1}{\maxAGIEval} & \deltacell{58.4}{-7.5}{\maxHumanEval} & \deltacell{50.8}{-8.5}{\maxHumanEvalPlus} & \deltacell{29.7}{-2.5}{\maxDROP} & \deltacell{24.1}{-1.8}{\maxGPQA} & \deltacell{32.8}{-11.9}{\maxGSMEightK} & \deltacell{37.3}{-17.0}{\maxIFEval} & \deltacell{43.3}{4.3}{\maxTruthfulQA} & \deltacell{38.5}{-6.4}{\maxOLMES} \\
& EAFT & K=20, Weight=1.0 & \deltacell{32.5}{-5.5}{\maxAGIEval} & \deltacell{55.8}{-10.1}{\maxHumanEval} & \deltacell{53.7}{-5.6}{\maxHumanEvalPlus} & \deltacell{29.8}{-2.4}{\maxDROP} & \deltacell{24.6}{-1.3}{\maxGPQA} & \deltacell{32.1}{-12.6}{\maxGSMEightK} & \deltacell{39.2}{-15.1}{\maxIFEval} & \deltacell{42.8}{3.8}{\maxTruthfulQA} & \deltacell{38.8}{-6.1}{\maxOLMES} \\
\midrule
\multirow{3}{*}{\shortstack[l]{Base Model\\Parameters}} & Freeze & N=1 & \deltacell{32.4}{-5.6}{\maxAGIEval} & \deltacell{56.9}{-9.0}{\maxHumanEval} & \deltacell{47.4}{-11.9}{\maxHumanEvalPlus} & \deltacell{29.3}{-2.9}{\maxDROP} & \deltacell{24.8}{-1.1}{\maxGPQA} & \deltacell{32.3}{-12.4}{\maxGSMEightK} & \deltacell{36.2}{-18.1}{\maxIFEval} & \deltacell{43.3}{4.3}{\maxTruthfulQA} & \deltacell{37.8}{-7.1}{\maxOLMES} \\
& SPD & $\lambda$=5 & \deltacell{33.8}{-4.2}{\maxAGIEval} & \deltacell{58.4}{-7.5}{\maxHumanEval} & \deltacell{54.9}{-4.4}{\maxHumanEvalPlus} & \deltacell{30.4}{-1.8}{\maxDROP} & \deltacell{24.6}{-1.3}{\maxGPQA} & \deltacell{35.6}{-9.1}{\maxGSMEightK} & \deltacell{42.5}{-11.8}{\maxIFEval} & \deltacell{\underline{44.9}}{5.9}{\maxTruthfulQA} & \deltacell{40.6}{-4.3}{\maxOLMES} \\
& LoRA & Rank=64, Alpha=64, lr=1e-5 & \deltacell{30.7}{-7.3}{\maxAGIEval} & \deltacell{56.9}{-9.0}{\maxHumanEval} & \deltacell{48.1}{-11.2}{\maxHumanEvalPlus} & \deltacell{28.1}{-4.1}{\maxDROP} & \deltacell{24.1}{-1.8}{\maxGPQA} & \deltacell{31.2}{-13.5}{\maxGSMEightK} & \deltacell{37.7}{-16.6}{\maxIFEval} & \deltacell{\textbf{46.8}}{7.8}{\maxTruthfulQA} & \deltacell{38.0}{-6.9}{\maxOLMES} \\
\bottomrule
\end{tabular}%
}
\caption{Per-task scores on the general benchmarks (as percentages) for the best-performing configuration of each method for Spanish (same model selection as in \Cref{tab:bestSpanishScores}).}
\label{tab:olmesSpanishDetailed}
\end{table*}

\begin{table*}[t]
\centering
\small
\renewcommand{\maxAGIEval}{16.2}
\renewcommand{\maxHumanEval}{3.9}
\renewcommand{\maxHumanEvalPlus}{6.1}
\renewcommand{\maxDROP}{36.4}
\renewcommand{\maxGPQA}{2.5}
\renewcommand{\maxGSMEightK}{12}
\renewcommand{\maxIFEval}{29.7}
\renewcommand{\maxTruthfulQA}{10.9}
\renewcommand{\maxOLMES}{13.8}

\fitwidth{%
\begin{tabular}{lllccccccccc}
\toprule
Anchoring & Method & Hyperparameters & AGI-Eval & HumanEval & HumanEval+ & DROP & GPQA & GSM8K & IFEval & TruthfulQA & General Avg \\
\midrule
\multicolumn{3}{l}{Llama 3.1 8B Instruct} & 64.5 & 85.3 & 79.8 & 61.5 & 26.8 & 83.2 & 78.7 & 55.0 & 66.8 \\
\midrule
\multicolumn{3}{l}{Standard SFT} & \deltacell{52.5}{-12.0}{\maxAGIEval} & \deltacell{82.1}{-3.2}{\maxHumanEval} & \deltacell{76.4}{-3.4}{\maxHumanEvalPlus} & \deltacell{36.4}{-25.1}{\maxDROP} & \deltacell{25.9}{-0.9}{\maxGPQA} & \deltacell{75.3}{-7.9}{\maxGSMEightK} & \deltacell{53.0}{-25.7}{\maxIFEval} & \deltacell{44.7}{-10.3}{\maxTruthfulQA} & \deltacell{55.8}{-11.0}{\maxOLMES} \\
\midrule
\multirow{12}{*}{\shortstack[l]{Auxiliary\\Data}} & \multirow{8}{*}{\shortstack[l]{Data\\Mixing}} & Tülu (full) & \deltacell{57.9}{-6.6}{\maxAGIEval} & \deltacell{87.1}{1.8}{\maxHumanEval} & \deltacell{78.9}{-0.9}{\maxHumanEvalPlus} & \deltacell{45.7}{-15.8}{\maxDROP} & \deltacell{27.7}{0.9}{\maxGPQA} & \deltacell{76.6}{-6.6}{\maxGSMEightK} & \deltacell{60.3}{-18.4}{\maxIFEval} & \deltacell{46.2}{-8.8}{\maxTruthfulQA} & \deltacell{60.0}{-6.8}{\maxOLMES} \\
& & Tülu (800) + MT-IF (1600) & \deltacell{58.3}{-6.2}{\maxAGIEval} & \deltacell{82.1}{-3.2}{\maxHumanEval} & \deltacell{76.2}{-3.6}{\maxHumanEvalPlus} & \deltacell{61.1}{-0.4}{\maxDROP} & \deltacell{27.0}{0.2}{\maxGPQA} & \deltacell{76.3}{-6.9}{\maxGSMEightK} & \deltacell{55.1}{-23.6}{\maxIFEval} & \deltacell{48.5}{-6.5}{\maxTruthfulQA} & \deltacell{60.6}{-6.2}{\maxOLMES} \\
& & \quad + diff MT-IF prompt & \deltacell{59.1}{-5.4}{\maxAGIEval} & \deltacell{80.8}{-4.5}{\maxHumanEval} & \deltacell{75.3}{-4.5}{\maxHumanEvalPlus} & \deltacell{62.5}{1.0}{\maxDROP} & \deltacell{27.5}{0.7}{\maxGPQA} & \deltacell{77.3}{-5.9}{\maxGSMEightK} & \deltacell{56.9}{-21.8}{\maxIFEval} & \deltacell{47.6}{-7.4}{\maxTruthfulQA} & \deltacell{60.9}{-5.9}{\maxOLMES} \\
& & MT-IF (1600) & \deltacell{52.7}{-11.8}{\maxAGIEval} & \deltacell{83.5}{-1.8}{\maxHumanEval} & \deltacell{77.0}{-2.8}{\maxHumanEvalPlus} & \deltacell{34.7}{-26.8}{\maxDROP} & \deltacell{28.3}{1.5}{\maxGPQA} & \deltacell{74.3}{-8.9}{\maxGSMEightK} & \deltacell{56.4}{-22.3}{\maxIFEval} & \deltacell{45.5}{-9.5}{\maxTruthfulQA} & \deltacell{56.5}{-10.3}{\maxOLMES} \\
& & Tülu (800) + Gender (800) & \deltacell{59.5}{-5.0}{\maxAGIEval} & \deltacell{81.6}{-3.7}{\maxHumanEval} & \deltacell{77.2}{-2.6}{\maxHumanEvalPlus} & \deltacell{62.0}{0.5}{\maxDROP} & \deltacell{28.3}{1.5}{\maxGPQA} & \deltacell{77.3}{-5.9}{\maxGSMEightK} & \deltacell{60.3}{-18.4}{\maxIFEval} & \deltacell{47.9}{-7.1}{\maxTruthfulQA} & \deltacell{61.7}{-5.1}{\maxOLMES} \\
& & \quad + diff Gender prompt & \deltacell{58.7}{-5.8}{\maxAGIEval} & \deltacell{83.7}{-1.6}{\maxHumanEval} & \deltacell{76.1}{-3.7}{\maxHumanEvalPlus} & \deltacell{63.1}{1.6}{\maxDROP} & \deltacell{27.0}{0.2}{\maxGPQA} & \deltacell{78.0}{-5.2}{\maxGSMEightK} & \deltacell{61.6}{-17.1}{\maxIFEval} & \deltacell{47.6}{-7.4}{\maxTruthfulQA} & \deltacell{62.0}{-4.8}{\maxOLMES} \\
& & Tülu (800) + Formality (800) & \deltacell{59.2}{-5.3}{\maxAGIEval} & \deltacell{83.8}{-1.5}{\maxHumanEval} & \deltacell{74.6}{-5.2}{\maxHumanEvalPlus} & \deltacell{62.3}{0.8}{\maxDROP} & \deltacell{27.7}{0.9}{\maxGPQA} & \deltacell{77.9}{-5.3}{\maxGSMEightK} & \deltacell{59.9}{-18.8}{\maxIFEval} & \deltacell{47.2}{-7.8}{\maxTruthfulQA} & \deltacell{61.6}{-5.2}{\maxOLMES} \\
& & \quad + diff Formality prompt & \deltacell{59.1}{-5.4}{\maxAGIEval} & \deltacell{83.8}{-1.5}{\maxHumanEval} & \deltacell{76.8}{-3.0}{\maxHumanEvalPlus} & \deltacell{63.0}{1.5}{\maxDROP} & \deltacell{27.5}{0.7}{\maxGPQA} & \deltacell{78.0}{-5.2}{\maxGSMEightK} & \deltacell{61.2}{-17.5}{\maxIFEval} & \deltacell{46.7}{-8.3}{\maxTruthfulQA} & \deltacell{62.0}{-4.8}{\maxOLMES} \\
\cmidrule(lr){2-12}
& Merge & Trans=0.5, Tülu+MT-IF=0.5 & \deltacell{58.1}{-6.4}{\maxAGIEval} & \deltacell{86.9}{1.6}{\maxHumanEval} & \deltacell{79.7}{-0.1}{\maxHumanEvalPlus} & \deltacell{47.1}{-14.4}{\maxDROP} & \deltacell{26.3}{-0.5}{\maxGPQA} & \deltacell{79.5}{-3.7}{\maxGSMEightK} & \deltacell{68.2}{-10.5}{\maxIFEval} & \deltacell{46.2}{-8.8}{\maxTruthfulQA} & \deltacell{61.5}{-5.3}{\maxOLMES} \\
\cmidrule(lr){2-12}
& \multirow{3}{*}{EWC} & $\lambda$=1e3, Tülu (10k) & \deltacell{60.3}{-4.2}{\maxAGIEval} & \deltacell{85.3}{0.0}{\maxHumanEval} & \deltacell{80.2}{0.4}{\maxHumanEvalPlus} & \deltacell{62.1}{0.6}{\maxDROP} & \deltacell{28.3}{1.5}{\maxGPQA} & \deltacell{82.5}{-0.7}{\maxGSMEightK} & \deltacell{72.1}{-6.6}{\maxIFEval} & \deltacell{50.0}{-5.0}{\maxTruthfulQA} & \deltacell{65.1}{-1.7}{\maxOLMES} \\
& & $\lambda$=1e3, Tülu (800) + MT-IF (1600) & \deltacell{57.3}{-7.2}{\maxAGIEval} & \deltacell{85.4}{0.1}{\maxHumanEval} & \deltacell{78.5}{-1.3}{\maxHumanEvalPlus} & \deltacell{64.2}{2.7}{\maxDROP} & \deltacell{25.2}{-1.6}{\maxGPQA} & \deltacell{77.7}{-5.5}{\maxGSMEightK} & \deltacell{69.7}{-9.0}{\maxIFEval} & \deltacell{46.5}{-8.5}{\maxTruthfulQA} & \deltacell{63.1}{-3.7}{\maxOLMES} \\
& & $\lambda$=1e3, MT-IF (1600) & \deltacell{56.8}{-7.7}{\maxAGIEval} & \deltacell{84.9}{-0.4}{\maxHumanEval} & \deltacell{78.7}{-1.1}{\maxHumanEvalPlus} & \deltacell{34.0}{-27.5}{\maxDROP} & \deltacell{25.7}{-1.1}{\maxGPQA} & \deltacell{76.9}{-6.3}{\maxGSMEightK} & \deltacell{60.6}{-18.1}{\maxIFEval} & \deltacell{47.3}{-7.7}{\maxTruthfulQA} & \deltacell{58.1}{-8.7}{\maxOLMES} \\
\midrule
\multirow{4}{*}{\shortstack[l]{Model\\Outputs}} & KL-Full & $\lambda$=0.05 & \deltacell{53.5}{-11.0}{\maxAGIEval} & \deltacell{83.6}{-1.7}{\maxHumanEval} & \deltacell{76.6}{-3.2}{\maxHumanEvalPlus} & \deltacell{46.8}{-14.7}{\maxDROP} & \deltacell{25.0}{-1.8}{\maxGPQA} & \deltacell{74.4}{-8.8}{\maxGSMEightK} & \deltacell{56.0}{-22.7}{\maxIFEval} & \deltacell{46.4}{-8.6}{\maxTruthfulQA} & \deltacell{57.8}{-9.0}{\maxOLMES} \\
& STM & X=0.25 & \deltacell{50.3}{-14.2}{\maxAGIEval} & \deltacell{82.1}{-3.2}{\maxHumanEval} & \deltacell{73.7}{-6.1}{\maxHumanEvalPlus} & \deltacell{25.1}{-36.4}{\maxDROP} & \deltacell{24.3}{-2.5}{\maxGPQA} & \deltacell{71.2}{-12}{\maxGSMEightK} & \deltacell{52.9}{-25.8}{\maxIFEval} & \deltacell{44.1}{-10.9}{\maxTruthfulQA} & \deltacell{53.0}{-13.8}{\maxOLMES} \\
& CC & X=0.25 & \deltacell{49.7}{-14.8}{\maxAGIEval} & \deltacell{83.2}{-2.1}{\maxHumanEval} & \deltacell{75.9}{-3.9}{\maxHumanEvalPlus} & \deltacell{30.2}{-31.3}{\maxDROP} & \deltacell{26.8}{0.0}{\maxGPQA} & \deltacell{73.6}{-9.6}{\maxGSMEightK} & \deltacell{55.1}{-23.6}{\maxIFEval} & \deltacell{45.4}{-9.6}{\maxTruthfulQA} & \deltacell{55.0}{-11.8}{\maxOLMES} \\
& EAFT & K=20, Weight=1.0 & \deltacell{52.3}{-12.2}{\maxAGIEval} & \deltacell{84.1}{-1.2}{\maxHumanEval} & \deltacell{75.4}{-4.4}{\maxHumanEvalPlus} & \deltacell{46.5}{-15.0}{\maxDROP} & \deltacell{25.0}{-1.8}{\maxGPQA} & \deltacell{76.0}{-9.2}{\maxGSMEightK} & \deltacell{58.0}{-20.7}{\maxIFEval} & \deltacell{45.5}{-4.5}{\maxTruthfulQA} & \deltacell{57.8}{-9.0}{\maxOLMES} \\
\midrule
\multirow{3}{*}{\shortstack[l]{Base Model\\Parameters}} & Freeze & N=4 & \deltacell{49.4}{-15.1}{\maxAGIEval} & \deltacell{82.9}{-2.4}{\maxHumanEval} & \deltacell{74.0}{-5.8}{\maxHumanEvalPlus} & \deltacell{31.2}{-30.3}{\maxDROP} & \deltacell{26.8}{0.0}{\maxGPQA} & \deltacell{75.4}{-7.8}{\maxGSMEightK} & \deltacell{49.0}{-29.7}{\maxIFEval} & \deltacell{45.5}{-9.5}{\maxTruthfulQA} & \deltacell{54.3}{-12.5}{\maxOLMES} \\
& SPD & $\lambda$=5 & \deltacell{61.4}{-3.1}{\maxAGIEval} & \deltacell{85.2}{-0.1}{\maxHumanEval} & \deltacell{79.4}{-0.4}{\maxHumanEvalPlus} & \deltacell{27.1}{-34.4}{\maxDROP} & \deltacell{24.6}{-2.2}{\maxGPQA} & \deltacell{81.7}{-1.5}{\maxGSMEightK} & \deltacell{69.1}{-9.6}{\maxIFEval} & \deltacell{53.2}{-1.8}{\maxTruthfulQA} & \deltacell{60.2}{-6.6}{\maxOLMES} \\
& LoRA & Rank=32, Alpha=64 & \deltacell{48.3}{-16.2}{\maxAGIEval} & \deltacell{81.4}{-3.9}{\maxHumanEval} & \deltacell{78.5}{-1.3}{\maxHumanEvalPlus} & \deltacell{65.8}{4.3}{\maxDROP} & \deltacell{24.6}{-2.2}{\maxGPQA} & \deltacell{78.5}{-4.7}{\maxGSMEightK} & \deltacell{65.2}{-13.5}{\maxIFEval} & \deltacell{47.9}{-7.1}{\maxTruthfulQA} & \deltacell{61.3}{-5.5}{\maxOLMES} \\

\bottomrule
\end{tabular}%
}
\caption{Per-task scores on the general benchmarks for bidirectional Spanish training on the Llama 3.1 8B Instruct model.}
\label{tab:bestSpanishScoresOLMESBreakdown8B}
\end{table*}

\begin{table*}[t]
\centering
\small
\renewcommand{\maxAGIEval}{16.6}
\renewcommand{\maxHumanEval}{3.1}
\renewcommand{\maxHumanEvalPlus}{4.7}
\renewcommand{\maxDROP}{26.2}
\renewcommand{\maxGPQA}{3.8}
\renewcommand{\maxGSMEightK}{8.7}
\renewcommand{\maxIFEval}{28.2}
\renewcommand{\maxTruthfulQA}{9.4}
\renewcommand{\maxOLMES}{10.8}
\fitwidth{%
\begin{tabular}{lllccccccccc}
\toprule
Anchoring & Method & Hyperparameters & AGI-Eval & HumanEval & HumanEval+ & DROP & GPQA & GSM8K & IFEval & TruthfulQA & General Avg \\
\midrule
\multicolumn{3}{l}{Llama 3.1 8B Instruct} & 64.5 & 85.3 & 79.8 & 61.5 & 26.8 & 83.2 & 78.7 & 55.0 & 66.8 \\
\midrule
\multicolumn{3}{l}{Standard SFT} & \deltacell{53.4}{-11.1}{\maxAGIEval} & \deltacell{85.2}{-0.1}{\maxHumanEval} & \deltacell{77.4}{-2.4}{\maxHumanEvalPlus} & \deltacell{48.7}{-12.8}{\maxDROP} & \deltacell{25.0}{-1.8}{\maxGPQA} & \deltacell{75.4}{-7.8}{\maxGSMEightK} & \deltacell{56.9}{-21.8}{\maxIFEval} & \deltacell{45.6}{-9.4}{\maxTruthfulQA} & \deltacell{58.5}{-8.3}{\maxOLMES} \\
\midrule
\multirow{9}{*}{\shortstack[l]{Auxiliary\\Data}} & \multirow{6}{*}{\shortstack[l]{Data\\Mixing}} & Tülu (full) & \deltacell{57.0}{-7.5}{\maxAGIEval} & \deltacell{84.0}{-1.3}{\maxHumanEval} & \deltacell{80.1}{0.3}{\maxHumanEvalPlus} & \deltacell{46.0}{-15.5}{\maxDROP} & \deltacell{29.9}{3.1}{\maxGPQA} & \deltacell{77.5}{-5.7}{\maxGSMEightK} & \deltacell{61.7}{-17.0}{\maxIFEval} & \deltacell{46.7}{-8.3}{\maxTruthfulQA} & \deltacell{60.4}{-6.4}{\maxOLMES} \\
& & Tülu (800) + MT-IF (1600) & \deltacell{58.9}{-5.6}{\maxAGIEval} & \deltacell{84.7}{-0.6}{\maxHumanEval} & \deltacell{77.9}{-1.9}{\maxHumanEvalPlus} & \deltacell{65.6}{4.1}{\maxDROP} & \deltacell{25.4}{-1.4}{\maxGPQA} & \deltacell{76.4}{-6.8}{\maxGSMEightK} & \deltacell{59.3}{-19.4}{\maxIFEval} & \deltacell{46.8}{-8.2}{\maxTruthfulQA} & \deltacell{61.9}{-4.9}{\maxOLMES} \\
& & \quad + diff MT-IF prompt & \deltacell{58.8}{-5.7}{\maxAGIEval} & \deltacell{83.2}{-2.1}{\maxHumanEval} & \deltacell{76.1}{-3.7}{\maxHumanEvalPlus} & \deltacell{65.6}{4.1}{\maxDROP} & \deltacell{28.1}{1.3}{\maxGPQA} & \deltacell{76.7}{-6.5}{\maxGSMEightK} & \deltacell{58.0}{-20.7}{\maxIFEval} & \deltacell{47.0}{-8.0}{\maxTruthfulQA} & \deltacell{61.7}{-5.1}{\maxOLMES} \\
& & MT-IF (1600) & \deltacell{54.0}{-10.5}{\maxAGIEval} & \deltacell{84.9}{-0.4}{\maxHumanEval} & \deltacell{78.5}{-1.3}{\maxHumanEvalPlus} & \deltacell{45.3}{-16.2}{\maxDROP} & \deltacell{30.6}{3.8}{\maxGPQA} & \deltacell{76.7}{-6.5}{\maxGSMEightK} & \deltacell{61.4}{-17.3}{\maxIFEval} & \deltacell{46.7}{-8.3}{\maxTruthfulQA} & \deltacell{59.8}{-7.0}{\maxOLMES} \\
& & Tülu (800) + Gender (800) & \deltacell{58.2}{-6.3}{\maxAGIEval} & \deltacell{83.8}{-1.5}{\maxHumanEval} & \deltacell{76.2}{-3.6}{\maxHumanEvalPlus} & \deltacell{65.1}{3.6}{\maxDROP} & \deltacell{26.3}{-0.5}{\maxGPQA} & \deltacell{78.2}{-5.0}{\maxGSMEightK} & \deltacell{59.5}{-19.2}{\maxIFEval} & \deltacell{48.0}{-7.0}{\maxTruthfulQA} & \deltacell{61.9}{-4.9}{\maxOLMES} \\
& & Tülu (800) + Formality (800) & \deltacell{58.0}{-6.5}{\maxAGIEval} & \deltacell{83.0}{-2.3}{\maxHumanEval} & \deltacell{77.6}{-2.2}{\maxHumanEvalPlus} & \deltacell{65.2}{3.7}{\maxDROP} & \deltacell{28.3}{1.5}{\maxGPQA} & \deltacell{78.1}{-5.1}{\maxGSMEightK} & \deltacell{61.4}{-17.3}{\maxIFEval} & \deltacell{48.3}{-6.7}{\maxTruthfulQA} & \deltacell{62.5}{-4.3}{\maxOLMES} \\
\cmidrule(lr){2-12}
& Merge & Trans=0.5, Tülu+MT-IF=0.5 & \deltacell{56.6}{-7.9}{\maxAGIEval} & \deltacell{88.4}{3.1}{\maxHumanEval} & \deltacell{78.8}{-1.0}{\maxHumanEvalPlus} & \deltacell{52.8}{-8.7}{\maxDROP} & \deltacell{28.1}{1.3}{\maxGPQA} & \deltacell{78.9}{-4.3}{\maxGSMEightK} & \deltacell{68.8}{-9.9}{\maxIFEval} & \deltacell{47.2}{-7.8}{\maxTruthfulQA} & \deltacell{62.5}{-4.3}{\maxOLMES} \\
\cmidrule(lr){2-12}
& \multirow{3}{*}{EWC} & $\lambda$=1e3, Tülu (10k) & \deltacell{60.7}{-3.8}{\maxAGIEval} & \deltacell{87.7}{2.4}{\maxHumanEval} & \deltacell{78.4}{-1.4}{\maxHumanEvalPlus} & \deltacell{64.0}{2.5}{\maxDROP} & \deltacell{28.6}{1.8}{\maxGPQA} & \deltacell{82.5}{-0.7}{\maxGSMEightK} & \deltacell{76.7}{-2.0}{\maxIFEval} & \deltacell{50.4}{-4.6}{\maxTruthfulQA} & \deltacell{66.1}{-0.7}{\maxOLMES} \\
& & $\lambda$=1e3, Tülu (800) + MT-IF (1600) & \deltacell{57.7}{-6.8}{\maxAGIEval} & \deltacell{84.6}{-0.7}{\maxHumanEval} & \deltacell{76.0}{-3.8}{\maxHumanEvalPlus} & \deltacell{45.4}{-16.1}{\maxDROP} & \deltacell{27.2}{0.4}{\maxGPQA} & \deltacell{77.7}{-5.5}{\maxGSMEightK} & \deltacell{66.5}{-12.2}{\maxIFEval} & \deltacell{47.6}{-7.4}{\maxTruthfulQA} & \deltacell{60.1}{-6.7}{\maxOLMES} \\
& & $\lambda$=1e3, MT-IF (1600) & \deltacell{55.9}{-8.6}{\maxAGIEval} & \deltacell{85.0}{-0.3}{\maxHumanEval} & \deltacell{77.0}{-2.8}{\maxHumanEvalPlus} & \deltacell{66.0}{4.5}{\maxDROP} & \deltacell{24.3}{-2.5}{\maxGPQA} & \deltacell{78.9}{-4.3}{\maxGSMEightK} & \deltacell{69.9}{-8.8}{\maxIFEval} & \deltacell{47.9}{-7.1}{\maxTruthfulQA} & \deltacell{63.3}{-3.5}{\maxOLMES} \\

\midrule
\multirow{4}{*}{\shortstack[l]{Model\\Outputs}} & KL-Full & $\lambda$=0.05 & \deltacell{53.8}{-10.7}{\maxAGIEval} & \deltacell{85.2}{-0.1}{\maxHumanEval} & \deltacell{77.3}{-2.5}{\maxHumanEvalPlus} & \deltacell{57.6}{-3.9}{\maxDROP} & \deltacell{27.9}{1.1}{\maxGPQA} & \deltacell{74.5}{-8.7}{\maxGSMEightK} & \deltacell{59.2}{-19.5}{\maxIFEval} & \deltacell{47.9}{-7.1}{\maxTruthfulQA} & \deltacell{60.4}{-6.4}{\maxOLMES} \\ 
& STM & X=0.25 & \deltacell{51.2}{-13.3}{\maxAGIEval} & \deltacell{85.0}{-0.3}{\maxHumanEval} & \deltacell{76.4}{-3.4}{\maxHumanEvalPlus} & \deltacell{35.3}{-26.2}{\maxDROP} & \deltacell{24.6}{-2.2}{\maxGPQA} & \deltacell{75.1}{-8.1}{\maxGSMEightK} & \deltacell{54.2}{-24.5}{\maxIFEval} & \deltacell{46.1}{-8.9}{\maxTruthfulQA} & \deltacell{56.0}{-10.8}{\maxOLMES} \\
& CC & X=0.25 & \deltacell{52.2}{-12.3}{\maxAGIEval} & \deltacell{85.9}{0.6}{\maxHumanEval} & \deltacell{78.2}{-1.6}{\maxHumanEvalPlus} & \deltacell{41.4}{-20.1}{\maxDROP} & \deltacell{25.7}{-1.1}{\maxGPQA} & \deltacell{75.0}{-8.2}{\maxGSMEightK} & \deltacell{55.1}{-23.6}{\maxIFEval} & \deltacell{46.8}{-8.2}{\maxTruthfulQA} & \deltacell{57.5}{-9.3}{\maxOLMES} \\
& EAFT & K=20, Weight=1.0 & \deltacell{52.3}{-12.2}{\maxAGIEval} & \deltacell{84.1}{-1.2}{\maxHumanEval} & \deltacell{80.8}{1.0}{\maxHumanEvalPlus} & \deltacell{55.0}{-6.5}{\maxDROP} & \deltacell{26.1}{-0.7}{\maxGPQA} & \deltacell{75.9}{-7.3}{\maxGSMEightK} & \deltacell{59.5}{-19.2}{\maxIFEval} & \deltacell{46.7}{-8.3}{\maxTruthfulQA} & \deltacell{60.1}{-6.7}{\maxOLMES} \\
\midrule
\multirow{3}{*}{\shortstack[l]{Base Model\\Parameters}} & Freeze & N=4 & \deltacell{47.9}{-16.6}{\maxAGIEval} & \deltacell{84.7}{-0.6}{\maxHumanEval} & \deltacell{75.1}{-4.7}{\maxHumanEvalPlus} & \deltacell{45.9}{-15.6}{\maxDROP} & \deltacell{25.7}{-1.1}{\maxGPQA} & \deltacell{77.0}{-6.2}{\maxGSMEightK} & \deltacell{50.5}{-28.2}{\maxIFEval} & \deltacell{48.3}{-6.7}{\maxTruthfulQA} & \deltacell{56.9}{-9.9}{\maxOLMES} \\
& SPD & $\lambda$=5 & \deltacell{59.9}{-4.6}{\maxAGIEval} & \deltacell{83.9}{-1.4}{\maxHumanEval} & \deltacell{80.2}{0.4}{\maxHumanEvalPlus} & \deltacell{46.4}{-15.1}{\maxDROP} & \deltacell{28.8}{2.0}{\maxGPQA} & \deltacell{82.7}{-0.5}{\maxGSMEightK} & \deltacell{71.0}{-7.7}{\maxIFEval} & \deltacell{52.6}{-2.4}{\maxTruthfulQA} & \deltacell{63.2}{-3.6}{\maxOLMES} \\
& LoRA & Rank=32, Alpha=64 & \deltacell{51.4}{-13.1}{\maxAGIEval} & \deltacell{84.9}{-0.4}{\maxHumanEval} & \deltacell{80.1}{0.3}{\maxHumanEvalPlus} & \deltacell{67.4}{5.9}{\maxDROP} & \deltacell{24.1}{-2.7}{\maxGPQA} & \deltacell{77.8}{-5.4}{\maxGSMEightK} & \deltacell{64.3}{-14.4}{\maxIFEval} & \deltacell{45.7}{-9.3}{\maxTruthfulQA} & \deltacell{62.0}{-4.8}{\maxOLMES} \\
\bottomrule
\end{tabular}%
}
\caption{Per-task scores on the general benchmarks for bidirectional Arabic training on the Llama 3.1 8B Instruct model.}
\label{tab:bestArabicScoresOLMESBreakdown8B}
\end{table*}

\renewcommand{\maxCOMETxling}{2.5}
\renewcommand{\maxBLEUxling}{8.9}
\renewcommand{\maxFormalAcc}{70.3}
\renewcommand{\maxInformalAcc}{38.4}
\renewcommand{\maxAvgAcc}{38.4}

\begin{table*}[t]
\refstepcounter{subsection}
\subsection*{\thesubsection\quad Stage 2: Detailed Formality, Grammatical Gender and Length Control Scores for Spanish Training Runs}
\centering
\small
\fitwidth{%
\begin{tabular}{lllccccccccccc}
\toprule
& & & \multicolumn{5}{c}{en$\rightarrow$es (N=600)} & \multicolumn{5}{c}{en$\rightarrow$de (N=600)} \\
\cmidrule(lr){4-8} \cmidrule(lr){9-13}
Anchoring & Method & Hyperparameters & COMET & BLEU & Formal M-Acc. & Informal M-Acc. & Avg. & COMET & BLEU & Formal M-Acc. & Informal M-Acc. & Avg. \\
\midrule
\multicolumn{3}{l}{Llama 3.1 8B Instruct} & 84.9 & 38.8 & 79.1 (M=359) & 94.0 (M=252) & 86.6 & 83.3 & 35.3 & 96.1 (M=434) & 91.8 (M=437) & 94.0 \\
\midrule
\multicolumn{3}{l}{Standard SFT} & \deltacell{85.7}{0.8}{\maxCOMETxling} & \deltacell{40.5}{1.7}{\maxBLEUxling} & \deltacell{50.0}{-29.1}{\maxFormalAcc} (M=302) & \deltacell{62.2}{-31.8}{\maxInformalAcc} (M=286) & \deltacell{56.1}{-30.5}{\maxAvgAcc} & \deltacell{81.8}{-1.5}{\maxCOMETxling} & \deltacell{31.8}{-3.5}{\maxBLEUxling} & \deltacell{58.4}{-37.7}{\maxFormalAcc} (M=416) & \deltacell{59.1}{-32.7}{\maxInformalAcc} (M=406) & \deltacell{58.8}{-35.2}{\maxAvgAcc} \\
\midrule
\multirow{11}{*}{\shortstack[l]{Auxiliary\\Data}} & \multirow{7}{*}{\shortstack[l]{Data\\Mixing}} & Tülu (full) & \deltacell{85.6}{0.7}{\maxCOMETxling} & \deltacell{40.9}{2.1}{\maxBLEUxling} & \deltacell{52.5}{-26.6}{\maxFormalAcc} (M=301) & \deltacell{66.4}{-27.6}{\maxInformalAcc} (M=283) & \deltacell{59.5}{-27.1}{\maxAvgAcc} & \deltacell{83.0}{-0.3}{\maxCOMETxling} & \deltacell{33.4}{-1.9}{\maxBLEUxling} & \deltacell{67.4}{-28.7}{\maxFormalAcc} (M=426) & \deltacell{57.3}{-34.5}{\maxInformalAcc} (M=419) & \deltacell{62.4}{-31.6}{\maxAvgAcc} \\
& & Tülu (800) + MT-IF (1600) & \deltacell{86.5}{1.6}{\maxCOMETxling} & \deltacell{47.7}{8.9}{\maxBLEUxling} & \deltacell{99.1}{20.0}{\maxFormalAcc} (M=425) & \deltacell{91.0}{-3.0}{\maxInformalAcc} (M=255) & \deltacell{95.1}{8.5}{\maxAvgAcc} & \deltacell{83.1}{-0.2}{\maxCOMETxling} & \deltacell{36.6}{1.3}{\maxBLEUxling} & \deltacell{98.5}{2.4}{\maxFormalAcc} (M=401) & \deltacell{87.7}{-4.1}{\maxInformalAcc} (M=447) & \deltacell{93.1}{-0.9}{\maxAvgAcc} \\
& & MT-IF (1600) & \deltacell{86.3}{1.4}{\maxCOMETxling} & \deltacell{44.8}{6.0}{\maxBLEUxling} & \deltacell{92.8}{13.7}{\maxFormalAcc} (M=391) & \deltacell{89.6}{-4.4}{\maxInformalAcc} (M=259) & \deltacell{91.2}{4.6}{\maxAvgAcc} & \deltacell{84.1}{0.8}{\maxCOMETxling} & \deltacell{41.5}{6.2}{\maxBLEUxling} & \deltacell{99.0}{2.9}{\maxFormalAcc} (M=480) & \deltacell{94.6}{2.8}{\maxInformalAcc} (M=482) & \deltacell{96.8}{2.8}{\maxAvgAcc} \\
& & Tülu (800) + Gender (800) & \deltacell{85.6}{0.7}{\maxCOMETxling} & \deltacell{40.5}{1.7}{\maxBLEUxling} & \deltacell{46.0}{-33.1}{\maxFormalAcc} (M=287) & \deltacell{61.3}{-32.7}{\maxInformalAcc} (M=284) & \deltacell{53.7}{-32.9}{\maxAvgAcc} & \deltacell{82.3}{-1.0}{\maxCOMETxling} & \deltacell{31.6}{-3.7}{\maxBLEUxling} & \deltacell{55.1}{-41.0}{\maxFormalAcc} (M=423) & \deltacell{60.0}{-31.8}{\maxInformalAcc} (M=425) & \deltacell{55.6}{-38.4}{\maxAvgAcc} \\
& & \quad + diff Gender prompt & \deltacell{85.9}{1.0}{\maxCOMETxling} & \deltacell{42.0}{3.2}{\maxBLEUxling} & \deltacell{46.2}{-32.9}{\maxFormalAcc} (M=292) & \deltacell{67.0}{-27.0}{\maxInformalAcc} (M=276) & \deltacell{56.6}{-30.0}{\maxAvgAcc} & \deltacell{81.7}{-1.6}{\maxCOMETxling} & \deltacell{31.7}{-3.6}{\maxBLEUxling} & \deltacell{52.9}{-43.2}{\maxFormalAcc} (M=435) & \deltacell{64.5}{-27.3}{\maxInformalAcc} (M=431) & \deltacell{58.7}{-35.3}{\maxAvgAcc} \\
& & Tülu (800) + Formality (800) & \deltacell{86.4}{1.5}{\maxCOMETxling} & \deltacell{47.2}{8.4}{\maxBLEUxling} & \deltacell{98.6}{19.5}{\maxFormalAcc} (M=427) & \deltacell{92.3}{-1.7}{\maxInformalAcc} (M=248) & \deltacell{95.5}{8.9}{\maxAvgAcc} & \deltacell{82.5}{-0.8}{\maxCOMETxling} & \deltacell{35.9}{0.6}{\maxBLEUxling} & \deltacell{98.7}{2.6}{\maxFormalAcc} (M=379) & \deltacell{88.0}{-3.8}{\maxInformalAcc} (M=458) & \deltacell{93.4}{-0.6}{\maxAvgAcc} \\
& & \quad + diff Formality prompt & \deltacell{85.5}{0.6}{\maxCOMETxling} & \deltacell{40.3}{1.5}{\maxBLEUxling} & \deltacell{45.9}{-33.2}{\maxFormalAcc} (M=290) & \deltacell{61.8}{-32.2}{\maxInformalAcc} (M=280) & \deltacell{53.8}{-32.8}{\maxAvgAcc} & \deltacell{81.5}{-1.8}{\maxCOMETxling} & \deltacell{31.0}{-4.3}{\maxBLEUxling} & \deltacell{54.9}{-41.2}{\maxFormalAcc} (M=430) & \deltacell{60.5}{-31.3}{\maxInformalAcc} (M=420) & \deltacell{57.7}{-36.3}{\maxAvgAcc} \\
\cmidrule(lr){2-13}
& Merge & Trans=0.5, Tülu+MT-IF=0.5 & \deltacell{85.7}{0.8}{\maxCOMETxling} & \deltacell{41.6}{2.8}{\maxBLEUxling} & \deltacell{47.2}{-31.9}{\maxFormalAcc} (M=299) & \deltacell{74.7}{-19.3}{\maxInformalAcc} (M=273) & \deltacell{61.0}{-25.6}{\maxAvgAcc} & \deltacell{82.9}{-0.4}{\maxCOMETxling} & \deltacell{34.9}{-0.4}{\maxBLEUxling} & \deltacell{67.8}{-28.3}{\maxFormalAcc} (M=451) & \deltacell{61.9}{-29.9}{\maxInformalAcc} (M=449) & \deltacell{64.9}{-29.1}{\maxAvgAcc} \\
\cmidrule(lr){2-13}
& \multirow{3}{*}{EWC} & $\lambda$=1e3, Tülu (10k) & \deltacell{85.7}{0.8}{\maxCOMETxling} & \deltacell{40.8}{2.0}{\maxBLEUxling} & \deltacell{52.3}{-26.8}{\maxFormalAcc} (M=304) & \deltacell{58.6}{-35.4}{\maxInformalAcc} (M=290) & \deltacell{55.5}{-31.1}{\maxAvgAcc} & \deltacell{82.3}{-1.0}{\maxCOMETxling} & \deltacell{32.9}{-2.4}{\maxBLEUxling} & \deltacell{65.3}{-30.8}{\maxFormalAcc} (M=429) & \deltacell{55.6}{-36.2}{\maxInformalAcc} (M=430) & \deltacell{60.5}{-33.5}{\maxAvgAcc} \\
& & $\lambda$=1e3, Tülu (800) + MT-IF (1600) & \deltacell{85.8}{0.9}{\maxCOMETxling} & \deltacell{40.9}{2.1}{\maxBLEUxling} & \deltacell{52.1}{-27.0}{\maxFormalAcc} (M=303) & \deltacell{64.2}{-29.8}{\maxInformalAcc} (M=285) & \deltacell{58.2}{-28.4}{\maxAvgAcc} & \deltacell{82.4}{-0.9}{\maxCOMETxling} & \deltacell{32.8}{-2.5}{\maxBLEUxling} & \deltacell{69.2}{-26.9}{\maxFormalAcc} (M=419) & \deltacell{54.7}{-37.1}{\maxInformalAcc} (M=422) & \deltacell{61.2}{-32.8}{\maxAvgAcc} \\
& & $\lambda$=1e3, MT-IF (1600) & \deltacell{85.7}{0.8}{\maxCOMETxling} & \deltacell{40.9}{2.1}{\maxBLEUxling} & \deltacell{53.6}{-25.5}{\maxFormalAcc} (M=308) & \deltacell{62.6}{-31.4}{\maxInformalAcc} (M=286) & \deltacell{58.1}{-28.5}{\maxAvgAcc} & \deltacell{82.4}{-0.9}{\maxCOMETxling} & \deltacell{32.8}{-2.5}{\maxBLEUxling} & \deltacell{70.0}{-26.1}{\maxFormalAcc} (M=427) & \deltacell{53.4}{-38.4}{\maxInformalAcc} (M=423) & \deltacell{61.7}{-32.3}{\maxAvgAcc} \\
\midrule
\multirow{4}{*}{\shortstack[l]{Model\\Outputs}} & KL-Full & $\lambda$=0.05 & \deltacell{85.7}{0.8}{\maxCOMETxling} & \deltacell{41.0}{2.2}{\maxBLEUxling} & \deltacell{47.5}{-31.6}{\maxFormalAcc} (M=295) & \deltacell{64.4}{-29.6}{\maxInformalAcc} (M=281) & \deltacell{56.0}{-30.6}{\maxAvgAcc} & \deltacell{82.1}{-1.2}{\maxCOMETxling} & \deltacell{32.0}{-3.3}{\maxBLEUxling} & \deltacell{54.6}{-41.5}{\maxFormalAcc} (M=425) & \deltacell{63.6}{-28.2}{\maxInformalAcc} (M=418) & \deltacell{59.1}{-34.9}{\maxAvgAcc} \\
& STM & X=0.25 & \deltacell{84.9}{0.0}{\maxCOMETxling} & \deltacell{36.7}{-2.1}{\maxBLEUxling} & \deltacell{54.4}{-24.7}{\maxFormalAcc} (M=298) & \deltacell{58.0}{-36.0}{\maxInformalAcc} (M=283) & \deltacell{56.2}{-30.4}{\maxAvgAcc} & \deltacell{80.8}{-2.5}{\maxCOMETxling} & \deltacell{29.2}{-6.1}{\maxBLEUxling} & \deltacell{60.6}{-35.5}{\maxFormalAcc} (M=376) & \deltacell{54.5}{-37.3}{\maxInformalAcc} (M=374) & \deltacell{57.6}{-36.4}{\maxAvgAcc} \\
& CC & X=0.25 & \deltacell{83.9}{-1.0}{\maxCOMETxling} & \deltacell{34.3}{-4.5}{\maxBLEUxling} & \deltacell{11.2}{-67.9}{\maxFormalAcc} (M=259) & \deltacell{93.7}{-0.3}{\maxInformalAcc} (M=255) & \deltacell{52.3}{-34.3}{\maxAvgAcc} & \deltacell{81.6}{-1.7}{\maxCOMETxling} & \deltacell{31.9}{-3.4}{\maxBLEUxling} & \deltacell{25.8}{-70.3}{\maxFormalAcc} (M=414) & \deltacell{86.2}{-5.6}{\maxInformalAcc} (M=413) & \deltacell{56.0}{-38.0}{\maxAvgAcc} \\
& EAFT & K=20, Weight=1.0 & \deltacell{85.7}{0.8}{\maxCOMETxling} & \deltacell{40.7}{1.9}{\maxBLEUxling} & \deltacell{48.1}{-31.0}{\maxFormalAcc} (M=291) & \deltacell{62.7}{-31.3}{\maxInformalAcc} (M=279) & \deltacell{55.4}{-31.2}{\maxAvgAcc} & \deltacell{82.0}{-1.3}{\maxCOMETxling} & \deltacell{32.1}{-3.2}{\maxBLEUxling} & \deltacell{52.5}{-43.6}{\maxFormalAcc} (M=434) & \deltacell{62.0}{-29.8}{\maxInformalAcc} (M=434) & \deltacell{57.3}{-36.7}{\maxAvgAcc} \\
\midrule
\multirow{3}{*}{\shortstack[l]{Base Model\\Parameters}} & Freeze & N=4 & \deltacell{85.7}{0.8}{\maxCOMETxling} & \deltacell{40.7}{1.9}{\maxBLEUxling} & \deltacell{49.2}{-29.9}{\maxFormalAcc} (M=299) & \deltacell{64.5}{-29.5}{\maxInformalAcc} (M=282) & \deltacell{56.9}{-29.7}{\maxAvgAcc} & \deltacell{82.3}{-1.0}{\maxCOMETxling} & \deltacell{31.9}{-3.4}{\maxBLEUxling} & \deltacell{58.0}{-38.1}{\maxFormalAcc} (M=419) & \deltacell{59.2}{-32.6}{\maxInformalAcc} (M=412) & \deltacell{58.6}{-35.4}{\maxAvgAcc} \\
& SPD & $\lambda$=5 & \deltacell{85.5}{0.6}{\maxCOMETxling} & \deltacell{41.0}{2.2}{\maxBLEUxling} & \deltacell{57.8}{-21.3}{\maxFormalAcc} (M=325) & \deltacell{86.1}{-7.9}{\maxInformalAcc} (M=267) & \deltacell{72.0}{-14.6}{\maxAvgAcc} & \deltacell{82.7}{-0.6}{\maxCOMETxling} & \deltacell{33.1}{-2.2}{\maxBLEUxling} & \deltacell{72.8}{-23.3}{\maxFormalAcc} (M=419) & \deltacell{78.4}{-13.4}{\maxInformalAcc} (M=426) & \deltacell{75.6}{-18.4}{\maxAvgAcc} \\
& LoRA & Rank=32, Alpha=64 & \deltacell{85.5}{0.6}{\maxCOMETxling} & \deltacell{40.2}{1.4}{\maxBLEUxling} & \deltacell{51.3}{-27.8}{\maxFormalAcc} (M=302) & \deltacell{66.9}{-27.1}{\maxInformalAcc} (M=287) & \deltacell{59.1}{-27.5}{\maxAvgAcc} & \deltacell{81.9}{-1.4}{\maxCOMETxling} & \deltacell{32.3}{-3.0}{\maxBLEUxling} & \deltacell{71.9}{-24.2}{\maxFormalAcc} (M=420) & \deltacell{54.2}{-37.6}{\maxInformalAcc} (M=421) & \deltacell{63.1}{-30.9}{\maxAvgAcc} \\
\bottomrule
\end{tabular}%
}
\caption{Formality Control scores (including number of matches used for accuracy scores) for bidirectional Spanish training on the Llama 3.1 8B Instruct model.}
\label{tab:spanish-formality-control-detailed}
\end{table*}

\renewcommand{\maxCOMETxling}{4.2}
\renewcommand{\maxBLEUxling}{12.1}
\renewcommand{\maxAccOne}{57.0}
\renewcommand{\maxAccTwo}{46.7}
\renewcommand{\maxAccThree}{51.0}
\begin{table*}[t]
\centering
\small
\fitwidth{%
\begin{tabular}{lllccccccccccc}
\toprule
& & & \multicolumn{5}{c}{en$\rightarrow$es (N=300)} & \multicolumn{5}{c}{en$\rightarrow$de (N=300)} \\
\cmidrule(lr){4-8} \cmidrule(lr){9-13}
Anchoring & Method & Hyperparameters & COMET & BLEU & Fem. Acc. & Masc. Acc. & Avg. & COMET & BLEU & Fem. Acc. & Masc. Acc. & Avg. \\
\midrule
\multicolumn{3}{l}{Llama 3.1 8B Instruct} & 83.1 & 44.5 & 42.3 & 60.7 & 51.5 & 81.2 & 32.8 & 63.3 & 50.7 & 57.0 \\
\midrule
\multicolumn{3}{l}{Standard SFT} & \deltacell{83.0}{-0.1}{\maxCOMETxling} & \deltacell{43.1}{-1.4}{\maxBLEUxling} & \deltacell{18.3}{-24.0}{\maxAccOne} & \deltacell{17.3}{-43.4}{\maxAccTwo} & \deltacell{17.8}{-33.7}{\maxAccThree} & \deltacell{81.4}{0.2}{\maxCOMETxling} & \deltacell{29.2}{-3.6}{\maxBLEUxling} & \deltacell{8.7}{-54.6}{\maxAccOne} & \deltacell{9.7}{-41.0}{\maxAccTwo} & \deltacell{9.2}{-47.8}{\maxAccThree} \\
\midrule
\multirow{12}{*}{\shortstack[l]{Auxiliary\\Data}} & \multirow{8}{*}{\shortstack[l]{Data\\Mixing}} & Tülu (full) & \deltacell{83.0}{-0.1}{\maxCOMETxling} & \deltacell{43.6}{-0.9}{\maxBLEUxling} & \deltacell{20.0}{-22.3}{\maxAccOne} & \deltacell{18.3}{-42.4}{\maxAccTwo} & \deltacell{19.2}{-32.3}{\maxAccThree} & \deltacell{82.5}{1.3}{\maxCOMETxling} & \deltacell{31.9}{-0.9}{\maxBLEUxling} & \deltacell{8.0}{-55.3}{\maxAccOne} & \deltacell{4.0}{-46.7}{\maxAccTwo} & \deltacell{6.0}{-51.0}{\maxAccThree} \\
& & Tülu (800) + MT-IF (1600) & \deltacell{86.2}{3.1}{\maxCOMETxling} & \deltacell{52.8}{8.3}{\maxBLEUxling} & \deltacell{81.7}{39.4}{\maxAccOne} & \deltacell{75.0}{14.3}{\maxAccTwo} & \deltacell{78.4}{26.9}{\maxAccThree} & \deltacell{85.3}{4.1}{\maxCOMETxling} & \deltacell{40.8}{8.0}{\maxBLEUxling} & \deltacell{84.3}{21.0}{\maxAccOne} & \deltacell{82.7}{32.0}{\maxAccTwo} & \deltacell{83.5}{26.5}{\maxAccThree} \\
& & \quad + diff MT-IF prompt & \deltacell{83.5}{0.4}{\maxCOMETxling} & \deltacell{40.9}{-3.6}{\maxBLEUxling} & \deltacell{22.0}{-22.3}{\maxAccOne} & \deltacell{19.7}{-41.0}{\maxAccTwo} & \deltacell{20.8}{-30.7}{\maxAccThree} & \deltacell{82.9}{1.7}{\maxCOMETxling} & \deltacell{32.5}{-0.3}{\maxBLEUxling} & \deltacell{19.7}{-43.6}{\maxAccOne} & \deltacell{15.3}{-35.4}{\maxAccTwo} & \deltacell{17.5}{-39.5}{\maxAccThree} \\
& & MT-IF (1600) & \deltacell{86.2}{3.1}{\maxCOMETxling} & \deltacell{52.5}{8.0}{\maxBLEUxling} & \deltacell{85.3}{43.0}{\maxAccOne} & \deltacell{78.7}{18.0}{\maxAccTwo} & \deltacell{82.0}{30.5}{\maxAccThree} & \deltacell{85.4}{4.2}{\maxCOMETxling} & \deltacell{41.4}{8.6}{\maxBLEUxling} & \deltacell{85.7}{22.4}{\maxAccOne} & \deltacell{84.3}{33.6}{\maxAccTwo} & \deltacell{85.0}{28.0}{\maxAccThree} \\
& & Tülu (800) + Gender (800) & \deltacell{86.2}{3.1}{\maxCOMETxling} & \deltacell{52.9}{8.4}{\maxBLEUxling} & \deltacell{80.0}{37.7}{\maxAccOne} & \deltacell{75.7}{15.0}{\maxAccTwo} & \deltacell{77.9}{26.4}{\maxAccThree} & \deltacell{85.4}{4.2}{\maxCOMETxling} & \deltacell{40.5}{7.7}{\maxBLEUxling} & \deltacell{83.0}{19.7}{\maxAccOne} & \deltacell{84.0}{33.3}{\maxAccTwo} & \deltacell{83.5}{26.5}{\maxAccThree} \\
& & \quad + diff Gender prompt & \deltacell{82.8}{-0.3}{\maxCOMETxling} & \deltacell{43.7}{-0.8}{\maxBLEUxling} & \deltacell{18.0}{-24.3}{\maxAccOne} & \deltacell{17.0}{-43.7}{\maxAccTwo} & \deltacell{17.5}{-34.0}{\maxAccThree} & \deltacell{81.5}{0.3}{\maxCOMETxling} & \deltacell{29.6}{-3.2}{\maxBLEUxling} & \deltacell{6.3}{-57.0}{\maxAccOne} & \deltacell{7.7}{-43.0}{\maxAccTwo} & \deltacell{7.0}{-50.0}{\maxAccThree} \\
& & Tülu (800) + Formality (800) & \deltacell{82.6}{-0.5}{\maxCOMETxling} & \deltacell{43.6}{-0.9}{\maxBLEUxling} & \deltacell{18.3}{-24.0}{\maxAccOne} & \deltacell{17.7}{-43.0}{\maxAccTwo} & \deltacell{18.0}{-33.5}{\maxAccThree} & \deltacell{81.3}{0.1}{\maxCOMETxling} & \deltacell{29.6}{-3.2}{\maxBLEUxling} & \deltacell{8.3}{-55.0}{\maxAccOne} & \deltacell{6.7}{-44.0}{\maxAccTwo} & \deltacell{7.5}{-49.5}{\maxAccThree} \\
& & \quad + diff Formality prompt & \deltacell{83.3}{0.2}{\maxCOMETxling} & \deltacell{44.6}{0.1}{\maxBLEUxling} & \deltacell{20.3}{-22.0}{\maxAccOne} & \deltacell{19.0}{-41.7}{\maxAccTwo} & \deltacell{19.7}{-31.8}{\maxAccThree} & \deltacell{81.7}{0.5}{\maxCOMETxling} & \deltacell{29.9}{-2.9}{\maxBLEUxling} & \deltacell{9.0}{-54.3}{\maxAccOne} & \deltacell{15.7}{-35.0}{\maxAccTwo} & \deltacell{12.3}{-44.7}{\maxAccThree} \\
\cmidrule(lr){2-13}
& Merge & Trans=0.5, Tülu+MT-IF=0.5 & \deltacell{82.9}{-0.2}{\maxCOMETxling} & \deltacell{32.4}{-12.1}{\maxBLEUxling} & \deltacell{22.7}{-19.6}{\maxAccOne} & \deltacell{24.0}{-36.7}{\maxAccTwo} & \deltacell{23.4}{-28.1}{\maxAccThree} & \deltacell{82.9}{1.7}{\maxCOMETxling} & \deltacell{32.4}{-0.4}{\maxBLEUxling} & \deltacell{22.7}{-40.6}{\maxAccOne} & \deltacell{24.0}{-26.7}{\maxAccTwo} & \deltacell{23.4}{-33.6}{\maxAccThree} \\
\cmidrule(lr){2-13}
& \multirow{3}{*}{EWC} & $\lambda$=1e3, Tülu (10k) & \deltacell{82.8}{-0.3}{\maxCOMETxling} & \deltacell{43.2}{-1.3}{\maxBLEUxling} & \deltacell{19.0}{-23.3}{\maxAccOne} & \deltacell{17.3}{-43.4}{\maxAccTwo} & \deltacell{18.2}{-33.3}{\maxAccThree} & \deltacell{82.0}{0.8}{\maxCOMETxling} & \deltacell{30.3}{-2.5}{\maxBLEUxling} & \deltacell{8.3}{-55.0}{\maxAccOne} & \deltacell{8.7}{-42.0}{\maxAccTwo} & \deltacell{8.5}{-48.5}{\maxAccThree} \\
& & $\lambda$=1e3, Tülu (800) + MT-IF (1600) & \deltacell{82.8}{-0.3}{\maxCOMETxling} & \deltacell{43.1}{-1.4}{\maxBLEUxling} & \deltacell{18.3}{-24.0}{\maxAccOne} & \deltacell{17.0}{-43.7}{\maxAccTwo} & \deltacell{17.7}{-33.8}{\maxAccThree} & \deltacell{81.9}{0.7}{\maxCOMETxling} & \deltacell{30.4}{-2.4}{\maxBLEUxling} & \deltacell{8.7}{-54.6}{\maxAccOne} & \deltacell{7.7}{-43.0}{\maxAccTwo} & \deltacell{8.2}{-48.8}{\maxAccThree} \\
& & $\lambda$=1e3, MT-IF (1600) & \deltacell{82.7}{-0.4}{\maxCOMETxling} & \deltacell{43.0}{-1.5}{\maxBLEUxling} & \deltacell{19.0}{-23.3}{\maxAccOne} & \deltacell{17.3}{-43.4}{\maxAccTwo} & \deltacell{18.2}{-33.3}{\maxAccThree} & \deltacell{81.8}{0.6}{\maxCOMETxling} & \deltacell{30.3}{-2.5}{\maxBLEUxling} & \deltacell{8.3}{-55.0}{\maxAccOne} & \deltacell{7.7}{-43.0}{\maxAccTwo} & \deltacell{8.0}{-49.0}{\maxAccThree} \\
\midrule
\multirow{4}{*}{\shortstack[l]{Model\\Outputs}} & KL-Full & $\lambda$=0.05 & \deltacell{83.3}{0.2}{\maxCOMETxling} & \deltacell{43.8}{-0.7}{\maxBLEUxling} & \deltacell{18.7}{-23.6}{\maxAccOne} & \deltacell{17.0}{-43.7}{\maxAccTwo} & \deltacell{17.9}{-33.6}{\maxAccThree} & \deltacell{81.7}{0.5}{\maxCOMETxling} & \deltacell{29.6}{-3.2}{\maxBLEUxling} & \deltacell{7.3}{-56.0}{\maxAccOne} & \deltacell{8.0}{-42.7}{\maxAccTwo} & \deltacell{7.7}{-49.3}{\maxAccThree} \\
& STM & X=0.25 & \deltacell{82.3}{-0.8}{\maxCOMETxling} & \deltacell{37.8}{-6.7}{\maxBLEUxling} & \deltacell{18.0}{-24.3}{\maxAccOne} & \deltacell{17.0}{-43.7}{\maxAccTwo} & \deltacell{17.5}{-34.0}{\maxAccThree} & \deltacell{81.2}{0.0}{\maxCOMETxling} & \deltacell{27.8}{-5.0}{\maxBLEUxling} & \deltacell{8.7}{-54.6}{\maxAccOne} & \deltacell{9.3}{-41.4}{\maxAccTwo} & \deltacell{9.0}{-48.0}{\maxAccThree} \\
& CC & X=0.25 & \deltacell{81.6}{-1.5}{\maxCOMETxling} & \deltacell{39.8}{-4.7}{\maxBLEUxling} & \deltacell{20.3}{-22.0}{\maxAccOne} & \deltacell{19.0}{-41.7}{\maxAccTwo} & \deltacell{19.7}{-31.8}{\maxAccThree} & \deltacell{79.8}{-1.4}{\maxCOMETxling} & \deltacell{28.2}{-4.6}{\maxBLEUxling} & \deltacell{8.3}{-55.0}{\maxAccOne} & \deltacell{10.3}{-40.4}{\maxAccTwo} & \deltacell{9.3}{-47.7}{\maxAccThree} \\
& EAFT & K=20, Weight=1.0 & \deltacell{83.2}{0.1}{\maxCOMETxling} & \deltacell{43.8}{-0.7}{\maxBLEUxling} & \deltacell{18.0}{-24.3}{\maxAccOne} & \deltacell{17.3}{-43.4}{\maxAccTwo} & \deltacell{17.7}{-33.8}{\maxAccThree} & \deltacell{81.8}{0.6}{\maxCOMETxling} & \deltacell{29.4}{-3.4}{\maxBLEUxling} & \deltacell{8.0}{-55.3}{\maxAccOne} & \deltacell{8.7}{-42.0}{\maxAccTwo} & \deltacell{8.4}{-48.6}{\maxAccThree} \\
\midrule
\multirow{3}{*}{\shortstack[l]{Base Model\\Parameters}} & Freeze & N=4 & \deltacell{83.2}{0.1}{\maxCOMETxling} & \deltacell{42.9}{-1.6}{\maxBLEUxling} & \deltacell{17.7}{-24.6}{\maxAccOne} & \deltacell{17.7}{-43.0}{\maxAccTwo} & \deltacell{17.7}{-33.8}{\maxAccThree} & \deltacell{82.1}{0.9}{\maxCOMETxling} & \deltacell{29.3}{-3.5}{\maxBLEUxling} & \deltacell{7.0}{-56.3}{\maxAccOne} & \deltacell{8.7}{-42.0}{\maxAccTwo} & \deltacell{7.9}{-49.1}{\maxAccThree} \\
& SPD & $\lambda$=5 & \deltacell{84.2}{1.1}{\maxCOMETxling} & \deltacell{46.0}{1.5}{\maxBLEUxling} & \deltacell{47.0}{4.7}{\maxAccOne} & \deltacell{30.0}{-30.7}{\maxAccTwo} & \deltacell{38.5}{-13.0}{\maxAccThree} & \deltacell{82.9}{1.7}{\maxCOMETxling} & \deltacell{32.2}{-0.6}{\maxBLEUxling} & \deltacell{39.3}{-24.0}{\maxAccOne} & \deltacell{38.3}{-12.4}{\maxAccTwo} & \deltacell{38.8}{-18.2}{\maxAccThree} \\
& LoRA & Rank=32, Alpha=64 & \deltacell{82.8}{-0.3}{\maxCOMETxling} & \deltacell{42.6}{-1.9}{\maxBLEUxling} & \deltacell{17.0}{-25.3}{\maxAccOne} & \deltacell{15.7}{-45.0}{\maxAccTwo} & \deltacell{16.4}{-35.1}{\maxAccThree} & \deltacell{82.0}{0.8}{\maxCOMETxling} & \deltacell{30.2}{-2.6}{\maxBLEUxling} & \deltacell{6.7}{-56.6}{\maxAccOne} & \deltacell{8.3}{-42.4}{\maxAccTwo} & \deltacell{7.5}{-49.5}{\maxAccThree} \\
\bottomrule
\end{tabular}%
}
\caption{Grammatical Gender Control scores for bidirectional Spanish training on the Llama 3.1 8B Instruct model. Fem. Acc. and Masc. Acc. are the accuracies of producing the requested feminine and masculine forms.}
\label{tab:spanish-gender-control}
\end{table*}

\renewcommand{\maxCOMETavg}{9.7}
\renewcommand{\maxPct}{60.80}
\begin{table*}
\centering
\small
\fitwidth{%
\begin{tabular}{lllcccccccc}
\toprule
& & & \multicolumn{8}{c}{en$\leftrightarrow$es} \\
\cmidrule(lr){4-11}
& & & \multicolumn{2}{c}{Baseline} & \multicolumn{3}{c}{Shorter} & \multicolumn{3}{c}{Longer} \\
\cmidrule(lr){4-5} \cmidrule(lr){6-8} \cmidrule(lr){9-11}
Anchoring & Method & Hyperparameters & Avg. COMET & Tgt/Src & Avg. COMET & \% Shorter & Tgt/Src & Avg. COMET & \% Longer & Tgt/Src \\
\midrule
\multicolumn{3}{l}{Llama 3.1 8B Instruct} & 86.0 & 1.040 & 86.0 & 80.35 & 0.941 & 84.0 & 70.20 & 1.349 \\
\midrule
\multicolumn{3}{l}{Standard SFT} & \deltacell{87.0}{1.0}{\maxCOMETavg} & 0.998 & \deltacell{86.6}{0.6}{\maxCOMETavg} & \deltacell{21.40}{-58.95}{\maxPct} & 0.994 & \deltacell{83.7}{-0.3}{\maxCOMETavg} & \deltacell{24.45}{-45.75}{\maxPct} & 1.010 \\
\midrule
\multirow{12}{*}{\shortstack[l]{Auxiliary\\Data}} & \multirow{8}{*}{\shortstack[l]{Data\\Mixing}} & Tülu (full) & \deltacell{86.9}{0.9}{\maxCOMETavg} & 0.997 & \deltacell{86.8}{0.8}{\maxCOMETavg} & \deltacell{26.90}{-53.45}{\maxPct} & 0.990 & \deltacell{86.0}{2.0}{\maxCOMETavg} & \deltacell{22.95}{-47.25}{\maxPct} & 1.004 \\
& & Tülu (800) + MT-IF (1600) & \deltacell{87.0}{1.0}{\maxCOMETavg} & 0.997 & \deltacell{86.9}{0.9}{\maxCOMETavg} & \deltacell{27.25}{-53.10}{\maxPct} & 0.987 & \deltacell{86.0}{2.0}{\maxCOMETavg} & \deltacell{27.70}{-42.50}{\maxPct} & 1.006 \\
& & \quad + diff MT-IF prompt & \deltacell{87.0}{1.0}{\maxCOMETavg} & 0.997 & \deltacell{86.9}{0.9}{\maxCOMETavg} & \deltacell{22.58}{-57.77}{\maxPct} & 0.981 & \deltacell{85.8}{1.8}{\maxCOMETavg} & \deltacell{22.97}{-47.23}{\maxPct} & 0.999 \\
& & MT-IF (1600) & \deltacell{87.0}{1.0}{\maxCOMETavg} & 0.998 & \deltacell{86.7}{0.7}{\maxCOMETavg} & \deltacell{23.30}{-57.05}{\maxPct} & 0.991 & \deltacell{84.1}{0.1}{\maxCOMETavg} & \deltacell{24.90}{-45.30}{\maxPct} & 0.999 \\
& & Tülu (800) + Gender (800) & \deltacell{87.0}{1.0}{\maxCOMETavg} & 0.998 & \deltacell{86.9}{0.9}{\maxCOMETavg} & \deltacell{21.10}{-59.25}{\maxPct} & 0.992 & \deltacell{86.9}{2.9}{\maxCOMETavg} & \deltacell{23.00}{-47.20}{\maxPct} & 1.001 \\
& & \quad + diff Gender prompt & \deltacell{87.0}{1.0}{\maxCOMETavg} & 0.998 & \deltacell{86.9}{0.9}{\maxCOMETavg} & \deltacell{19.55}{-60.80}{\maxPct} & 0.995 & \deltacell{86.3}{2.3}{\maxCOMETavg} & \deltacell{23.45}{-46.75}{\maxPct} & 1.000 \\
& & Tülu (800) + Formality (800) & \deltacell{87.0}{1.0}{\maxCOMETavg} & 0.998 & \deltacell{86.9}{0.9}{\maxCOMETavg} & \deltacell{21.65}{-58.70}{\maxPct} & 0.994 & \deltacell{86.4}{2.4}{\maxCOMETavg} & \deltacell{24.50}{-45.70}{\maxPct} & 1.002 \\
& & \quad + diff Formality prompt & \deltacell{87.0}{1.0}{\maxCOMETavg} & 0.997 & \deltacell{86.9}{0.9}{\maxCOMETavg} & \deltacell{20.70}{-59.65}{\maxPct} & 0.992 & \deltacell{86.6}{2.6}{\maxCOMETavg} & \deltacell{22.80}{-47.40}{\maxPct} & 1.001 \\
\cmidrule(lr){2-11}
& Merge & Trans=0.5, Tülu+MT-IF=0.5 & \deltacell{87.0}{1.0}{\maxCOMETavg} & 0.996 & \deltacell{86.9}{0.9}{\maxCOMETavg} & \deltacell{27.25}{-53.10}{\maxPct} & 0.988 & \deltacell{85.7}{1.7}{\maxCOMETavg} & \deltacell{24.65}{-45.55}{\maxPct} & 1.001 \\
\cmidrule(lr){2-11}
& \multirow{3}{*}{EWC} & $\lambda$=1e3, Tülu (10k) & \deltacell{87.0}{1.0}{\maxCOMETavg} & 1.000 & \deltacell{84.2}{-1.8}{\maxCOMETavg} & \deltacell{27.10}{-53.25}{\maxPct} & 1.027 & \deltacell{74.3}{-9.7}{\maxCOMETavg} & \deltacell{37.20}{-33.00}{\maxPct} & 1.103 \\
& & $\lambda$=1e3, Tülu (800) + MT-IF (1600) & \deltacell{87.0}{1.0}{\maxCOMETavg} & 0.999 & \deltacell{85.8}{-0.2}{\maxCOMETavg} & \deltacell{27.10}{-53.25}{\maxPct} & 1.006 & \deltacell{77.4}{-6.6}{\maxCOMETavg} & \deltacell{31.30}{-38.90}{\maxPct} & 1.040 \\
& & $\lambda$=1e3, MT-IF (1600) & \deltacell{87.0}{1.0}{\maxCOMETavg} & 0.998 & \deltacell{86.0}{0.0}{\maxCOMETavg} & \deltacell{26.60}{-53.75}{\maxPct} & 0.992 & \deltacell{79.0}{-5.0}{\maxCOMETavg} & \deltacell{29.35}{-40.85}{\maxPct} & 1.028 \\
\midrule
\multirow{4}{*}{\shortstack[l]{Model\\Outputs}} & KL-Full & $\lambda$=0.05 & \deltacell{86.9}{0.9}{\maxCOMETavg} & 1.002 & \deltacell{86.8}{0.8}{\maxCOMETavg} & \deltacell{20.80}{-59.55}{\maxPct} & 1.000 & \deltacell{85.5}{1.5}{\maxCOMETavg} & \deltacell{24.00}{-46.20}{\maxPct} & 1.008 \\
& STM & X=0.25 & \deltacell{86.6}{0.6}{\maxCOMETavg} & 0.993 & \deltacell{86.2}{0.2}{\maxCOMETavg} & \deltacell{24.00}{-56.35}{\maxPct} & 0.989 & \deltacell{83.5}{-0.5}{\maxCOMETavg} & \deltacell{23.60}{-46.60}{\maxPct} & 0.998 \\
& CC & X=0.25 & \deltacell{87.0}{1.0}{\maxCOMETavg} & 0.997 & \deltacell{86.3}{0.3}{\maxCOMETavg} & \deltacell{20.45}{-59.90}{\maxPct} & 0.992 & \deltacell{83.1}{-0.9}{\maxCOMETavg} & \deltacell{24.50}{-45.70}{\maxPct} & 0.996 \\
& EAFT & K=20, Weight=1.0 & \deltacell{87.0}{1.0}{\maxCOMETavg} & 0.998 & \deltacell{86.6}{0.6}{\maxCOMETavg} & \deltacell{19.75}{-60.60}{\maxPct} & 0.998 & \deltacell{83.2}{-0.8}{\maxCOMETavg} & \deltacell{25.30}{-44.90}{\maxPct} & 1.014 \\
\midrule
\multirow{3}{*}{\shortstack[l]{Base Model\\Parameters}} & Freeze & N=4 & \deltacell{87.0}{1.0}{\maxCOMETavg} & 0.997 & \deltacell{86.9}{0.9}{\maxCOMETavg} & \deltacell{23.25}{-57.10}{\maxPct} & 0.990 & \deltacell{86.3}{2.3}{\maxCOMETavg} & \deltacell{22.85}{-47.35}{\maxPct} & 1.002 \\
& SPD & $\lambda$=5 & \deltacell{86.9}{0.9}{\maxCOMETavg} & 0.998 & \deltacell{86.7}{0.7}{\maxCOMETavg} & \deltacell{58.35}{-22.00}{\maxPct} & 0.951 & \deltacell{86.7}{2.7}{\maxCOMETavg} & \deltacell{40.60}{-29.60}{\maxPct} & 1.015 \\
& LoRA & Rank=32, Alpha=64 & \deltacell{86.9}{0.9}{\maxCOMETavg} & 0.997 & \deltacell{86.8}{0.8}{\maxCOMETavg} & \deltacell{21.15}{-59.20}{\maxPct} & 1.000 & \deltacell{85.1}{1.1}{\maxCOMETavg} & \deltacell{26.90}{-43.30}{\maxPct} & 1.044 \\
\bottomrule
\end{tabular}%
}
\caption{Effect of length instruction (none/shorter/longer) on length control scores for en$\leftrightarrow$es translation with Llama 3.1 8B Instruct, evaluated on FLORES (bidirectional, 2x N=$1012$).}
\label{tab:spanish-length-control}
\end{table*}

\begin{table*}[t]
\refstepcounter{subsection}
\subsection*{\thesubsection\quad Stage 2: Arabic Scores}
\centering
\small
\renewcommand{\maxCOMET}{2.4}
\renewcommand{\maxBLEU}{6.6}
\renewcommand{\maxOLMES}{10.8}
\renewcommand{\maxMTIF}{42.4}

\fitwidth{%
\begin{tabular}{lllcccccc}
\toprule
& & & \multicolumn{2}{c}{ar$\leftrightarrow$en} & \multicolumn{2}{c}{de$\leftrightarrow$en} & & ar \\
\cmidrule(lr){4-5} \cmidrule(lr){6-7} \cmidrule(lr){9-9}
Anchoring & Method & Hyperparameters & COMET & BLEU & COMET & BLEU & General Avg & MT-IF Avg \\
\midrule
\multicolumn{3}{l}{Llama 3.1 8B Instruct} & 84.4 & 26.4 & 87.6 & 37.3 & 66.8 & 66.8 \\
\midrule
\multicolumn{3}{l}{Standard SFT} & \deltacell{86.7}{2.3}{\maxCOMET} & \deltacell{32.9}{6.5}{\maxBLEU} & \deltacell{87.2}{-0.4}{\maxCOMET} & \deltacell{37.2}{-0.1}{\maxBLEU} & \deltacell{58.5}{-8.3}{\maxOLMES} & \deltacell{28.9}{-37.8}{\maxMTIF} \\
\midrule
\multirow{9}{*}{\shortstack[l]{Auxiliary\\Data}} & \multirow{6}{*}{\shortstack[l]{Data\\Mixing}} & Tülu (full) & \deltacell{86.7}{2.3}{\maxCOMET} & \deltacell{32.5}{6.1}{\maxBLEU} & \deltacell{87.7}{0.1}{\maxCOMET} & \deltacell{38.0}{0.7}{\maxBLEU} & \deltacell{60.4}{-6.4}{\maxOLMES} & \deltacell{26.9}{-39.8}{\maxMTIF} \\
& & Tülu (800) + MT-IF (1600) & \deltacell{86.8}{2.4}{\maxCOMET} & \deltacell{32.7}{6.3}{\maxBLEU} & \deltacell{87.8}{0.2}{\maxCOMET} & \deltacell{38.4}{1.1}{\maxBLEU} & \deltacell{61.9}{-4.9}{\maxOLMES} & \deltacell{51.8}{-15.0}{\maxMTIF} \\
& & \quad + diff MT-IF prompt & \deltacell{86.8}{2.4}{\maxCOMET} & \deltacell{32.6}{6.2}{\maxBLEU} & \deltacell{87.8}{0.2}{\maxCOMET} & \deltacell{38.2}{0.9}{\maxBLEU} & \deltacell{61.7}{-5.1}{\maxOLMES} & \deltacell{27.3}{-39.5}{\maxMTIF} \\
& & MT-IF (1600) & \deltacell{86.7}{2.3}{\maxCOMET} & \deltacell{32.8}{6.4}{\maxBLEU} & \deltacell{87.4}{-0.2}{\maxCOMET} & \deltacell{37.6}{0.3}{\maxBLEU} & \deltacell{59.8}{-7.0}{\maxOLMES} & \deltacell{55.1}{-11.7}{\maxMTIF} \\
& & Tülu (800) + Gender (800) & \deltacell{86.7}{2.3}{\maxCOMET} & \deltacell{32.6}{6.2}{\maxBLEU} & \deltacell{87.4}{-0.2}{\maxCOMET} & \deltacell{37.6}{0.3}{\maxBLEU} & \deltacell{61.9}{-4.9}{\maxOLMES} & \deltacell{53.0}{-13.8}{\maxMTIF} \\
& & Tülu (800) + Formality (800) & \deltacell{86.7}{2.3}{\maxCOMET} & \deltacell{32.7}{6.3}{\maxBLEU} & \deltacell{87.4}{-0.2}{\maxCOMET} & \deltacell{37.8}{0.5}{\maxBLEU} & \deltacell{62.5}{-4.3}{\maxOLMES} & \deltacell{24.4}{-42.4}{\maxMTIF} \\
\cmidrule(lr){2-9}
& Merge & Trans=0.5, Tülu+MT-IF=0.5 & \deltacell{86.4}{2.0}{\maxCOMET} & \deltacell{31.8}{5.4}{\maxBLEU} & \deltacell{87.9}{0.3}{\maxCOMET} & \deltacell{38.5}{1.2}{\maxBLEU} & \deltacell{62.5}{-4.3}{\maxOLMES} & \deltacell{33.2}{-33.6}{\maxMTIF} \\
\cmidrule(lr){2-9}
& \multirow{3}{*}{EWC} & $\lambda$=1e3, Tülu (10k) & \deltacell{86.6}{2.2}{\maxCOMET} & \deltacell{32.3}{5.9}{\maxBLEU} & \deltacell{87.4}{-0.2}{\maxCOMET} & \deltacell{37.7}{0.4}{\maxBLEU} & \deltacell{66.1}{-0.7}{\maxOLMES} & \deltacell{32.7}{-34.0}{\maxMTIF} \\
& & $\lambda$=1e3, Tülu (800) + MT-IF (1600) & \deltacell{86.6}{2.2}{\maxCOMET} & \deltacell{32.7}{6.3}{\maxBLEU} & \deltacell{87.3}{-0.3}{\maxCOMET} & \deltacell{37.4}{0.1}{\maxBLEU} & \deltacell{63.3}{-3.5}{\maxOLMES} & \deltacell{28.2}{-38.6}{\maxMTIF} \\
& & $\lambda$=1e3, MT-IF (1600) & \deltacell{86.7}{2.3}{\maxCOMET} & \deltacell{32.7}{6.3}{\maxBLEU} & \deltacell{87.5}{-0.1}{\maxCOMET} & \deltacell{37.4}{0.1}{\maxBLEU} & \deltacell{60.1}{-6.7}{\maxOLMES} & \deltacell{29.3}{-37.55}{\maxMTIF} \\
\midrule
\multirow{4}{*}{\shortstack[l]{Model\\Outputs}} & KL-Full & $\lambda$=0.05 & \deltacell{86.8}{2.4}{\maxCOMET} & \deltacell{33.1}{6.7}{\maxBLEU} & \deltacell{87.4}{-0.2}{\maxCOMET} & \deltacell{37.7}{0.4}{\maxBLEU} & \deltacell{60.4}{-6.4}{\maxOLMES} & \deltacell{26.1}{-40.6}{\maxMTIF} \\
& STM & X=0.25 & \deltacell{86.2}{1.8}{\maxCOMET} & \deltacell{31.0}{4.3}{\maxBLEU} & \deltacell{87.1}{-0.5}{\maxCOMET} & \deltacell{36.6}{-0.7}{\maxBLEU} & \deltacell{56.0}{-10.8}{\maxOLMES} & \deltacell{30.0}{-36.8}{\maxMTIF} \\
& CC & X=0.25 & \deltacell{86.8}{2.4}{\maxCOMET} & \deltacell{33.0}{6.6}{\maxBLEU} & \deltacell{87.0}{-0.6}{\maxCOMET} & \deltacell{36.9}{-0.4}{\maxBLEU} & \deltacell{57.5}{-9.3}{\maxOLMES} & \deltacell{25.2}{-41.6}{\maxMTIF} \\
& EAFT & K=20, Weight=1.0 & \deltacell{86.8}{2.4}{\maxCOMET} & \deltacell{32.9}{6.5}{\maxBLEU} & \deltacell{87.3}{-0.3}{\maxCOMET} & \deltacell{37.4}{0.1}{\maxBLEU} & \deltacell{60.1}{-6.7}{\maxOLMES} & \deltacell{24.9}{-41.9}{\maxMTIF} \\
\midrule
\multirow{3}{*}{\shortstack[l]{Base Model\\Parameters}} & Freeze & N=4 & \deltacell{86.6}{2.2}{\maxCOMET} & \deltacell{32.6}{6.2}{\maxBLEU} & \deltacell{87.3}{-0.3}{\maxCOMET} & \deltacell{37.1}{-0.2}{\maxBLEU} & \deltacell{56.9}{-9.9}{\maxOLMES} & \deltacell{25.5}{-41.2}{\maxMTIF} \\
& SPD & $\lambda$=5 & \deltacell{86.6}{2.2}{\maxCOMET} & \deltacell{32.0}{5.6}{\maxBLEU} & \deltacell{87.5}{-0.1}{\maxCOMET} & \deltacell{37.5}{0.2}{\maxBLEU} & \deltacell{63.2}{-3.6}{\maxOLMES} & \deltacell{39.7}{-27.1}{\maxMTIF} \\
& LoRA & Rank=32, Alpha=64 & \deltacell{86.2}{1.8}{\maxCOMET} & \deltacell{31.1}{4.7}{\maxBLEU} & \deltacell{87.4}{-0.2}{\maxCOMET} & \deltacell{37.2}{-0.1}{\maxBLEU} & \deltacell{62.0}{-4.8}{\maxOLMES} & \deltacell{27.3}{-39.5}{\maxMTIF} \\
\bottomrule
\end{tabular}%
}
\caption{MT, General Avg, and Avg MT-IF scores for bidirectional Arabic training on the Llama 3.1 8B Instruct model. Color shade of cells encodes the absolute difference to the base model, normalized within each metric group (across languages).}
\label{tab:arabic-bidrectional-overview}
\end{table*}

\renewcommand{\maxCOMETavg}{12.1}
\renewcommand{\maxPct}{49.44}

\begin{table*}[t]
\centering
\small
\fitwidth{%
\begin{tabular}{lllcccccccc}
\toprule
& & & \multicolumn{8}{c}{ar$\leftrightarrow$en} \\
\cmidrule(lr){4-11}
& & & \multicolumn{2}{c}{Baseline} & \multicolumn{3}{c}{Shorter} & \multicolumn{3}{c}{Longer} \\
\cmidrule(lr){4-5} \cmidrule(lr){6-8} \cmidrule(lr){9-11}
Anchoring & Method & Hyperparameters & Avg. COMET & Tgt/Src & Avg. COMET & \% Shorter & Tgt/Src & Avg. COMET & \% Longer & Tgt/Src \\
\midrule
\multicolumn{3}{l}{Llama 3.1 8B Instruct} & 84.4 & 1.069 & 84.0 & 78.4 & 0.942 & 83.2 & 66.8 & 1.329 \\
\midrule
\multicolumn{3}{l}{Standard SFT} & \deltacell{86.7}{2.3}{\maxCOMETavg} & 1.006 & \deltacell{84.5}{0.5}{\maxCOMETavg} & \deltacell{34.3}{-44.1}{\maxPct} & 1.003 & \deltacell{78.3}{-4.9}{\maxCOMETavg} & \deltacell{33.2}{-33.6}{\maxPct} & 1.061 \\
\midrule
\multirow{9}{*}{\shortstack[l]{Auxiliary\\Data}} & \multirow{6}{*}{\shortstack[l]{Data\\Mixing}} & Tülu (full) & \deltacell{86.6}{2.2}{\maxCOMETavg} & 0.998 & \deltacell{86.3}{2.3}{\maxCOMETavg} & \deltacell{37.7}{-40.7}{\maxPct} & 0.978 & \deltacell{82.7}{-0.5}{\maxCOMETavg} & \deltacell{30.0}{-36.8}{\maxPct} & 0.997 \\
& & Tülu (800) + MT-IF (1600) & \deltacell{86.8}{2.4}{\maxCOMETavg} & 1.005 & \deltacell{86.3}{2.3}{\maxCOMETavg} & \deltacell{36.3}{-42.1}{\maxPct} & 0.982 & \deltacell{80.0}{-3.2}{\maxCOMETavg} & \deltacell{32.1}{-34.7}{\maxPct} & 1.021 \\
& & \quad + diff MT-IF prompt & \deltacell{86.8}{2.4}{\maxCOMETavg} & 1.004 & \deltacell{85.1}{1.1}{\maxCOMETavg} & \deltacell{37.5}{-40.9}{\maxPct} & 0.981 & \deltacell{77.5}{-5.7}{\maxCOMETavg} & \deltacell{31.2}{-35.6}{\maxPct} & 1.018 \\
& & MT-IF (1600) & \deltacell{86.7}{2.3}{\maxCOMETavg} & 1.005 & \deltacell{86.4}{2.4}{\maxCOMETavg} & \deltacell{36.5}{-41.9}{\maxPct} & 0.991 & \deltacell{82.4}{-0.8}{\maxCOMETavg} & \deltacell{36.5}{-30.3}{\maxPct} & 1.060 \\
& & Tülu (800) + Gender (800) & \deltacell{86.6}{2.2}{\maxCOMETavg} & 1.005 & \deltacell{86.4}{2.4}{\maxCOMETavg} & \deltacell{34.3}{-44.1}{\maxPct} & 0.996 & \deltacell{82.5}{-0.7}{\maxCOMETavg} & \deltacell{30.1}{-36.7}{\maxPct} & 1.012 \\
& & Tülu (800) + Formality (800) & \deltacell{86.7}{2.3}{\maxCOMETavg} & 1.004 & \deltacell{86.2}{2.2}{\maxCOMETavg} & \deltacell{32.4}{-46.0}{\maxPct} & 1.002 & \deltacell{81.1}{-2.1}{\maxCOMETavg} & \deltacell{30.6}{-36.2}{\maxPct} & 1.012 \\
\cmidrule(lr){2-11}
& Merge & Trans=0.5, Tülu+MT-IF=0.5 & \deltacell{86.3}{1.9}{\maxCOMETavg} & 1.005 & \deltacell{86.2}{2.2}{\maxCOMETavg} & \deltacell{35.2}{-43.2}{\maxPct} & 0.987 & \deltacell{84.5}{1.3}{\maxCOMETavg} & \deltacell{29.0}{-37.8}{\maxPct} & 1.002 \\
\cmidrule(lr){2-11}
& \multirow{3}{*}{EWC} & $\lambda$=1e3, Tülu (10k) & \deltacell{86.6}{2.2}{\maxCOMETavg} & 1.007 & \deltacell{80.5}{-3.5}{\maxCOMETavg} & \deltacell{43.0}{-35.4}{\maxPct} & 0.989 & \deltacell{71.1}{-12.1}{\maxCOMETavg} & \deltacell{43.0}{-23.8}{\maxPct} & 1.218 \\
& & $\lambda$=1e3, Tülu (800) + MT-IF (1600) & \deltacell{86.6}{2.2}{\maxCOMETavg} & 1.009 & \deltacell{83.5}{-0.5}{\maxCOMETavg} & \deltacell{35.2}{-43.2}{\maxPct} & 1.021 & \deltacell{77.5}{-5.7}{\maxCOMETavg} & \deltacell{38.5}{-28.3}{\maxPct} & 1.100 \\
& & $\lambda$=1e3, MT-IF (1600) & \deltacell{86.7}{2.3}{\maxCOMETavg} & 1.006 & \deltacell{83.3}{-0.7}{\maxCOMETavg} & \deltacell{37.0}{-41.4}{\maxPct} & 1.016 & \deltacell{77.1}{-6.1}{\maxCOMETavg} & \deltacell{39.0}{-27.8}{\maxPct} & 1.093 \\
\midrule
\multirow{4}{*}{\shortstack[l]{Model\\Outputs}} & KL-Full & $\lambda$=0.05 & \deltacell{86.8}{2.4}{\maxCOMETavg} & 1.012 & \deltacell{85.4}{1.4}{\maxCOMETavg} & \deltacell{32.7}{-45.7}{\maxPct} & 1.009 & \deltacell{79.1}{-4.1}{\maxCOMETavg} & \deltacell{36.1}{-30.7}{\maxPct} & 1.099 \\
& STM & X=0.25 & \deltacell{86.2}{1.8}{\maxCOMETavg} & 1.003 & \deltacell{84.1}{0.1}{\maxCOMETavg} & \deltacell{35.5}{-42.9}{\maxPct} & 0.982 & \deltacell{76.1}{-7.1}{\maxCOMETavg} & \deltacell{27.6}{-39.2}{\maxPct} & 0.999 \\
& CC & X=0.25 & \deltacell{86.8}{2.4}{\maxCOMETavg} & 1.011 & \deltacell{85.4}{1.4}{\maxCOMETavg} & \deltacell{33.1}{-45.3}{\maxPct} & 1.006 & \deltacell{80.0}{-3.2}{\maxCOMETavg} & \deltacell{31.1}{-35.7}{\maxPct} & 1.057 \\
& EAFT & K=20, Weight=1.0 & \deltacell{86.7}{2.3}{\maxCOMETavg} & 1.008 & \deltacell{85.7}{1.7}{\maxCOMETavg} & \deltacell{32.4}{-46.0}{\maxPct} & 0.993 & \deltacell{79.5}{-3.7}{\maxCOMETavg} & \deltacell{28.1}{-38.7}{\maxPct} & 1.006 \\
\midrule
\multirow{3}{*}{\shortstack[l]{Base Model\\Parameters}} & Freeze & N=4 & \deltacell{86.6}{2.2}{\maxCOMETavg} & 1.004 & \deltacell{86.5}{2.5}{\maxCOMETavg} & \deltacell{32.9}{-45.5}{\maxPct} & 0.990 & \deltacell{80.1}{-3.1}{\maxCOMETavg} & \deltacell{32.3}{-34.5}{\maxPct} & 1.045 \\
& SPD & $\lambda$=5 & \deltacell{86.5}{2.1}{\maxCOMETavg} & 1.009 & \deltacell{83.2}{-0.8}{\maxCOMETavg} & \deltacell{49.8}{-28.6}{\maxPct} & 0.988 & \deltacell{75.2}{-8.0}{\maxCOMETavg} & \deltacell{49.4}{-17.4}{\maxPct} & 1.183 \\
& LoRA & Rank=32, Alpha=64 & \deltacell{86.1}{1.7}{\maxCOMETavg} & 1.011 & \deltacell{81.2}{-2.8}{\maxCOMETavg} & \deltacell{41.1}{-37.3}{\maxPct} & 0.977 & \deltacell{73.9}{-9.3}{\maxCOMETavg} & \deltacell{28.4}{-38.4}{\maxPct} & 0.999 \\
\bottomrule
\end{tabular}%
}
\caption{Effect of length instructions (none/shorter/longer) on length control scores for Arabic $\leftrightarrow$ English translation with Llama 3.1 8B Instruct, evaluated on FLORES (bidirectional, 2x N=$1012$).}
\label{tab:arabic-length-control}
\end{table*}

\begin{table*}[t]
\centering
\small
\renewcommand{\maxCOMETxling}{4.7}
\renewcommand{\maxBLEUxling}{8.4}
\renewcommand{\maxFemAcc}{57.3}
\renewcommand{\maxMascAcc}{43.4}
\renewcommand{\maxAvgAcc}{50.0}
\fitwidth{%
\begin{tabular}{lllccccccccccc}
\toprule
& & & \multicolumn{5}{c}{en$\rightarrow$ar (N=300)} & \multicolumn{5}{c}{en$\rightarrow$de (N=300)} \\
\cmidrule(lr){4-8} \cmidrule(lr){9-13}
Anchoring & Method & Hyperparameters & COMET & BLEU & Fem. Acc. & Masc. Acc. & Avg. & COMET & BLEU & Fem. Acc. & Masc. Acc. & Avg. \\
\midrule
\multicolumn{3}{l}{Llama 3.1 8B Instruct} & 76.6 & 17.3 & 70.3 & 51.7 & 61.0 & 81.2 & 32.8 & 63.3 & 50.7 & 57.0 \\
\midrule
\multicolumn{3}{l}{Standard SFT} & \deltacell{76.2}{-0.4}{\maxCOMETxling} & \deltacell{16.0}{-1.3}{\maxBLEUxling} & \deltacell{30.0}{-40.3}{\maxFemAcc} & \deltacell{18.0}{-33.7}{\maxMascAcc} & \deltacell{24.0}{-37.0}{\maxAvgAcc} & \deltacell{80.3}{-0.9}{\maxCOMETxling} & \deltacell{28.6}{-4.2}{\maxBLEUxling} & \deltacell{8.3}{-55.0}{\maxFemAcc} & \deltacell{8.7}{-42.0}{\maxMascAcc} & \deltacell{8.5}{-48.5}{\maxAvgAcc} \\
\midrule
\multirow{9}{*}{\shortstack[l]{Auxiliary\\Data}} & \multirow{6}{*}{\shortstack[l]{Data\\Mixing}} & Tülu (full) & \deltacell{78.9}{2.3}{\maxCOMETxling} & \deltacell{17.6}{0.3}{\maxBLEUxling} & \deltacell{23.0}{-47.3}{\maxFemAcc} & \deltacell{17.0}{-34.7}{\maxMascAcc} & \deltacell{20.0}{-41.0}{\maxAvgAcc} & \deltacell{82.2}{1.0}{\maxCOMETxling} & \deltacell{31.7}{-1.1}{\maxBLEUxling} & \deltacell{10.0}{-53.3}{\maxFemAcc} & \deltacell{12.3}{-38.4}{\maxMascAcc} & \deltacell{11.2}{-45.8}{\maxAvgAcc} \\
& & Tülu (800) + MT-IF (1600) & \deltacell{80.9}{4.3}{\maxCOMETxling} & \deltacell{21.2}{3.9}{\maxBLEUxling} & \deltacell{73.7}{3.4}{\maxFemAcc} & \deltacell{65.0}{13.3}{\maxMascAcc} & \deltacell{69.4}{8.4}{\maxAvgAcc} & \deltacell{85.2}{4.0}{\maxCOMETxling} & \deltacell{41.2}{8.4}{\maxBLEUxling} & \deltacell{79.7}{16.4}{\maxFemAcc} & \deltacell{80.0}{29.3}{\maxMascAcc} & \deltacell{79.9}{22.9}{\maxAvgAcc} \\
& & \quad + diff MT-IF prompt & \deltacell{79.5}{2.9}{\maxCOMETxling} & \deltacell{17.9}{0.6}{\maxBLEUxling} & \deltacell{23.0}{-47.3}{\maxFemAcc} & \deltacell{17.3}{-34.4}{\maxMascAcc} & \deltacell{20.2}{-40.8}{\maxAvgAcc} & \deltacell{83.3}{2.1}{\maxCOMETxling} & \deltacell{35.2}{2.4}{\maxBLEUxling} & \deltacell{36.7}{-26.6}{\maxFemAcc} & \deltacell{38.0}{-12.7}{\maxMascAcc} & \deltacell{37.3}{-19.7}{\maxAvgAcc} \\
& & MT-IF (1600) & \deltacell{80.8}{4.2}{\maxCOMETxling} & \deltacell{22.5}{5.2}{\maxBLEUxling} & \deltacell{73.3}{3.0}{\maxFemAcc} & \deltacell{74.0}{22.3}{\maxMascAcc} & \deltacell{73.7}{12.7}{\maxAvgAcc} & \deltacell{83.8}{2.6}{\maxCOMETxling} & \deltacell{39.1}{6.3}{\maxBLEUxling} & \deltacell{80.7}{17.4}{\maxFemAcc} & \deltacell{83.0}{32.3}{\maxMascAcc} & \deltacell{81.8}{24.8}{\maxAvgAcc} \\
& & Tülu (800) + Gender (800) & \deltacell{80.9}{4.3}{\maxCOMETxling} & \deltacell{22.7}{5.4}{\maxBLEUxling} & \deltacell{72.7}{2.4}{\maxFemAcc} & \deltacell{75.0}{23.3}{\maxMascAcc} & \deltacell{73.8}{12.8}{\maxAvgAcc} & \deltacell{83.9}{2.7}{\maxCOMETxling} & \deltacell{39.1}{6.3}{\maxBLEUxling} & \deltacell{81.3}{18.0}{\maxFemAcc} & \deltacell{82.7}{32.0}{\maxMascAcc} & \deltacell{82.0}{25.0}{\maxAvgAcc} \\
& & Tülu (800) + Formality (800) & \deltacell{79.1}{2.5}{\maxCOMETxling} & \deltacell{17.7}{0.4}{\maxBLEUxling} & \deltacell{19.0}{-51.3}{\maxFemAcc} & \deltacell{15.7}{-36.0}{\maxMascAcc} & \deltacell{17.3}{-43.7}{\maxAvgAcc} & \deltacell{80.7}{-0.5}{\maxCOMETxling} & \deltacell{29.9}{-2.9}{\maxBLEUxling} & \deltacell{7.3}{-56.0}{\maxFemAcc} & \deltacell{9.7}{-41.0}{\maxMascAcc} & \deltacell{8.5}{-48.5}{\maxAvgAcc} \\
\cmidrule(lr){2-13}
& Merge & Trans=0.5, Tülu+MT-IF=0.5 & \deltacell{79.7}{3.1}{\maxCOMETxling} & \deltacell{18.4}{1.1}{\maxBLEUxling} & \deltacell{42.3}{-28.0}{\maxFemAcc} & \deltacell{26.3}{-25.4}{\maxMascAcc} & \deltacell{34.3}{-26.7}{\maxAvgAcc} & \deltacell{83.0}{1.8}{\maxCOMETxling} & \deltacell{34.9}{2.1}{\maxBLEUxling} & \deltacell{36.0}{-27.3}{\maxFemAcc} & \deltacell{35.3}{-15.4}{\maxMascAcc} & \deltacell{35.7}{-21.3}{\maxAvgAcc} \\
\cmidrule(lr){2-13}
& \multirow{3}{*}{EWC} & $\lambda$=1e3, Tülu (10k) & \deltacell{76.8}{0.2}{\maxCOMETxling} & \deltacell{16.0}{-1.3}{\maxBLEUxling} & \deltacell{23.3}{-47.0}{\maxFemAcc} & \deltacell{21.7}{-30.0}{\maxMascAcc} & \deltacell{22.5}{-38.5}{\maxAvgAcc} & \deltacell{81.1}{-0.1}{\maxCOMETxling} & \deltacell{29.9}{-2.9}{\maxBLEUxling} & \deltacell{13.7}{-49.6}{\maxFemAcc} & \deltacell{11.3}{-39.4}{\maxMascAcc} & \deltacell{12.5}{-44.5}{\maxAvgAcc} \\
& & $\lambda$=1e3, Tülu (800) + MT-IF (1600) & \deltacell{78.2}{1.6}{\maxCOMETxling} & \deltacell{17.1}{-0.2}{\maxBLEUxling} & \deltacell{20.3}{-50.0}{\maxFemAcc} & \deltacell{18.7}{-33.0}{\maxMascAcc} & \deltacell{19.5}{-41.5}{\maxAvgAcc} & \deltacell{81.3}{0.1}{\maxCOMETxling} & \deltacell{29.8}{-3.0}{\maxBLEUxling} & \deltacell{11.0}{-52.3}{\maxFemAcc} & \deltacell{9.0}{-41.7}{\maxMascAcc} & \deltacell{10.0}{-47.0}{\maxAvgAcc} \\
& & $\lambda$=1e3, MT-IF (1600) & \deltacell{78.1}{1.5}{\maxCOMETxling} & \deltacell{17.3}{0}{\maxBLEUxling} & \deltacell{21.7}{-48.6}{\maxFemAcc} & \deltacell{19.3}{-32.4}{\maxMascAcc} & \deltacell{20.5}{-40.5}{\maxAvgAcc} & \deltacell{81.1}{-0.1}{\maxCOMETxling} & \deltacell{29.6}{-3.2}{\maxBLEUxling} & \deltacell{11.0}{-52.3}{\maxFemAcc} & \deltacell{8.7}{-42.0}{\maxMascAcc} & \deltacell{9.9}{-47.1}{\maxAvgAcc} \\
\midrule
\multirow{3}{*}{\shortstack[l]{Model\\Outputs}} & KL Full & $\lambda$=0.05 & \deltacell{78.7}{2.1}{\maxCOMETxling} & \deltacell{17.2}{-0.1}{\maxBLEUxling} & \deltacell{19.3}{-51.0}{\maxFemAcc} & \deltacell{16.3}{-35.4}{\maxMascAcc} & \deltacell{17.8}{-43.2}{\maxAvgAcc} & \deltacell{80.9}{-0.3}{\maxCOMETxling} & \deltacell{29.5}{-3.3}{\maxBLEUxling} & \deltacell{7.3}{-56.0}{\maxFemAcc} & \deltacell{8.3}{-42.7}{\maxMascAcc} & \deltacell{7.8}{-49.2}{\maxAvgAcc} \\
& STM & X=0.25 & \deltacell{74.9}{-1.7}{\maxCOMETxling} & \deltacell{14.0}{-3.3}{\maxBLEUxling} & \deltacell{28.7}{-41.6}{\maxFemAcc} & \deltacell{28.3}{-23.4}{\maxMascAcc} & \deltacell{28.5}{-32.5}{\maxAvgAcc} & \deltacell{80.5}{-0.7}{\maxCOMETxling} & \deltacell{28.0}{-4.8}{\maxBLEUxling} & \deltacell{8.0}{-55.3}{\maxFemAcc} & \deltacell{8.0}{-42.7}{\maxMascAcc} & \deltacell{8.0}{-49.0}{\maxAvgAcc} \\
& CC & X=0.25 & \deltacell{78.3}{1.7}{\maxCOMETxling} & \deltacell{16.7}{-0.6}{\maxBLEUxling} & \deltacell{20.0}{-50.3}{\maxFemAcc} & \deltacell{16.7}{-35.0}{\maxMascAcc} & \deltacell{18.3}{-42.7}{\maxAvgAcc} & \deltacell{79.8}{-1.4}{\maxCOMETxling} & \deltacell{28.2}{-4.6}{\maxBLEUxling} & \deltacell{8.3}{-55.0}{\maxFemAcc} & \deltacell{10.3}{-40.4}{\maxMascAcc} & \deltacell{9.3}{-47.7}{\maxAvgAcc} \\
& EAFT & K=20, Weight=1.0 & \deltacell{78.4}{1.8}{\maxCOMETxling} & \deltacell{17.0}{-0.3}{\maxBLEUxling} & \deltacell{19.7}{-50.6}{\maxFemAcc} & \deltacell{19.3}{-32.4}{\maxMascAcc} & \deltacell{19.5}{-41.5}{\maxAvgAcc} & \deltacell{80.8}{-0.4}{\maxCOMETxling} & \deltacell{29.6}{-3.2}{\maxBLEUxling} & \deltacell{6.7}{-56.6}{\maxFemAcc} & \deltacell{7.3}{-43.4}{\maxMascAcc} & \deltacell{7.0}{-50.0}{\maxAvgAcc} \\
\midrule
\multirow{3}{*}{\shortstack[l]{Base Model\\Parameters}} & Freeze & N=4 & \deltacell{79.2}{2.6}{\maxCOMETxling} & \deltacell{18.1}{0.8}{\maxBLEUxling} & \deltacell{20.0}{-50.3}{\maxFemAcc} & \deltacell{17.0}{-34.7}{\maxMascAcc} & \deltacell{18.5}{-42.5}{\maxAvgAcc} & \deltacell{80.5}{-0.7}{\maxCOMETxling} & \deltacell{28.5}{-4.3}{\maxBLEUxling} & \deltacell{10.0}{-53.3}{\maxFemAcc} & \deltacell{10.0}{-40.7}{\maxMascAcc} & \deltacell{10.0}{-47.0}{\maxAvgAcc} \\
& SPD & $\lambda$=5 & \deltacell{79.6}{3.0}{\maxCOMETxling} & \deltacell{18.3}{1.0}{\maxBLEUxling} & \deltacell{35.0}{-35.3}{\maxFemAcc} & \deltacell{24.3}{-27.4}{\maxMascAcc} & \deltacell{29.7}{-31.3}{\maxAvgAcc} & \deltacell{82.0}{0.8}{\maxCOMETxling} & \deltacell{32.1}{-0.7}{\maxBLEUxling} & \deltacell{31.3}{-32.0}{\maxFemAcc} & \deltacell{24.7}{-26.0}{\maxMascAcc} & \deltacell{28.0}{-29.0}{\maxAvgAcc} \\
& LoRA & Rank=32, Alpha=64 & \deltacell{78.4}{1.8}{\maxCOMETxling} & \deltacell{16.7}{-0.6}{\maxBLEUxling} & \deltacell{21.7}{-48.6}{\maxFemAcc} & \deltacell{18.0}{-33.7}{\maxMascAcc} & \deltacell{19.8}{-41.2}{\maxAvgAcc} & \deltacell{80.8}{-0.4}{\maxCOMETxling} & \deltacell{28.9}{-3.9}{\maxBLEUxling} & \deltacell{6.0}{-57.3}{\maxFemAcc} & \deltacell{8.0}{-42.7}{\maxMascAcc} & \deltacell{7.0}{-50.0}{\maxAvgAcc} \\
\bottomrule
\end{tabular}%
}
\caption{Grammatical Gender Control scores for bidirectional Arabic training on the Llama 3.1 8B Instruct model. Fem. Acc. and Masc. Acc. are the accuracies of producing the requested feminine and masculine forms.}
\label{tab:arabic-gender-control}
\end{table*}

\end{document}